\documentclass[mnsc,nonblindrev]{informs3}
\OneAndAHalfSpacedXI

\usepackage{natbib}
 \bibpunct[, ]{(}{)}{,}{a}{}{,}%
 \def\bibfont{\small}%

\TheoremsNumberedThrough     
\ECRepeatTheorems

\EquationsNumberedThrough    

\usepackage{amssymb}
\usepackage{mathtools}
\usepackage{thmtools}
\usepackage{thm-restate}
\usepackage{enumitem}
\usepackage[mathscr]{euscript}
\usepackage{xargs}
\usepackage[ruled]{algorithm2e}
\usepackage{float}

\newcommand{\shrinkrow}{\renewcommand{\arraystretch}{0.5}}

\newcommand{\defeq}{\vcentcolon=}
\newcommand{\eqdef}{=\vcentcolon}

\newcommand{\EDi}{\mathbb{E}_{\{(\mathbf{x}_{j},t_{j},\delta_{j})\}_{j=1}^{N_i}\sim \mathcal{D}_i}}

\newcommand{\EDixy}{\mathbb{E}_{\{(\mathbf{x}_{j},t_j,\delta_j)\}_{j=1}^{N_i}\sim \mathcal{D}_i}}
\newcommand{\constraintterma}[2][\{(\mathbf{x}_{j},t_j,\delta_j)\}_{j=1}^{N_i}\sim \mathcal{D}_i]{\mathbb{E}_{{#1}}\left[\left(\frac{1}{N_i}\sum_{j=1}^{N_i} {#2}-\SKM\right)^2 \right] }
\newcommand{\Samplexy}[1][N_i]{\{\mathbf{x}_{j},t_{j},\delta_{j}\}_{j=1}^{{#1}}}

\newcommand{\LDRSA}[1][\phi]{\mathcal{L}_{DRSA}(#1,\,\mathbf{x},t,\delta)}

\newcommand{\SKM}[1][t]{S_{KM}(#1;\{(\mathbf{x}_{j},t_{j},\delta_{ j})\}_{j=1}^{{N_i}})}

\newcommand{\SKMProp}[1][t]{S_{KM}(#1;\{(\mathbf{x}_{n_i,j}^i,t_{n_i,j}^i,\delta_{n_i, j}^i)\}_{j=1}^{{N_i}})}
\newcommand{\GRADUATE}{\mbox{GRADUATE}\xspace}
\newcommand{\MCBoostSurv}{\mbox{MCBoostSurv}\xspace}

\begin{document}


\RUNAUTHOR{Suttaket and Kok}

\RUNTITLE{Constrained Optimization for Multicalibrated Survival Analysis}

\TITLE{In-Training Multicalibrated Survival Analysis for Healthcare via Constrained Optimization}

\ARTICLEAUTHORS{%
\AUTHOR{Thiti Suttaket, Stanley Kok}
\AFF{Department of Information Systems and Analytics, National University of Singapore, 13 Computing Drive, Singapore 117417, \EMAIL{e0384140@u.nus.edu},\EMAIL{skok@comp.nus.edu.sg}} 
} 
\ABSTRACT{%
Survival analysis is an important problem in healthcare because it models the relationship between an individual’s covariates and the onset time of an event of interest (e.g., death). It is important for survival models to be \emph{well-calibrated} (i.e., for their predicted probabilities to be close to ground-truth probabilities) because badly calibrated systems can result in erroneous clinical decisions. Existing survival models are typically calibrated at the population level only, and thus run the risk of being poorly calibrated for one or more minority subpopulations. We propose a model called GRADUATE that achieves \emph{multicalibration} by ensuring that all subpopulations are well-calibrated too. GRADUATE frames multicalibration as a constrained optimization problem, and optimizes both calibration and discrimination in-training to achieve a good balance between them. We mathematically prove that the optimization method used yields a solution that is both near-optimal and feasible with high probability. Empirical comparisons against state-of-the-art baselines on real-world clinical datasets demonstrate GRADUATE's efficacy. In a detailed analysis, we elucidate the shortcomings of the baselines vis-\`{a}-vis  GRADUATE's strengths.
}%


\maketitle
\section{Introduction}

Information systems are pervasively used in medical management tasks and clinical care applications such as physician staffing~\citep{lipscomb1998combining},
treatment planning~\citep{chen2021designing}, and predictive modeling~\citep{harutyunyan2019multitask}. 
For the task of predictive modeling, many systems have narrowly focused on predicting the onset of an event (e.g., disease) at a \emph{fixed} time point in the future (e.g., after a year from the point of diagnosis). However, it may be of more utility to generalize the task to one of learning a probability distribution over all possible time points of event onset, a problem that is addressed by \emph{survival analysis}. 

In survival analysis, we model the relationship between an individual’s features (covariates) and the stochastic process underlying some event of interest (e.g., disease onset), and represent the relationship as a probability distribution over the time taken till event occurrence. Systems for survival analysis are typically concerned with discriminative performance, i.e., their ability to correctly \emph{rank} individuals by their predicted probabilities of event occurrence vis-\`{a}-vis observed outcomes. However, especially for clinical applications, it is at least as important for the systems to be \emph{well-calibrated}~\citep{pepe2013methods}, i.e., for their predicted probabilities to be numerically close to ground-truth probabilities at all time points. This is because poorly calibrated systems can provide misleading predictions, which in turn can result in potentially harmful medical decisions~\citep{vancalster2019achilles, vancalster2015calibration}.

State-of-the-art survival analysis systems, such as RPS~\citep{kamran2021estimating} and X-Cal~\citep{goldstein2020x}, are typically built upon deep learning, and are calibrated only marginally, i.e., they are calibrated over the entire population represented in their training data. They adopt the traditional supervised learning approach of choosing a model that minimizes some expected loss on the training data, and thus run the risk of learning a model that is good for the majority sub-population, but perform poorly on one or more minority sub-populations (whose contributions to the expected loss are relatively minuscule). This characteristic can have serious legal and ethical implications if the minority sub-populations are defined by protected traits such as race and gender.  Such problems are well-documented in many applications of information systems such as natural language processing~\citep{bolukbasi2016man} and image classification~\citep{buolamwini2018gender}. Survival analysis systems adopt the same supervised learning paradigm as those systems, and thus suffer from the same ``fairness'' problem. Even if fairness is not an issue, it may still be important to ensure that all sub-populations of interest are well-calibrated. For example, in a dataset of diabetic patients, a small sub-population of patients are administered a specialized medication to prevent eye complications, and it may be of medical interest to make sure that the sub-population is well-calibrated (in conjunction with the entire population) to accurately reflect its probability of developing eye problems over time.

To date, no system ensures that pre-identified sub-populations are well-calibrated in survival analysis except MCBoostSurv~\citep{becker2021multicalibration}. It is a post-hoc system that encapsulates an existing survival analysis model that has already been trained, and \emph{post-processes} the latter's predictions using a validation dataset to achieve multi-subpopulation calibration (\emph{multicalibration} in short). However, because it is a post-hoc system, it decouples the learning of a good predictive (discriminative\footnote{Discriminative in the sense of being able to rank individuals in the correct order by their predicted probabilities of event occurrence, and not in the sense of being biased against people with protected traits.}) model from the search for a well-calibrated one, and does not simultaneously find a good balance between both discrimination and calibration. Furthermore, by being constrained to using a (relatively small) validation dataset for calibration, MCBoostSurv fails to utilize the entire training dataset to better inform its calibration. These traits of MCBoostSurv are in contrast to those of the aforementioned systems of~\cite{kamran2021estimating} and~\cite{goldstein2020x}, which optimize both discrimination and calibration \emph{in-training}, and bring the full training set to bear in simultaneously achieving both good discrimination and good calibration (albeit for the entire population only). 

To overcome the shortcomings of the abovementioned systems, we propose a model called \GRADUATE (short for \textit{\underline{GRAD}ient-\underline{U}tilized calibr\underline{ATE}}) that optimizes both discrimination and calibration \emph{in-training}, and concurrently ensures that its predictions are well-calibrated over all time points for all pre-identified (possibly overlapping) subpopulations of interest. \GRADUATE utilizes constrained optimization~\citep{beck2017first} to minimize a discriminative loss function while posing the multicalibration requirement as constraints, one for each sub-population. Through rigorous mathematical analysis, we demonstrate that this optimization approach yields solutions with a high probability of being both near-optimal and feasible within the multicalibration constraints imposed by the problem.

In sum, our contributions are threefold.
\begin{itemize}
    \item We propose a novel survival analysis model for in-training multicalibration called \GRADUATE. Table~\ref{tab:GRADUATE_contribution} highlights its contribution in comparison to existing survival analysis models. To our knowledge, \GRADUATE is the first system that simultaneously achieves \emph{multicalibration} and a balance between calibration and discrimination \emph{during its training process}.
    \item We rigorously demonstrate, through mathematical analysis, that the optimization approach yields solutions with a high probability of being both near-optimal and feasible within the multicalibration constraints imposed by the constrained optimization problem.
    \item We empirically compare our \GRADUATE model to four state-of-the-art baselines on three real-world medical datasets, and show that \GRADUATE is not only better calibrated both marginally and at the sub-population level, but also achieves the best trade-off between discrimination and calibration.
\end{itemize}

\bgroup
\def\arraystretch{1.3}
\footnotesize
\begin{table}[ht]
    \centering
    \footnotesize
\begin{tabular}{|c|>{\centering\arraybackslash}p{2.1cm}|>{\centering\arraybackslash}p{2.1cm}|>{\centering\arraybackslash}p{2.3cm}|>{\centering\arraybackslash}p{2.5cm}|} 
\cline{1-5}
 
  & \multicolumn{4}{c|}{\normalsize{\textbf{Model  Attributes}}}\rule{0pt}{3ex} \\ [2pt] \cline{2-5}
 & 
  Discrimination &
  \begin{tabular}[c]{@{}c@{}}Marginal \\Calibration\end{tabular} &
  Multicalibration &
\begin{tabular}[c]{@{}c@{}}In-Training \\(Multi)calibration\end{tabular}
 \\
  \hline
\multicolumn{1}{|c|}{\shrinkrow\begin{tabular}[c]{@{}c@{}}Discriminative\\ Survival \\Analysis Model\end{tabular}}        
            & \checkmark
            & \scalebox{1.5}{$\times$}
            & \scalebox{1.5}{$\times$}
            & \scalebox{1.5}{$\times$}
            \\ \hline
\multicolumn{1}{|c|}{X-Cal}   
            & \checkmark
            & \checkmark
            & \scalebox{1.5}{$\times$}
            & \checkmark
            \\ \hline  
\multicolumn{1}{|c|}{MCBoostSurv}   
            & \checkmark
            & \checkmark
            & \checkmark
            & \scalebox{1.5}{$\times$}
            \\ \hline  
\multicolumn{1}{|c|}{RPS}
            & \checkmark
            & \checkmark
            & \scalebox{1.5}{$\times$}
            & \checkmark
            \\ \hline
\multicolumn{1}{|c|}{GRADUATE}
            & \checkmark
            & \checkmark
            & \checkmark
            & \checkmark
            \\ \hline
\end{tabular}
\vspace{0.3cm}
    \caption{Model Attributes Comparison.}
	\label{tab:GRADUATE_contribution}
\end{table}
\egroup
The remainder of this paper is structured as follows:

Section~\ref{sec:related_work} reviews related work in the area of survival analysis and calibration approaches in this domain.

Section~\ref{sec:background} establishes essential background for our work. It covers problem formulation, the Kaplan-Meier estimator, and Deep Recurrent Survival Analysis (DRSA; \cite{ren2019deep}), a model achieving state-of-the-art discriminative performance in survival analysis.

Section~\ref{sec:GRADUATE_model} presents \GRADUATE, our proposed in-training multicalibrated survival analysis model, and its corresponding optimization algorithm.

Section~\ref{sec:proof} presents the mathematical proofs that underpin the theoretical aspects of our methodology.

Section~\ref{sec:experiments} describes the experimental setup, including the datasets, evaluation metrics, and results obtained from applying the proposed method. The limitations of the baseline methods compared to our proposed approach is presented in Section~\ref{sec:discussion}.

Section \ref{sec:practical_implications} explores the practical implications of our proposed model, \GRADUATE. It discusses how it can be applied in real-world healthcare settings, and the potential benefits in clinical practices. 

Finally, Section~\ref{sec:conclusion} summarizes the key findings of this study and outlines directions for future research.

\section{Related Work} \label{sec:related_work}

Since survival analysis plays a crucial role in healthcare, a wide variety of techniques have been developed for it~\citep{kleinbaum2004survival}. Traditional statistics-based methods can be divided into three categories, viz., non-parametric, semi-parametric, and parametric methods. Non-parametric methods, such as the Kaplan-Meier (KM) estimator~\citep{kaplan1958nonparametric}, are based on counting statistics. Semi-parametric methods, such as the Cox proportional hazards model~\citep{cox1972regression} and its variants (e.g.,~\cite{tibshirani1997lasso}), assume some form for the base probability distribution that has scaling coefficients which can be tuned to optimize survival rate predictions. Parametric methods assume that survival times follow a particular distribution such as the Weibull distribution~\citep{baghestani2016survival} and the Gompertz distribution~\citep{garg1970maximum}. Because these methods are based on counts or assume a rigid distributional form for survival functions, they do not generalize well in real-world scenarios where data are sparse or the distributional assumptions are not valid~\citep{jin2020deep}.

Aside from statistics, machine learning techniques have also been incorporated into models for survival analysis, e.g., survival random forest~\citep{gordon1985tree} and support vector regression~\citep{khan2008support}. In recent years, deep learning has also been utilized for survival analysis. DeepSurv~\citep{katzman2018deepsurv} and the system of~\cite{luck2017deep} are both deep-learning extensions of the standard Cox proportional hazards model. \cite{chapfuwa2018adversarial} adopt a conditional generative adversarial network to predict the event time conditioned on covariates, and implicitly specify a survival distribution via sampling. DeepHit~\citep{lee2018deephit} uses a recurrent neural network to model the relationship between covariates and event occurrence probability over discrete time. Deep Recurrent Survival Analysis (DRSA; \cite{ren2019deep}) represents survival functions using the probability chain rule, and in turn models that as a recurrent neural network. DRSA achieves state-of-the-art discriminative performance for survival analysis.

Realizing the importance of good calibration for survival analysis, several novel methods have recently been proposed to improve the calibration of survival functions. \cite{haider2020effective} use the probability integral transform~\citep{angus1994pit} to measure the uniformity of the cumulative distribution function for survival time. They then use that uniformity to define a metric called \emph{D-Calibration} (Equation~\ref{eq:dcal}) to quantify how well a model is calibrated. \cite{goldstein2020x} propose a differentiable version of D-Calibration called \textit{X-Cal}, and use it as the objective function to train a deep learning model.
\cite{avati2020countdown} extend the continuous ranked probability score (CRPS) from the field of meteorology to handle censored data, and develop a new objective function for training survival models called \emph{survival-CRPS}. However, survival-CRPS requires potentially restrictive distributional assumptions during training. After that, \cite{kamran2021estimating} propose to minimize a CRPS variant called the \emph{rank probability score} (RPS) to calibrate survival functions. Intuitively, when minimizing either RPS or survival-CRPS for uncensored individuals, we push to 1 and 0 the values of a survival function at times respectively before and after the ground-truth onset of an event. Similarly, for censored individuals, the survival function at times before censoring are pushed to values of 1. All these methods are concerned with calibration at the level of the entire population only.

In the field of algorithmic fairness, \cite{hebert2018multicalibration} introduce an algorithm to ensure that the predictions of a binary classifier is multicalibrated. The algorithm post-processes the predictions of the binary classifier using a held-out validation set to ensure that the predictions are well-calibrated for multiple pre-identified protected subgroups of individuals in the data.
\MCBoostSurv~\citep{becker2021multicalibration} extends the multicalibration algorithm to survival analysis by optimizing the integrated Brier score. 

\section{Background}\label{sec:background}

\subsection{Problem Setup}

In survival analysis, we seek to learn a time-to-event model in a supervised fashion. In contrast to traditional supervised learning, the data in survival analysis take the form of $D\defeq\{\,(\mathbf{x}_i,t_i,\delta_i)\,\}_{i=1}^n$, 
where $n$ is the number of individuals (or examples). These $n$ examples jointly follow a distribution denoted as $\mathcal{D}$, and each individual triplet $(\mathbf{x}_i, t_i, \delta_i)$ follows a distribution $\mathfrak{D}$. However, $\{(\mathbf{x}_i, t_i, \delta_i)\}_{i=1}^n$ are assumed to be independent, it is therefore common to denote $\mathcal{D}$ as $\mathfrak{D}^n$. Here, 
$\mathbf{x}_i\!\in\!\mathbb{R}^d$, $\delta_i$, and $t_i$ respectively denote 
individual $i$'s features (covariates), censoring status, and observed event onset time or censoring time (whichever is earlier). We term data of such form \emph{survival data}. To elucidate $\delta_i$ and $t_i$, let $\tilde{t}_i$ be the ground-truth occurrence time of an event of interest, and let $c_i$ be the ground-truth censoring time after which data are unavailable for individual $i$  (e.g., the time of the last follow up with individual $i$ in a medical study). If $\tilde{t}_i \leq c_i$, then the event of interest is observed, and we set $\delta_i = 1$ and $t_i =\tilde{t}_i$; otherwise the event is not observed before individual $i$'s data is censored\footnote{This is known as \emph{right} censoring because the event occurs \emph{after} the censoring time.}, and we set $\delta_i = 0$ and $t_i = c_i$. To reiterate, if $\delta_i=0$, we do not observe the actual time of event occurrence for individual $i$. 
(As in standard practice, we assume $\tilde{t}_i$ and $c_i$ are independent given individual $i$'s features $\mathbf{x}_i$, i.e., $\tilde{t}_i \perp c_i | \mathbf{x}_i$.) 

Following recent work~\citep{lee2018deephit,ren2019deep,kamran2021estimating},
we focus on \emph{discrete-time} survival analysis, in which a time horizon is discretized into $\tau$ intervals $\{1,\ldots,\tau\}$. Given data $D$, we aim to learn a model that maps individual $i$'s features $\mathbf{x}_i$ to individualized estimates of $P(T\!=\!t|\mathbf{x}_i)$ where $T$ is a random variable representing the (discrete) time when an event of interest occurred, and \linebreak $t\in \{1,\ldots,\tau\}$. Using these estimates, we derive the \emph{survival curve} or \emph{survival function} \linebreak
$S(t|\mathbf{x}_i) \defeq  P(T\!>\!t|\mathbf{x}_i)=\sum_{j>t} P(T\!=\!j|\mathbf{x}_i)$.
Similarly, we can estimate the \emph{cumulative incidence function} or \emph{cumulative distribution function} $F(t|\mathbf{x}_i)\defeq 1-S(t|\mathbf{x}_i)=\sum_{j\!\leq\!t} P(T\!=\!j|\mathbf{x}_i)$. 
Intuitively, $S(t|\mathbf{x}_i)$ is the probability of not observing an event until time $t$ for individual $i$ with features $\mathbf{x}_i$, and conversely, $F(t|\mathbf{x}_i)$ is the probability of observing the event within time $t$ for individual $i$.

\subsection{Kaplan-Meier Estimator}

One of the earliest techniques for deriving the survival function is the Kaplan-Meier (KM) estimator~\citep{kaplan1958nonparametric}. Given survival data $\{\,(\mathbf{x}_i,t_i,\delta_i)\,\}_{i=1}^n$, the KM estimator first considers the unique times of event onset $\{\tau_i\}_{i=1}^N$. Then it counts the number of event onsets $e_i$ at time $\tau_i$ as
$e_i \defeq \sum_{j=1}^n \mathbb{I}[t_j = \tau_i] \delta_j$, and calculates the number of individuals $k_i$ at time $\tau_i$ for whom the event of interest could occur as 
$k_i \defeq \sum_{j=1}^n \mathbb{I}[t_j \geq \tau_i]$ (where $\mathbb{I}$ is the indicator function). Finally, the KM estimator computes the survival function as

\begin{equation}\label{eq:bg_kaplan_meier}
S_{KM}(t|\{\,(\mathbf{x}_i,t_i,\delta_i)\,\}_{i=1}^n) \defeq \prod_{i=1}^{N} \left( 1- \frac{e_i}{k_i}\right)^{\mathbb{I}[\tau_i \leq t]} .
\end{equation}

Intuitively, the KM estimator is simply the product of the probabilities of the event of interest \emph{not} occurring from time 1 to $\tau_1$, from time $\tau_1$ to $\tau_2$, and so on, up till time $t$. As $t$ increases, Equation~\ref{eq:bg_kaplan_meier} multiplies more probabilities and thus gets smaller. Consequently, $S_{KM}(t|\{\,(\mathbf{x}_i,t_i,\delta_i)\,\}_{i=1}^n)$ decreases monotonically in a piecewise constant manner. It is important to note that the KM estimator effectively estimates the expected value of the survival function over the population $\mathcal{D}$, specifically $\mathbb{E}_{(\mathbf{x}, t, \delta) \sim \mathcal{D}} [P(T > t|\mathbf{x})]$, which reflects the probability that an event has not occurred by time $t$ of the population $\mathcal{D}$. In our GRADUATE model, we used the KM estimator to provide reference ground-truth survival curves for both subpopulations and the entire population.

\subsection{Deep Recurrent Survival Analysis (DRSA)}

Because our GRADUATE model and all the baseline systems in our empirical comparisons utilize the DRSA model~\citep{ren2019deep}, we provide some brief but salient details about DRSA to elucidate its workings.
DRSA discretizes a time horizon into $\tau$ intervals $\{1,\ldots,\tau\}$, and models the sequential patterns present in survival analysis with a recurrent neural network (RNN). Specifically, it uses a linear chain of long short-term memory (LSTM) units~\citep{hochreiter1997lstm} to build its RNN, with one LSTM unit for each timestep. At timestep $t$, individual $i$'s features $\mathbf{x}_i$ is concatenated with $t$ to form a vector, which is then fed into the corresponding LSTM unit as input. The LSTM unit at timestep $t$ outputs the instantaneous hazard rate \linebreak $h_{\theta,t}(\mathbf{x}_i) = P(T=t| T\!>\!t\!-\!1, \mathbf{x}_i,\phi^\theta)$, given the parameters $\theta$ of the LSTM unit of DRSA model $\phi$, features $\mathbf{x}_i$, and the fact that the event  has not occurred at previous timesteps. 
The survival function can then be estimated as \linebreak $S(t|\mathbf{x}_i)=\prod_{l:l\leq t} (1-h_{\theta,l}(\mathbf{x}_i))$.
Note that our choice of using DRSA as a base model for our \linebreak GRADUATE system is due to its state-of-the-art discriminative performance.
To train the parameters of the DRSA model, the following loss function $\mathcal{L}_{DRSA}$ is minimized over a training set $\{\,(\mathbf{x}_j,t_j,\delta_j)\,\}_{j=1}^{N_1}$ where $(\mathbf{x}_j,t_j,\delta_j) \sim \mathfrak{D}_1$. 
In doing so, we simultaneously minimize the negative log-likelihoods that (a) an event occurs at its observed time ($T\!=\!t_i$) over the \emph{uncensored} data (first term on the second line), (b) the true event time occurs \emph{before or at} the maximum time $\tau$ ($T\!\leq\!\tau$) over the \emph{uncensored} data (second term on the second line), and (c) the event occurs \emph{after} the censoring time ($T\!>\!t_i$) over the \emph{censored} data (first term on the third line).
\begin{align}
    &\mathbb{E}_{(\mathbf{x},t,\delta) \sim \mathfrak{D}_1} \left[\mathcal{L}_{DRSA}(\phi^\theta,\,\mathbf{x},t,\delta) \right] \nonumber\\
    &= \mathbb{E}_{(\mathbf{x},t,\delta) \sim \mathfrak{D}_1} \left[
    -\mathbb{I}\{\delta=1\}\log P(T=t|\mathbf{x},\phi^\theta)
    -\mathbb{I}\{\delta=1\}\log P(T\leq\tau|\mathbf{x},\phi^\theta) \right. \nonumber \\
    &\quad\quad\quad\quad\quad \left.-\mathbb{I}\{\delta=0\}\log P(T>t|\mathbf{x},\phi^\theta) \right] \nonumber \\ 
    &=  \mathbb{E}_{(\mathbf{x},t,\delta) \sim \mathfrak{D}_1} \left[-\mathbb{I}\{\delta=1\}\left(\log h_{\theta,t}(\mathbf{x})+\sum_{t^\prime| t^\prime<t}\log(1-h_{\theta,t^\prime}(\mathbf{x})) \right) \right.\nonumber \\
     & \quad\quad\quad\quad \left.-\mathbb{I}\{\delta=1\} \log \left( 1- \prod_{t^\prime|t^\prime< \tau} \left( 1-h_{\theta,t^\prime}(\mathbf{x})\right)\right) -\mathbb{I}\{\delta=0\} \sum_{t^\prime|t^\prime < t}\log \left(  1-h_{\theta,t^\prime}(\mathbf{x})\right) \right] \label{eq:drsa}
\end{align}

\section{Our GRADUATE Model} \label{sec:GRADUATE_model}

A predictive model is well-calibrated if its predicted probabilities agree with actual ground-truth probabilities. In the context of survival analysis, the expected number of event onsets under a  perfectly calibrated survival model at each discrete timestep should be the same as the ground-truth number of event onsets at the same timestep. (We discuss more about calibration in  survival analysis when we describe the ECE metric in the \emph{Evaluation Metrics} section.) To capture this intuition of good calibration, we marginalize, over the entire population $\mathfrak{D}_1$, the individualized survival function $S(t|\mathbf{x};\theta)$ that is produced by a survival model with parameters $\theta$, and require the resultant marginalized survival function to be the same as the ground-truth population-level survival function. More formally, we require
\begin{equation} \label{eq:model_1}
    \mathbb{E}_{(\mathbf{x},\cdot,\cdot)\sim \mathfrak{D}_1} \left[\, S(t|\mathbf{x};\theta) \,\right] = P(T > t) \;\;\; \forall t \in \{1, \ldots, \tau\},
\end{equation}
where $T$ is a random variable for the time of event onset, and $P(T > t)$ is the ground-truth population-level survival function. $P(T > t)$ can be approximated from finite $N_1$ samples $\{\,(\mathbf{x}_j,t_j,\delta_j)\,\}_{j=1}^{N_1}$ with the Kaplan-Meier (KM) estimator $S_{KM}(t|\{\,(\mathbf{x}_j,t_j,\delta_j)\,\}_{j=1}^n)$, a non-parametric model that gives an unbiased estimate of the survival function at the population level (see \emph{Background} section for more details).  Similarly, we impose this requirement on any subpopulation of interest $\mathfrak{D}_i \subset \mathfrak{D}_1$ as
\begin{equation} \label{eq:model_1_5}
\mathbb{E}_{(\mathbf{x},\cdot,\cdot)\sim \mathfrak{D}_i} \left[\, S(t|\mathbf{x};\theta) \,\right] = P_{\mathfrak{D}_i}(T > t) \;\;\; \text{ for any } t \in \{1, \ldots, \tau\} \text{ and } i \in \{2, \ldots, m\}
\end{equation}
and $P_{\mathfrak{D}_i}(T > t)$ is a survival function of subpopulation $\mathfrak{D}_i$.
In practical, we only have access to a limited number of individuals from the (sub)population $\mathfrak{D}_i$. To represent this finite set of samples, we employ a collection of tuples comprising random variables, denoted as $\{ (\mathbf{x}^i_{1,j},t^i_{1,j},\delta^i_{1,j})\}_{j=1}^{N_i}$ , where $N_i$ corresponds to the number of individuals available from (sub)population $\mathfrak{D}_i$ \footnote{The subscript $1$ is used for consistency with the notation in Section~\ref{sec:proof}, where a more general case is addressed.}. The estimation of $\mathbb{E}_{(\mathbf{x},\cdot,\cdot)\sim \mathfrak{D}_i} \left[\, S(t|\mathbf{x};\theta) \,\right]$ can be approximated using $\frac{1}{N_i} \sum_{j=1}^{N_i} S(t|\mathbf{x}^i_{1,j};\theta)$, while $P_{\mathfrak{D}_i}(T > t)$ is estimated using the Kaplan-Meier estimator $S_{KM}(t|\{ (\mathbf{x}^i_{1,j},t^i_{1,j},\delta^i_{1,j})\}_{j=1}^{N_i})$, and from Equation \ref{eq:model_1_5}, we have
\begin{equation} \label{eq:model_1_7.5}
\frac{1}{N_i} \sum_{j=1}^{N_i} S(t|\mathbf{x}^i_{1,j};\theta) = S_{KM}(t|\{ (\mathbf{x}^i_{1,j},t^i_{1,j},\delta^i_{1,j})\}_{j=1}^{N_i}) \;\;\; \text{ for any } t \in \{1, \ldots, \tau\} \text{ and } i \in \{1, \ldots, m\}.
\end{equation}
The (exact) equality constraints in Equation~\ref{eq:model_1_7.5} above are too strict, and we loosen it by introducing a ``slack" variable $c_i$, and using a function $dist(\cdot)$ that measures the ``distance'' \emph{over all timesteps} ($t \in \{1,\ldots,\tau \}$) between the marginalized survival function $\frac{1}{N_i} \sum_{j=1}^{N_i} S(t|\mathbf{x}^i_{1,j};\theta)$ that is produced by a survival model with parameters $\theta$ (left-hand side of the equality), and the ground-truth survival function given by the Kaplan-Meier estimator $S_{KM}(t|\{ (\mathbf{x}^i_{1,j},t^i_{1,j},\delta^i_{1,j})\}_{j=1}^{N_i})$ (right-hand side), i.e.,
\begin{equation}
dist\left( \Big\{ \frac{1}{N_i} \sum_{j=1}^{N_i} S(t^\prime|\mathbf{x}^i_{1,j};\theta) \Big\}_{t^\prime=1}^\tau, \Big\{S_{KM}(t^\prime|\{ (\mathbf{x}^i_{1,j},t^i_{1,j},\delta^i_{1,j})\}_{j=1}^{N_i})\Big\}_{t^\prime=1}^\tau \right) \leq c_i,  
\end{equation}
(We will provide two definitions of $dist(\cdot)$ later.) for $i=1,\ldots,m$, where we define vectors $\Big\{ \frac{1}{N_i} \sum_{j=1}^{N_i} S(t^\prime|\mathbf{x}^i_{1,j};\theta) \Big\}_{t^\prime=1}^\tau \textrm{ and } \Big\{S_{KM}(t^\prime|\{ (\mathbf{x}^i_{1,j},t^i_{1,j},\delta^i_{1,j})\}_{j=1}^{N_i}) \Big\}_{t^\prime=1}^\tau$ as 
\begin{align*}
    &\Big\{ \frac{1}{N_i} \sum_{j=1}^{N_i} S(t^\prime|\mathbf{x}^i_{1,j};\theta) \Big\}_{t^\prime=1}^\tau=\left(\frac{1}{N_i} \sum_{j=1}^{N_i} S(1|\mathbf{x}^i_{1,j};\theta),\ldots,\frac{1}{N_i} \sum_{j=1}^{N_i} S(\tau|\mathbf{x}^i_{1,j};\theta)\right)  \\
&\Big\{S_{KM}(t^\prime|\{ (\mathbf{x}^i_{1,j},t^i_{1,j},\delta^i_{1,j})\}_{j=1}^{N_i}) \Big\}_{t^\prime=1}^\tau=\left(S_{KM}(1|\{ (\mathbf{x}^i_{1,j},t^i_{1,j},\delta^i_{1,j})\}_{j=1}^{N_i}),\ldots,S_{KM}(\tau|\{ (\mathbf{x}_{i,j},t_{i,j},\delta_{i,j})\}_{j=1}^{N_i}) \right).
\end{align*}

 To achieve good discriminative performance, our GRADUATE model chooses to optimize the loss function of DRSA (Equation~\ref{eq:drsa}) using finite samples from the distribution $\mathfrak{D}_1$, i.e., $\frac{1}{N_1}\sum_{j=1}^{N_1}\mathcal{L}_{DRSA}(\phi^\theta, \,\mathbf{x}^1_{1,j},t^1_{1,j},\delta^1_{1,j}) $. We combine this discriminative loss function with the aforementioned requirements, and pose them as the following constrained optimization problem. (Our model can work with any alternative discrete time survival model other than DRSA by simply using the alternative's loss function. We use DRSA because of its state-of-the-art discriminative performance.)
\begin{equation} \label{eq:model_3} 
\begin{aligned}
\min_{\theta} \quad & \frac{1}{N_1}\sum_{j=1}^{N_1}\mathcal{L}_{DRSA}(\phi^\theta, \,\mathbf{x}^1_{1,j},t^1_{1,j},\delta^1_{1,j})\\
\textrm{subject to} 
\quad & dist\left( \Big\{ \frac{1}{N_i} \sum_{j=1}^{N_i} S(t^\prime|\mathbf{x}^i_{1,j};\theta) \Big\}_{t^\prime=1}^\tau, \Big\{S_{KM}(t^\prime|\{ (\mathbf{x}^i_{1,j},t^i_{1,j},\delta^i_{1,j})\}_{j=1}^{N_i})\Big\}_{t^\prime=1}^\tau \right) \leq c_i 
\end{aligned}    
\end{equation}
for $i=1,\ldots,m$ where $\phi^\theta$ represents a model $\phi$ with parameters $\theta$, 
$\mathfrak{D}_1$ is the entire population, 
$\mathfrak{D}_i$ for $i>1$ is a subpopulation, i.e., $\mathfrak{D}_i \subset \mathfrak{D}_1$ for $i=2,\ldots,m$, and $S_{KM}(t|\{ (\mathbf{x}^i_{1,j},t^i_{1,j},\delta^i_{1,j})\}_{j=1}^{N_i})$ is the KM estimator of $P_{\mathfrak{D}_i}(T>t)$ using $\{ (\mathbf{x}^i_{1,j},t^i_{1,j},\delta^i_{1,j})\}_{j=1}^{N_i}$ , where $N_i$ is the number of individuals available from (sub)population $\mathfrak{D}_i$. 

The constraint in Problem~\ref{eq:model_3} intuitively requires the ``distance'' between $\Big\{ \frac{1}{N_i} \sum_{j=1}^{N_i} S(t^\prime|\mathbf{x}^i_{1,j};\theta) \Big\}_{t^\prime=1}^\tau$ and $\Big\{S_{KM}(t^\prime|\{ (\mathbf{x}^i_{1,j},t^i_{1,j},\delta^i_{1,j})\}_{j=1}^{N_i})\Big\}_{t^\prime=1}^\tau$ to be at most $c_i$.
If the loss function $\frac{1}{N_1}\sum_{j=1}^{N_1}\mathcal{L}_{DRSA}(\phi^\theta, \,\mathbf{x}^1_{1,j},t^1_{1,j},\delta^1_{1,j})$ in Problem~\ref{eq:model_3} is convex, the problem can be easily solved as a convex program~\citep{boyd2004convex}. However,  $\mathbb{E}_{(\mathbf{x},t,\delta)\sim \mathfrak{D}_1}\left[\mathcal{L}_{DRSA}(\phi^\theta, \,\mathbf{x},t,\delta) \right]$ is the loss function of a complex recurrent neural network and is non-convex in its parameters $\theta$. 

To simplify notation, define 
\[dist_i \defeq dist\left( \Big\{ \frac{1}{N_i} \sum_{j=1}^{N_i} S(t^\prime|\mathbf{x}^i_{1,j};\theta) \Big\}_{t^\prime=1}^\tau, \Big\{S_{KM}(t^\prime|\{ (\mathbf{x}^i_{1,j},t^i_{1,j},\delta^i_{1,j})\}_{j=1}^{N_i})\Big\}_{t^\prime=1}^\tau \right)\] 
to be the left-hand term in the constraints of Problem~\ref{eq:model_3} for (sub)population $i$. 
%
One way to solve Problem~\ref{eq:model_3} is by combining the loss function with the constraints through penalizing the violation of each constraint $i$ with a manually-specified fixed cost $\mu_i$, i.e., \linebreak
$\frac{1}{N_1}\sum_{j=1}^{N_1}\mathcal{L}_{DRSA}(\phi^\theta, \,\mathbf{x}^1_{1,j},t^1_{1,j},\delta^1_{1,j}) - \sum_{i=1}^{m}\mu_i (dist_i - c_i)$.
The resultant function is then minimized.

Though seemingly straightforward, this approach has several drawbacks.
First, manually selecting the $\mu_i$ costs becomes impractical when the number of constraints is large, as in our case where each constraint corresponds to a subpopulation (we have as many as 25 subpopulations in our experiments). Second, selecting an appropriate value for each $\mu_i$ require painstaking effort -- if the value is too large, the effect of the original loss function is diminished; if the value is too small, the constraints are not enforced. Third, the problem of selecting good values is exacerbated when the constraints are dependent on one another (as is likely when subpopulations overlap). For example, if two constraints are positively correlated,  setting the $\mu_i$'s to be the same for all $i$'s means that the correlated constraints receive more attention than the others. Fourth, it is not clear what semantics are imbued in each $\mu_i$ to allow for intuitive tuning.

This approach of manually selecting $\mu_i$ is actually used by two baselines in our experiments, viz., X-Cal~\citep{goldstein2020x} and RPS~\citep{kamran2021estimating}. These systems only have \emph{one} constraint each, which render the manual approach feasible for them. 

In contrast, our GRADUATE model automatically learns the cost  $\mu_i$ terms, as well as, the parameters $\theta$ of the loss function from data by using the primal-dual algorithm~\citep{beck2017first}. We first define the Lagrangian~\citep{boyd2004convex} corresponding to Problem~\ref{eq:model_3} as 
\begin{align} 
\Hat{L}(\phi^\theta,\mathbf{\mu}) = & \frac{1}{N_1}\sum_{j=1}^{N_1}\mathcal{L}_{DRSA}(\phi^\theta, \,\mathbf{x}^1_{1,j},t^1_{1,j},\delta^1_{1,j}) + \nonumber\\ 
&\sum_{i=1}^{m} \mu_i\left(dist\left(\Big\{ \frac{1}{N_i} \sum_{j=1}^{N_i} S(t^\prime|\mathbf{x}^i_{1,j};\theta) \Big\}_{t^\prime=1}^\tau, \Big\{S_{KM}(t^\prime|\{ (\mathbf{x}^i_{1,j},t^i_{1,j},\delta^i_{1,j})\}_{j=1}^{N_i})\Big\}_{t^\prime=1}^\tau\right) - c_i\right) \label{eq:bg_primal_dual_2}
\end{align}
where $\mu_1,\mu_2,\ldots,\mu_{m}$ are non-negative dual variables.

Next, we randomly initialize the parameters $\mu_1,\ldots \mu_{m}$. Then we iteratively minimize, using stochastic gradient descent, $\Hat{L}(\phi^\theta,\mathbf{\mu})$ over $\theta$ holding each $\mu_i$ fixed to its current value. After that, we update each $\mu_i$ by a penalty $p_i(\theta)$ that measures how much constraint $i$ is violated, i.e., \mbox{$p_i(\theta) = dist\left(\Big\{ \frac{1}{N_i} \sum_{j=1}^{N_i} S(t^\prime|\mathbf{x}^i_{1,j};\theta) \Big\}_{t^\prime=1}^\tau, \Big\{S_{KM}(t^\prime|\{ (\mathbf{x}^i_{1,j},t^i_{1,j},\delta^i_{1,j})\}_{j=1}^{N_i})\Big\}_{t^\prime=1}^\tau\right) - c_i$}\xspace. Note that when computing $p_i(\theta)$, we use the $\theta$ parameters that we have obtained from the previous step. In short, we solve $\max_{\mathbf{\mu} \in \mathbb{R}_+^{m}} \min_{\theta} \Hat{L}(\phi^\theta,\mathbf{\mu})$ by alternating between minimizing the Lagrangian with respect to $\theta$ for fixed $\mu_i$'s, and then updating the dual variables $\mu_i$'s using the resulting minimizing $\theta$.  This procedure is summarized in Algorithm \ref{alg:bg_primal_dual_1}.

\begin{algorithm}[H]
\caption{Primal-Dual Method}\label{alg:bg_primal_dual_1}
Randomly initialize $\mathbf{\mu}^{(0)} \defeq (\mu_1^{(0)},\ldots, \mu_m^{(0)})$\;
\For{$j=1,\ldots,T$}{
$\theta^{(j)}=\argmin_{\theta} \Hat{L}(\phi^\theta,\mathbf{\mu}^{(j-1)})$ \\\tcc{with stochastic gradient descent.}
\For{$i=1,\ldots,m$}{
$\mu_i^{(j)}=\mu_i^{(j-1)} + \eta\, p_i(\theta^{(j)}) $ 
\\\tcc{$\eta$ is a step-size hyperparameter.}
}
}
\end{algorithm}

Next we provide two definitions for \mbox{$dist\left(\Big\{ \frac{1}{N_i} \sum_{j=1}^{N_i} S(t^\prime|\mathbf{x}^i_{1,j};\theta) \Big\}_{t^\prime=1}^\tau, \Big\{S_{KM}(t^\prime|\{ (\mathbf{x}^i_{1,j},t^i_{1,j},\delta^i_{1,j})\}_{j=1}^{N_i})\Big\}_{t^\prime=1}^\tau\right)$}\xspace, each of which measures the ``distance'' between the marginalized survival function $\Big\{ \frac{1}{N_i} \sum_{j=1}^{N_i} S(t^\prime|\mathbf{x}^i_{1,j};\theta) \Big\}_{t^\prime=1}^\tau$
produced by a survival model with parameters $\theta$, and the ground-truth survival function given by the Kaplan-Meier estimator $\Big\{S_{KM}(t^\prime|\{ (\mathbf{x}^i_{1,j},t^i_{1,j},\delta^i_{1,j})\}_{j=1}^{N_i})\Big\}_{t^\prime=1}^\tau$.

\subsection{$L^2$-Norm} \label{subsec:L2_Norm}

We define the distance between the marginalization of individualized survival curves and the ground-truth (sub)population survival curve
as the $L^2$-norm of the difference between the two curves, i.e.,
\begin{align}
    &dist\left(\Big\{ \frac{1}{N_i} \sum_{j=1}^{N_i} S(t|\mathbf{x}^i_{1,j};\theta) \Big\}_{t=1}^\tau, \Big\{S_{KM}(t|\{ (\mathbf{x}^i_{1,j},t^i_{1,j},\delta^i_{1,j})\}_{j=1}^{N_i})\Big\}_{t=1}^\tau\right)
= \nonumber \\& \frac{1}{\tau}\sum_{t=1}^\tau\left(\frac{1}{N_i} \sum_{j=1}^{N_i} S(t|\mathbf{x}^i_{1,j};\theta) - S_{KM}(t|\{ (\mathbf{x}^i_{1,j},t^i_{1,j},\delta^i_{1,j})\}_{j=1}^{N_i}) \right)^2. \label{eq:l2_norm_def}
\end{align}

\subsection{Variance-Adjusted Norm}

In the $L^2$-norm above, we treat every discrete timestep equally. However, the survival function of the Kaplan-Meier estimator at later timesteps has fewer individuals at risk (i.e., individuals for whom the event of interest can occur). Therefore, one would expect the variance to be higher at the tail of the survival function. The variance for (sub)population $\mathfrak{D}_i$ according to Greenwood's formula~\citep{kleinbaum2004survival} is given by
\begin{equation*}
    V\hat{a}r\left[S_{KM}(t|\{ (\mathbf{x}^i_{1,j},t^i_{1,j},\delta^i_{1,j})\}_{j=1}^{N_i})\right]=\left(S_{KM}(t|\{ (\mathbf{x}^i_{1,j},t^i_{1,j},\delta^i_{1,j})\}_{j=1}^{N_i})\right)^2\sum_{s:\tau_s\leq t}\left(\frac{e_s}{k_s(k_s-e_s)} \right),
\end{equation*}
where, for the unique times of event onset $\{\tau_i\}^{N}_{i=1}$, $k_s$ is the number of individuals up to time $\tau_s$ among (sub)population $\mathfrak{D}_i$ for whom the event of interest has not occurred and whom have not been censored, and $e_s$ is the number of individuals experiencing event onset at time $\tau_s$ among (sub)population $\mathfrak{D}_i$. 
As another distance measure, we thus propose the following to take the changing variance into account.
\begin{align*}
&dist\left(\Big\{ \frac{1}{N_i} \sum_{j=1}^{N_i} S(t|\mathbf{x}^i_{1,j};\theta) \Big\}_{t=1}^\tau, \Big\{S_{KM}(t|\{ (\mathbf{x}^i_{1,j},t^i_{1,j},\delta^i_{1,j})\}_{j=1}^{N_i})\Big\}_{t=1}^\tau\right) =\\
 &\max_{t:1\leq t \leq \tau}\frac{\left|\frac{1}{N_i} \sum_{j=1}^{N_i} S(t|\mathbf{x}^i_{1,j};\theta)-S_{KM}(t|\{ (\mathbf{x}^i_{1,j},t^i_{1,j},\delta^i_{1,j})\}_{j=1}^{N_i})\right|}{\sqrt{V\hat{a}r\left(S_{KM}(t|\{ (\mathbf{x}^i_{1,j},t^i_{1,j},\delta^i_{1,j})\}_{j=1}^{N_i})\right)}}.
\end{align*}
Please note that to we use absolute value of the difference between predicted and actual survival functions in the numerator because the fraction can then be interpreted as Z score; $c_i$ can then be selected using this intuition e.g., letting $c_i=1.96$ for 95\% confidence interval.  

\section{Mathematical Proof}~\label{sec:proof}
In Section~\ref{sec:GRADUATE_model}, we employed Algorithm~\ref{alg:bg_primal_dual_1} to find a solution to the constrained optimization problem formulated in Problem~\ref{eq:model_3} (repeated here for convenience):
\begin{equation*} \tag{\ref{eq:model_3}}
\begin{aligned}
\min_{\theta} \quad & \frac{1}{N_1}\sum_{j=1}^{N_1}\mathcal{L}_{DRSA}(\phi^\theta, \,\mathbf{x}^1_{1,j},t^1_{1,j},\delta^1_{1,j})\\
\textrm{subject to} 
\quad & dist\left( \Big\{ \frac{1}{N_i} \sum_{j=1}^{N_i} S(t^\prime|\mathbf{x}^i_{1,j};\theta) \Big\}_{t^\prime=1}^\tau, \Big\{S_{KM}(t^\prime|\{ (\mathbf{x}^i_{1,j},t^i_{1,j},\delta^i_{1,j})\}_{j=1}^{N_i})\Big\}_{t^\prime=1}^\tau \right) \leq c_i  \text{, $i=1,\ldots,m$.}
\end{aligned}    
\end{equation*}
To achieve a practical solution, we leverage two key strategies in the formulation of Problem~\ref{eq:model_3}:
\begin{enumerate}
    \item Parameterized Models: We employ a class of parameterized models, denoted $\mathcal{P}$ in which $\phi^\theta \in \mathcal{P}$, instead of a broader class, $\mathcal{H}$ that encompasses $\phi$.
    \item Monte Carlo Estimates: We utilize Monte Carlo estimates of expectation values rather than relying on theoretical expectations.
\end{enumerate}
Without these strategies, the formulation of Problem \ref{eq:model_3} transforms into Problem \ref{eq:model_3_PrimalOpt} as shown next:
\begin{equation} \label{eq:model_3_PrimalOpt}
\begin{aligned}
\min_{\phi \in \mathcal{H}} \quad & \mathbb{E}_{(\mathbf{x},t,\delta)\sim \mathfrak{D}_1}\left[\LDRSA\right] \\ 
\textrm{s.t.} \quad & \mathbb{E}_{\{(\mathbf{x}_{j},t_{j},\delta_{j})\}_{j=1}^{N_i}\sim \mathcal{D}_i}\left[dist\left( \Big\{ \frac{1}{N_i} \sum_{j=1}^{N_i} S(t^\prime|\mathbf{x}_{j}) \Big\}_{t^\prime=1}^\tau, \Big\{S_{KM}(t^\prime|\{ (\mathbf{x}_{j},t_{j},\delta_{j})\}_{j=1}^{N_i})\Big\}_{t^\prime=1}^\tau \right)\right] \leq c_i  \\
& \text{, $i=1,\ldots,m$.}
\end{aligned}   
\end{equation}

In this section, we demonstrate that Algorithm~\ref{alg:bg_primal_dual_1} acts as a learner for Problem~\ref{eq:model_3_PrimalOpt}, achieving a \textit{good} solution according to the framework of Probably Approximately Correct Constrained (PACC) learning, formally defined in Definition~\ref{definition:chamon_thesis_3}. We specifically consider the case where the $L^ 
2$-norm (Section~\ref{subsec:L2_Norm}) is used as the distance measure $dist$ in Problem~\ref{eq:model_3_PrimalOpt} to quantify the discrepancy between the marginalized individualized survival curves and the Kaplan-Meier survival curve of a (sub)population.
 
In particular, by applying the $L^2$-norm for $dist$ as shown in Equation \ref{eq:l2_norm_def}, Problem \ref{eq:model_3_PrimalOpt} becomes
\begin{align} \label{eq:P_CSO}
\min_{\phi \in \mathcal{H}} \quad & \mathbb{E}_{(\mathbf{x},t,\delta)\sim \mathfrak{D}_1}\left[\LDRSA\right] \nonumber\\ 
\textrm{s.t.} \quad & \frac{1}{\tau}\sum_{t=1}^\tau\mathbb{E}_{\{(\mathbf{x}_{j},t_{j},\delta_{j})\}_{j=1}^{N_i}\sim \mathcal{D}_i}\left[\left(\frac{1}{N_i}\sum_{j=1}^{N_i}S(t|\mathbf{x}_j)-\SKM\right)^2\right] \leq c_i, \nonumber\\
& \text{, $i=1,\ldots,m$.}\tag{\texttt{Primal-Opt}}
\end{align}
\ref{eq:P_CSO} is the key constrained optimization problem we aim to solve.

This section shows that solving \ref{eq:Hat_D_CSL} (or equivalently, applying Algorithm~\ref{alg:bg_primal_dual_1}) leads to an approximate solution of \ref{eq:P_CSO}. Theorem~\ref{thm:chamon_thesis_3} characterizes the difference between the exact solution (\ref{eq:P_CSO}) and the approximate solution obtained from \ref{eq:Hat_D_CSL} (or Algorithm~\ref{alg:bg_primal_dual_1}). This suggests that with high probability, solving \ref{eq:Hat_D_CSL} (or following Algorithm~\ref{alg:bg_primal_dual_1}) can result in a solution that is both approximately optimal and feasible within the constraints imposed in \ref{eq:P_CSO}.
\begin{equation} \label{eq:Hat_D_CSL}
\Hat{D}^*_\theta=\max_{\mathbf{\mu} \in \mathbb{R}_+^{m}} \min_{\theta \in \mathbb{R}^p} \Hat{L}(\phi^\theta,\mathbf{\mu}) \tag{\texttt{Empirical-Dual-Opt$_\theta$}} 
\end{equation}
\hspace{3em} where
\begin{align}
    &\hspace{3em} \Hat{L}(\phi^\theta,\mathbf{\mu})=\frac{1}{N_1}\sum_{j=1}^{N_1}\mathcal{L}_{DRSA}(\phi^\theta,\,\mathbf{x}_{1,j}^{1},t_{1,j}^{1},\delta_{1,j}^{1}) \nonumber \\
    &\hspace{3em} +\sum_{i=1}^m \mu_i\left[ \frac{1}{\tau}\sum_{t=1}^\tau\left(\frac{1}{\mathcal{N}_i}\sum_{n_i=1}^{\mathcal{N}_i}\left(\frac{1}{N_i}\sum_{j=1}^{N_i}S(t|\mathbf{x}_{n_i,j}^{i};\theta)-\SKMProp\right)^2 \right)- c_i \right]. \label{eq:chamon_thesis_4.4}
\end{align}
$\Hat{L}(\phi^\theta,\mathbf{\mu})$ in Equation~\ref{eq:chamon_thesis_4.4} is more general than the one in Equation~\ref{eq:bg_primal_dual_2}. In Equation~\ref{eq:chamon_thesis_4.4}, we assume we have multiple sets of samples available for each subpopulation. The number of sets of samples for subpopulation $i$ is denoted by $\mathcal{N}_i$. Each set contains $N_i$ samples (or individuals). In contrast, Equation~\ref{eq:bg_primal_dual_2} only assumes a single set of samples is available for each subpopulation. We use $\mathbf{x}_{n_i,j}^{i}$ to denote the $j$\textsuperscript{th} individual from the $n_i$\textsuperscript{th} set of samples in subpopulation $i$.
\subsection{Notation}
In this subsection, we introduce and explain the notations that are used throughout this section.
\begin{enumerate}[label=\arabic*., listparindent=1.5em]
    \item $\{ A_t\}_{t=1}^\tau$ represents $\tau$-dimensional vector $(A_1, \ldots,\ A_\tau)$.
    \item $\mathcal{H}$ is a hypothesis class containing functions $\phi:\mathcal{X} \rightarrow \mathbb{R}^\tau$ and $\phi(\mathbf{x}) = \{S(t|\mathbf{x})\}_{t=1}^\tau$. $S(t\mid \mathbf{x})$ represents the survival probability at time $t$ of an individual with features $\mathbf{x} \in \mathcal{X}$ or $P(T > t|\mathbf{x}, \phi)$. However, the generality of $\mathcal{H}$ makes it challenging to work with practically. We can narrow our focus to a subset called $\mathcal{P} \subset \mathcal{H}$. This subset comprises parameterized models, such as recurrent neural networks, where the parameters, weights and biases, are denoted as $\theta$. To denote a function within $\mathcal{P}$, we append the superscript $\theta$ to $\phi$, i.e., $\mathcal{P}$ consists of functions $\phi^\theta:\mathcal{X} \rightarrow \mathbb{R}^\tau$, with $\phi^\theta(\mathbf{x}) = \{S(t|\mathbf{x};\theta)\}_{t=1}^\tau$, wherein survival probabilities are modeled using a function $\phi^\theta$ parameterized by $\theta \in \mathbb{R}^p$.
    \item $h_{\theta,t}(\mathbf{x})$ represents the instantaneous hazard rate, a term used to model the death of the individual with features represented by a vector $\mathbf{x}$ at time $t$ given that he survives the earlier timestep, and the model is parameterized by $\theta$. Note that $h_{\theta,t}(\mathbf{x})=P(T=t|T>t-1,\mathbf{x},\phi^\theta)$, and having instantaneous hazard rates $\{h_{\theta,t}(\mathbf{x})\}_{t=1}^\tau$ is equivalent to having survival probabilities $\{S(t|\mathbf{x};\theta)\}_{t=1}^\tau$, i.e., $S(t\mid \mathbf{x};\theta) = \prod_{t^\prime; t^\prime \leq t} (1 - h_{\theta,t^\prime}(\mathbf{x}))$.
    \item For each (sub)population $i=1,\ldots,m$, let $\{(\mathbf{x}_{j},t_j,\delta_j)\}_{j=1}^{N_i} \sim \mathcal{D}_i$ with $\mathbf{x}_{j} \in \mathcal{X}$. The density function of $\Samplexy$ is denoted by $f_{i}\left(\Samplexy \right)$. Note that $\mathcal{D}_i$ represents a distribution of a vector of $N_i$ tuples of random variables $(\mathbf{x}_j,t_j,\delta_j)$ for $j=1,\ldots, N_i$ which are independent and identically distributed with distribution $\mathfrak{D}_i$. 
    \item $\mathfrak{D}_1$ represents entire population, and is the distribution of $\left(\mathbf{x},t,\delta\right)$ over the entire population, and the density function of $(\mathbf{x},t,\delta)$ is denoted by $f_{1}(\mathbf{x},t,\delta)$.
    \item From $\int_{\mathcal{Z}} F_1(\mathbf{x},t,\delta)f_1(\mathbf{x},t,\delta) d(\mathbf{x},t,\delta)$, a differential element $d(\mathbf{x},t,\delta)$ indicates that the integration is performed with respect to all three variables $\mathbf{x},\; t, \text{ and } \delta$. \label{notation:differential}
    \item Similar to \ref{notation:differential}, the differential element $d\left(\Samplexy[N_i]\right)$ in the integral $\int_{\mathcal{Z}} F_2\left(\Samplexy[N_i]\right) f_{i}\left(\Samplexy[N_i]\right)d\left(\Samplexy[N_i]\right)$ indicates that the integration is performed with respect to all variables $\mathbf{x}_1,\ldots,\mathbf{x}_{N_i}, t_1,\ldots,t_{N_i},\delta_1,\ldots,\delta_{N_i} $, when $F_2\left(\Samplexy[N_i]\right)$ is any arbitrary function of $\mathbf{x}_1,\ldots,\mathbf{x}_{N_i}, t_1,\ldots,t_{N_i},\delta_1,\ldots,\delta_{N_i} $.
\end{enumerate}
\subsection{Mathematical Proof of Theorem \ref{thm:chamon_thesis_3}} 
Before proceeding with the proofs for Theorem \ref{thm:chamon_thesis_3}, let us again present the constrained optimization problem of interest, \ref{eq:P_CSO}.
\begin{align*}
P^*=\min_{\phi \in \mathcal{H}} \quad & \mathbb{E}_{(\mathbf{x},t,\delta)\sim \mathfrak{D}_1}\left[\LDRSA\right] \nonumber\\ 
\textrm{s.t.} \quad & \frac{1}{\tau}\sum_{t=1}^\tau\mathbb{E}_{\{(\mathbf{x}_{j},t_{j},\delta_{j})\}_{j=1}^{N_i}\sim \mathcal{D}_i}\left[\left(\frac{1}{N_i}\sum_{j=1}^{N_i}S(t|\mathbf{x}_j)-\SKM\right)^2\right] \leq c_i, \tag{\ref{eq:P_CSO}}
\end{align*}
where
\begin{align}
\label{eq:drsa_before_expectation}
    \LDRSA=
     &-\mathbb{I}\{\delta=1\}\log P(T=t|\mathbf{x},\phi)
    -\mathbb{I}\{\delta=1\}\log P(T\leq\tau|\mathbf{x},\phi) \\ \nonumber
    &\, -\mathbb{I}\{\delta=0\}\log P(T>t|\mathbf{x},\phi)
\end{align}
for $i=1,\ldots,m$. Since $\phi(\mathbf{x})\defeq \{S(t|\mathbf{x})\}_{t=1}^\tau$ and instantaneous hazard rates $\{h_t(\mathbf{x})\}_{t=1}^\tau$ is equivalent to survival probabilities $\{S(t|\mathbf{x})\}_{t=1}^\tau$ (having one you can derive another), we will express $\mathcal{L}_{DRSA}$ in terms of $\{h_t(\mathbf{x})\}_{t=1}^\tau$ to show that $\mathcal{L}_{DRSA}$ is a function of $\phi$ and $(\mathbf{x},t,\delta)$.
\begin{align}
    \LDRSA=
     &-\mathbb{I}\{\delta=1\}\log P(T=t|\mathbf{x},\phi)
    -\mathbb{I}\{\delta=1\}\log P(T\leq\tau|\mathbf{x},\phi) \nonumber \\ 
    &\, -\mathbb{I}\{\delta=0\}\log P(T>t|\mathbf{x},\phi) \nonumber \\
    = & -\mathbb{I}\{\delta=1\}\left(\log h_{t}(\mathbf{x})+\sum_{t^\prime| t^\prime<t}\log(1-h_{t^\prime}(\mathbf{x})) \right) \nonumber \\
     &\; -\mathbb{I}\{\delta=1\} \log \left( 1- \prod_{t^\prime|t^\prime< \tau} \left( 1-h_{t^\prime}(\mathbf{x})\right)\right) \nonumber \\
    &\; -\mathbb{I}\{\delta=0\} \sum_{t^\prime|t^\prime < t}\log \left(  1-h_{t^\prime}(\mathbf{x})\right) \nonumber
\end{align}
We are interested in finding the solution $P^*$ as stated in \ref{eq:P_CSO}. The dual problem of \ref{eq:P_CSO} can be formulated in \ref{eq:D_CSL}. 
Later in Proposition \ref{Proposition1}, we demonstrate that the solution to the primal problem \ref{eq:P_CSO} is equal to that of \ref{eq:D_CSL} due to strong duality, i.e., $P^* = D^*$.  

\begin{equation} \label{eq:D_CSL}
D^*=\max_{\mathbf{\mu} \in \mathbb{R}_+^{m}} \min_{\phi \in \mathcal{H}} L(\phi,\mathbf{\mu}) \tag{\texttt{Dual-Opt}} 
\end{equation}
where $\phi(\mathbf{x})=\{S(t|\mathbf{x})\}_{t=1}^\tau$ and
\begin{align}
    \label{eq:Lagrangian}
    L(\phi,\mu) \defeq &\mathbb{E}_{(\mathbf{x},t,\delta)\sim \mathfrak{D}_1}\left[\LDRSA\right] \nonumber \\
    &+ \sum_{i=1}^m \mu_i \left(\frac{1}{\tau}\sum_{t=1}^\tau \mathbb{E}_{\{(\mathbf{x}_{j},t_{j},\delta_{j})\}_{j=1}^{N_i}\sim \mathcal{D}_i}\left[\left(\frac{1}{N_i}\sum_{j=1}^{N_i}S(t|\mathbf{x}_j)-\SKM\right)^2\right] -c_i \right).
\end{align}
However, both \ref{eq:P_CSO} and \ref{eq:D_CSL} involve optimization with respect to a function $\phi$ from a hypothesis class $\mathcal{H}$ (as stated in Assumption \ref{assumption_5}), which is too general to be practical. Therefore, we can consider a parameterized class $\mathcal{P}$ that is a subset of $\mathcal{H}$. As a result, the dual problem \ref{eq:D_CSL} becomes \ref{eq:Dtheta_CSL}. Later in Proposition \ref{Proposition2}, it can be demonstrated that the solution of \ref{eq:Dtheta_CSL} or $D^*_\theta$ is feasible for \ref{eq:P_CSO}, and the difference between $D^*_\theta$ and $P^*$ can be bounded by the approximation quality of functions in $\mathcal{P}$ compared to those in the more general class $\mathcal{H}$.
\begin{align} \label{eq:Dtheta_CSL}
    D^*_\theta=\max_{\mu \in \mathbb{R}_+^m} \min_{\theta \in \mathbb{R}^p} L(\phi^\theta,\mu)
    \tag{\texttt{Dual-Opt$_\theta$}}
\end{align}
where $\phi^\theta(x)=\{S(t|x;\theta)\}_{t=1}^\tau$ and $L$ is the Lagrangian in Equation \ref{eq:Lagrangian}.

In addition to approximating the function $\phi$ in $\mathcal{H}$ through parametrization, practical implementation also requires approximating the expectation by taking Monte Carlo sampling from $\mathfrak{D}_1$ or $\mathcal{D}_i$ for $i \in \{1,\ldots,m\}$. Therefore, the primal problem given by \ref{eq:P_CSO} can be reformulated as the problem stated in \ref{eq:PIV}, as shown below.
\begin{align} \label{eq:PIV}
\Hat{P}^*_\theta=&\min_{\theta} \quad  \frac{1}{N_1}\sum_{j=1}^{N_1}\mathcal{L}_{DRSA}(\phi^\theta,\,\mathbf{x}_{1,j}^1,t_{1,j}^1,\delta_{1,j}^1) \nonumber\\
=&\min_{\{S(t|\mathbf{x};\theta)\}_{t=1}^\tau \in \mathcal{P}} \quad  \frac{1}{N_1}\sum_{j=1}^{N_1}\mathcal{L}_{DRSA}(\phi^\theta,\,\mathbf{x}_{1,j}^1,t_{1,j}^1,\delta_{1,j}^1) \nonumber\\
& \textrm{s.t.} \quad  \frac{1}{\tau}\sum_{t=1}^\tau\left(\frac{1}{\mathcal{N}_i}\sum_{n_i=1}^{\mathcal{N}_i}\left(\frac{1}{N_i}\sum_{j=1}^{N_i}S(t|\mathbf{x}_{n_i,j}^{i};\theta)-\SKMProp\right)^2 \right)\leq c_i \tag{\texttt{Empirical-Primal-Opt$_\theta$}}
\end{align}
The parameterized empirical Lagrangian of \ref{eq:P_CSO} is defined as 
\begin{align*}
    \Hat{L}(\phi^\theta,\mathbf{\mu})=&\frac{1}{N_1}\sum_{j=1}^{N_1}\mathcal{L}_{DRSA}(\phi^\theta,\,\mathbf{x}_{1,j}^{1},t_{1,j}^{1},\delta_{1,j}^{1}) \nonumber \\
    &+\sum_{i=1}^m \mu_i\left[ \frac{1}{\tau}\sum_{t=1}^\tau\left(\frac{1}{\mathcal{N}_i}\sum_{n_i=1}^{\mathcal{N}_i}\left(\frac{1}{N_i}\sum_{j=1}^{N_i}S(t|\mathbf{x}_{n_i,j}^{i};\theta)-\SKMProp\right)^2 \right)- c_i \right]. \tag{\ref{eq:chamon_thesis_4.4}}
\end{align*}
The empirical dual problem of \ref{eq:P_CSO} is 
\begin{equation*} 
\Hat{D}^*_\theta=\max_{\mathbf{\mu} \in \mathbb{R}_+^{m}} \min_{\theta \in \mathbb{R}^p} \Hat{L}(\phi^\theta,\mathbf{\mu}). \tag{\texttt{Empirical-Dual-Opt$_\theta$}} 
\end{equation*}

In Proposition \ref{Proposition3}, we demonstrate that the distance between the solution $\Hat{D}^*_\theta$ of \ref{eq:Hat_D_CSL} and the solution $D^*_\theta$ of \ref{eq:Dtheta_CSL} can be bounded. As previously mentioned, Proposition \ref{Proposition2} shows the distance between $D^\theta$ and $P^*$ is also bounded. By combining these two results in Proposition \ref{Proposition2} and \ref{Proposition3}, we derive the main result presented in Theorem \ref{thm:chamon_thesis_3}. In that theorem, we find a solution that, with high probability, is approximately optimal and feasible with respect to \ref{eq:P_CSO} (or PACC for Survival Analysis which is formally defined in Definition \ref{definition:chamon_thesis_3}).  
\\
\begin{definition}[PACC Learnability for Survival Analysis] \label{definition:chamon_thesis_3}
    A hypothesis class $\mathcal{H}$ is probably approximately correct constrained (PACC) learnable for survival analysis if for every $\epsilon, \Bar{\delta} \in (0,1)$ and every distribution $\mathcal{D}_i$, $i=1,\ldots,m$, a $\phi_\dagger \in \mathcal{H}$ can be obtained using $\mathcal{N}(\epsilon,\delta,m)$ samples from each $\mathcal{D}_i$, that is, with probably $1-\Bar{\delta}$,
    \begin{itemize}
        \item probably approximately optimal, i.e.,
        \[
        \Big | \mathbb{E}_{(\mathbf{x},t,\delta)\sim \mathfrak{D}_1}\left[\LDRSA[\phi_\dagger]\right] - P^* \Big | \leq \epsilon \textrm{, and}
        \]
        \item probably approximately feasible, i.e., 
        \[
            \constraintterma{S_{\dagger}(t|\mathbf{x}_j)} \leq c_i+\epsilon \textrm{, for all } i = 1,\ldots,m,
        \]
    \end{itemize}
    where 
    $\phi_\dagger(\mathbf{x})=\{S_{\dagger}(t|\mathbf{x})\}_{t=1}^\tau$.
\end{definition}

\begin{assumption} \label{assumption_3}
There exist parameters $\Tilde{\theta}$ and $\hat{\theta}$ within $\mathbb{R}^p$ such that the survival functions $S(t|\mathbf{x};\Tilde{\theta})$ and $S(t|\mathbf{x};\hat{\theta})$ are strictly feasible for \ref{eq:P_CSO} (with perturbed constraints) and \ref{eq:PIV} respectively. Specifically, for every $i=1,\ldots,m$,
\begin{align}
    \frac{1}{\tau}\sum_{t=1}^\tau \mathbb{E}_{\{(\mathbf{x}_{j},t_{j},\delta_{j})\}_{j=1}^{N_i}\sim \mathcal{D}_i}\left[\left(\frac{1}{N_i}\sum_{j=1}^{N_i}S(t|\mathbf{x}_{j};\Tilde{\theta})-\SKM\right)^2\right] \leq c_i - \xi_\tau \nu  -\Psi \\   
    \textrm{, and \quad}\frac{1}{\tau}\sum_{t=1}^\tau\left(\frac{1}{\mathcal{N}_i}\sum_{n_i=1}^{\mathcal{N}_i}\left(\frac{1}{N_i}\sum_{j=1}^{N_i}S(t|\mathbf{x}_{n_i,j}^{i};\Hat{\theta})-\SKMProp\right)^2 \right) \leq c_i-\Psi
\end{align}
with $\nu$ as in Assumption \ref{assumption_5}, $\xi_\tau \defeq \tau+1$, , and  $\Psi > 0$.
\end{assumption}

\begin{assumption} \label{assumption_4}
The range of probabilities $S(t|\mathbf{x})$ for $\mathbf{x} \in \mathcal{X}$ and $t \in \{1,\ldots,\tau\}$ which are utilized in the calculation of $\mathcal{L}_{DRSA}$ and the constraints $\left(\frac{1}{N_i}\sum_{j=1}^{N_i}S(t|\mathbf{x}_j)-\SKM\right)^2$ is a closed set denoted as $[\Tilde{\mathcal{A}},\Tilde{\mathcal{B}}]$, which is a proper subset of the interval $[0,1]$, i.e., $[\Tilde{\mathcal{A}},\Tilde{\mathcal{B}}] \subset [0,1]$. This means that $\Tilde{\mathcal{A}}$ can be arbitrary small but larger than $0$. Similarly, $\Tilde{\mathcal{B}}$ can be arbitrarily close to, but lower than, $1$.

As a result, $\mathcal{L}_{DRSA}$ and $\left(\frac{1}{N_i}\sum_{j=1}^{N_i}S(t|\mathbf{x}_j)-\SKM\right)^2$ are bounded, i.e., $\exists B\in \mathbb{R} \textrm{\;such that\;} 0< \mathcal{L}_{DRSA} \leq B$ and $0\leq \left(\frac{1}{N_i}\sum_{j=1}^{N_i}S(t|\mathbf{x}_j)-\SKM\right)^2 \leq B$. We refer to these terms, namely $\mathcal{L}_{DRSA}$ and $\left(\frac{1}{N_i}\sum_{j=1}^{N_i}S(t|\mathbf{x}_j)-\SKM\right)^2$, as being \emph{B-bounded}.
\end{assumption}

\begin{assumption} \label{assumption_5}
    The hypothesis class $\mathcal{H}$ is a convex closed functional space. The parameterized class $\mathcal{P}$, which is a subset of $\mathcal{H}$, can be learned in a Probably Approximately Correct (PAC) manner. Furthermore, there exists a positive real number $\nu$ such that for every function $\phi$ in $\mathcal{H}$, defined as $\{S(t|\mathbf{x})\}_{t=1}^\tau$, there is a corresponding $\phi^{\theta^\prime}(\mathbf{x})$ in $P$, defined as $\{S(t|\mathbf{x};\theta^\prime)\}_{t=1}^\tau$ for which the corresponding instantaneous hazard rate
\[
\lvert h_{\theta^\prime,t}(\mathbf{x})-h_{t}(\mathbf{x}) \rvert \leq \nu
\]
for $t=1,\ldots,\tau$. Note that the instantaneous hazard rate can be computed from the survival probability according to the formula $h_t(\mathbf{x})=1-\frac{S(t|\mathbf{x})}{\prod_{t^\prime; t^\prime < t} 1-h_{t^\prime}(\mathbf{x})}$. Meanwhile, $h_{1}(\mathbf{x})$ is defined as $1-S(1|\mathbf{x})$.
\end{assumption}

\begin{assumption}
    \label{assumption:non_atomic}
    For $i=1,\ldots,m$, let $\Samplexy[N_i] \sim \mathcal{D}_i$. It is assumed that features $\mathbf{x}$ consist of continuous variables. Therefore, the density function $f_{i}\left(\Samplexy[N_i]\right)$ has no point masses or Dirac deltas.  
\end{assumption}

\begin{assumption}
    \label{assumption:non_atomic_0}   
    Let $\left(\mathbf{x},t,\delta\right) \sim \mathfrak{D}_1$, and the density function of $(\mathbf{x},t,\delta)$ is denoted by $f_{0}(\mathbf{x},t,\delta)$. Consistent with Assumption \ref{assumption:non_atomic}, it is assumed that the features $\mathbf{x}$ consist of continuous variables. Therefore, the density function $f_{0}\left(\mathbf{x},t,\delta\right)$ does not have any point masses or Dirac deltas.
\end{assumption}

\begin{proposition} \label{Proposition1}
Under Assumptions \ref{assumption_3}, \ref{assumption:non_atomic} and \ref{assumption:non_atomic_0}, the dual problem \ref{eq:D_CSL} and the primal problem \ref{eq:P_CSO} exhibit strong duality, i.e., $P^*=D^*$.
\end{proposition}
\proof{\underline{Proof}:}
     The optimal value of dual problem \ref{eq:D_CSL} provides a lower bound on the optimal value of its primal counterpart \ref{eq:P_CSO}, i.e., $D^* \leq P^*$ \citep{boyd2004convex}.

    To conclude that $P^* = D^*$, we have to prove that $D^* \geq P^*$. This can be accomplished by demonstrating that, although \ref{eq:P_CSO} is a non-convex program, the ranges of its cost and constraints form a convex set under the assumptions stated in the proposition. Explicitly, we define the epigraph set for the cost-constraints as follows:
    \begin{align}
        \mathcal{C}=\{(s_0,\mathbf{s})\in\mathbb{R}^{m+1}|\,&\exists \phi \in \mathcal{H} \textrm{\, s.t. \,} 
        \mathbb{E}_{(\mathbf{x},t,\delta)\sim \mathfrak{D}_1}\left[\LDRSA\right] \leq s_0 \textrm{\, and \,} \nonumber \\
        &\frac{1}{\tau}\sum_{t=1}^\tau\mathbb{E}_{\{(\mathbf{x}_{j},t_{j},\delta_{j})\}_{j=1}^{N_i}\sim \mathcal{D}_i}\left[\left(\frac{1}{N_i}\sum_{j=1}^{N_i}S(t|\mathbf{x}_j)-\SKM \right)^2 \right]  \leq s_i \} \label{eq:chamon_thesis_A_24}
    \end{align}
    where the vector $\mathbf{s} \in \mathbb{R}^{m}$ collects the $s_i, i=1,\ldots,m$. Then, the following holds:
\begin{lemma}
\label{lemma:chamon_thesis_lemma8}
Under the assumptions of Proposition \ref{Proposition1}, the cost-constraints set $\mathcal{C}$ in Equation \ref{eq:chamon_thesis_A_24} is a non-empty convex set.
\end{lemma}

With Lemma \ref{lemma:chamon_thesis_lemma8} which will be proven later, we can utilize the result from convex geometry stated in \textit{Supporting Hyperplane Theorem} below to show $D^* \geq P^*$, thus establishing strong duality for \ref{eq:P_CSO}.

\noindent\textbf{Supporting Hyperplane Theorem (Proposition 1.5.1 \cite{bertsekas2009convex})} 
    \textit{Let $\mathcal{A} \subset \mathbb{R}^n$ be a nonempty convex set. If $\Tilde{x} \in \mathbb{R}^n$ is not in the interior of $\mathcal{A}$, then there exists a hyperplane passing through $\Tilde{x}$ such that $\mathcal{A}$ is in one of its closed halfspaces, i.e., there exists $p \in \mathbb{R}^n$ and $p \ne 0$ such that $p^T \Tilde{x} \leq p^T x$ for all $x \in \mathcal{A}$.} 

Observe from \ref{eq:P_CSO} that the point $(P^*,c)$, where $c \in \mathbb{R}^m$ collects the values of $c_i$, cannot lie in the interior of $\mathcal{C}$. Otherwise, there would exist an $\epsilon > 0$ such that $(P^*-\epsilon,c) \in \mathcal{C}$, which would contradict the optimality of $P^*$.

As a result, \textit{Supporting Hyperplane Theorem} guarantees the existence of a non-zero vector $(\mu_0,\mu) \in \mathbb{R}^{m+1}$ such that
\begin{equation} \label{eq:chamon_thesis_A_25}
    \mu_0 s_0 + \mu^T s \geq \mu_0 P^* + \mu^T c, \textrm{\quad for all\quad} (s_0,s) \in \mathcal{C}.
\end{equation}
It is worth noting that the hyperplanes defined in Equation \ref{eq:chamon_thesis_A_25} use the same notation as the dual problem \ref{eq:D_CSL}, which foreshadows the fact that they actually encompass the values of the Lagrangian (Equation \ref{eq:Lagrangian}).

Note from Equation \ref{eq:chamon_thesis_A_24} that $\mathcal{C}$ is unbounded above. In other words, if $(s_0,s) \in \mathcal{C}$, then $(s^\prime_0,s^\prime) \in \mathcal{C}$ for all $(s^\prime_0,s^\prime) \succeq (s_0,s)$. As a result, Equation \ref{eq:chamon_thesis_A_25} can only hold if $\mu_i \geq 0$ for all $i=0,\ldots,m$. Otherwise, there would exist a vector in $\mathcal{C}$ for which the left-hand side of Equation \ref{eq:chamon_thesis_A_25} evaluates to an arbitrarily number that is negative enough to violate \textit{Supporting Hyperplane Theorem}. We next show that $\mu_0 \neq 0$ by contradiction. This allows us to divide by $\mu_0$.

\begin{claim}
    \label{Claim:Proposition1_Claim1}
    $\mu_0 \neq 0$
\end{claim}
\proof{Proof of Claim \ref{Claim:Proposition1_Claim1}.}
Suppose $\mu_0=0$. Then Equation \ref{eq:chamon_thesis_A_25} simplifies to
\begin{equation}
\label{eq:chamon_thesis_A_26}
    \mu^T s \geq \mu^T c \iff \mu^T (s-c) \geq 0 \textrm{,\quad for all \quad} (s_0,s) \in \mathcal{C}. 
\end{equation}
However, this contradicts the existence of the strictly feasible point $\phi^\prime$. Specifically, for any non-zero $\mu$, there exists $(s_0,s) \in \mathcal{C}$ achieved by $\phi^\prime$ based on Assumption \ref{assumption_3}, such that $s_i^\prime < c_i$ for all $i$. This contradicts Equation \ref{eq:chamon_thesis_A_26}.

Since $\mu_0 \neq 0$, Equation \ref{eq:chamon_thesis_A_25} can be rewritten as
\[
    s_0+\Tilde{\mu}^T s \geq P^*+\Tilde{\mu}^T c \textrm{,\quad for all \quad} (s_0,s) \in \mathcal{C}, 
\]
where $\Tilde{\mu}=\mu/\mu_0$. From the definition of $\mathcal{C}$ in Equation \ref{eq:chamon_thesis_A_24}, this implies that
\begin{align}
\label{eq:chamon_thesis_A_27}
    &\mathbb{E}_{(\mathbf{x},t,\delta)\sim \mathfrak{D}_1}\left[\LDRSA\right] \nonumber \\
    &+ \sum_{i=1}^m \Tilde{\mu}_i \left(\frac{1}{\tau}\sum_{t=1}^\tau\mathbb{E}_{\{(\mathbf{x}_j,t_{j},\delta_{j})\}_{j=1}^{N_i}\sim \mathcal{D}_i}\left[\left(\frac{1}{N_i}\sum_{j=1}^{N_i}S(t|\mathbf{x}_j)-\SKM\right)^2\right] -c_i \right) \geq P^*,
\end{align}
for all $\phi \in \mathcal{H}$. 

The left-hand side of Equation \ref{eq:chamon_thesis_A_27} is the Langrangian (Equation \ref{eq:Lagrangian}). It therefore implies that $L(\phi,\Tilde{\mu}) \geq P^*$. This also holds for the minimum of $L(\phi,\Tilde{\mu})$, which implies that $D^* \geq P^*$. Thus, strong duality holds for \ref{eq:P_CSO}.
\endproof{\Halmos}
\begingroup
\def\thelemma{\ref{lemma:chamon_thesis_lemma8}}
\begin{lemma}
With Assumptions \ref{assumption:non_atomic} and \ref{assumption:non_atomic_0}, the cost-constraints set $\mathcal{C}$ in Equation \ref{eq:chamon_thesis_A_24} is a non-empty convex set.
\end{lemma}
\addtocounter{lemma}{-1}
\endgroup
\proof{\underline{Proof}:}
    Let $(s_0,\mathbf{s}),(s'_0,\mathbf{s'}) \in \mathcal{C}$ be achieved by $\phi,\phi' \in \mathcal{H}$, i.e., 
    \begin{equation} \label{eq:Lemma8_equation1}
        \mathbb{E}_{(\mathbf{x},t,\delta)\sim \mathfrak{D}_1}\left[\LDRSA \right]\leq s_0
        \text{ and }        \mathbb{E}_{(\mathbf{x},t,\delta)\sim \mathfrak{D}_1}\left[\LDRSA[\phi^\prime] \right]\leq s^\prime_0
    \end{equation}
    and for $i=1,\ldots,m$
    \begin{align} 
        \frac{1}{\tau}\sum_{t=1}^\tau \EDi\left[\left(\frac{1}{N_i}\sum_{j=1}^{N_i}S(t|\mathbf{x}_j)-\SKM[t]\right)^2 \right]  \leq s_i \label{eq:Lemma8_equation2a} \\
        \frac{1}{\tau}\sum_{t=1}^\tau \EDi\left[\left(\frac{1}{N_i}\sum_{j=1}^{N_i}S^\prime(t|\mathbf{x}_j)-\SKM[t]\right)^2 \right]  \leq s^\prime_i .\label{eq:Lemma8_equation2b}
    \end{align}
    
    Our goal is to construct $\phi_\lambda(\mathbf{x})\defeq\{S_\lambda(t|\mathbf{x})\}_{t=1}^\tau$ such that 
    \begin{equation} \label{eq:Lemma8_LDRSA}
        \mathbb{E}_{(\mathbf{x},t,\delta)\sim \mathfrak{D}_1}\left[\LDRSA[\phi_\lambda] \right]\leq \lambda s_0 + (1-\lambda) s'_0
    \end{equation}
    and for $i=1,\ldots,m$
    \begin{equation} \label{eq:chamon_thesis_A_29_1}
        \frac{1}{\tau}\sum_{t=1}^\tau \EDi\left[\left(\frac{1}{N_i}\sum_{j=1}^{N_i}S_\lambda(t|\mathbf{x}_j)-\SKM[t]\right)^2 \right]  \leq \lambda s_i + (1-\lambda) s'_i
    \end{equation}
    for all $0 \leq \lambda \leq 1$.
To proceed, construct the $2(m\tau+1) \times 1$ vector measure
    \begin{align} \label{eq:chamon_thesis_A_37}
    \mathfrak{q}(\mathcal{Z})=
        \begin{bmatrix}
            \int_{\mathcal{Z}}\LDRSA f_{0}(\mathbf{x},t,\delta)d(\mathbf{x},t,\delta)
            \\
            \int_{\mathcal{Z}}\LDRSA[\phi^\prime] f_{0}(\mathbf{x},t,\delta)d(\mathbf{x},t,\delta)
            \\
            \vdots
            \\
            \int_{\mathcal{Z}}\left(\frac{1}{N_i}\sum_{j=1}^{N_i}S(t|\mathbf{x}_j)-\SKM\right)^2 f_{i}\left(\Samplexy[N_i]\right)d\left(\Samplexy[N_i]\right) \\
            \int_{\mathcal{Z}}\left(\frac{1}{N_i}\sum_{j=1}^{N_i}S^\prime(t|\mathbf{x}_j)-\SKM\right)^2 f_{i}\left(\Samplexy[N_i]\right)d\left(\Samplexy[N_i]\right)
            \\
            \vdots
            \\
            \int_{\mathcal{Z}}\left(\frac{1}{N_m}\sum_{j=1}^{N_m}S(\tau|\mathbf{x}_j)-\SKM[\tau]\right)^2 f_{i}\left(\Samplexy[N_i]\right)d\left(\Samplexy[N_i]\right) \\
            \int_{\mathcal{Z}}\left(\frac{1}{N_m}\sum_{j=1}^{N_m}S^\prime(\tau|\mathbf{x}_j)-\SKM[\tau]\right)^2 f_{i}\left(\Samplexy[N_i]\right)d\left(\Samplexy[N_i]\right)
        \end{bmatrix}
    \end{align}
    and, since both $f_0\left(\mathbf{x},t,\delta\right)$ and $f_{i}\left(\Samplexy[N_i]\right)$ do not contain point masses or Dirac deltas, $\mathfrak{q}$ is non-atomic. 
    
    \noindent\textbf{Lyapunov's Convexity Theorem \cite{diestel1977vector}}
\textit{Let $\mathfrak{q}: \mathcal{B} \rightarrow \mathbb{R}^n$ be a finite dimensional vector measure over the measurable space $(\Omega,\mathcal{B})$. If $\mathfrak{q}$ is non-atomic, then its range is convex, i.e., the set $\{\mathfrak{q}(\mathcal{Z})|\mathcal{Z} \in \mathcal{B}\}$ is a convex set.}
    
    Hence, from \textit{Lyapunov's Convexity Theorem}, there exists a set $\mathcal{T}_{\lambda} \in \mathcal{B}$ such that
    \begin{equation} \label{eq:chamon_thesis_A_38}
        \mathfrak{q}(\mathcal{T}_{\lambda})=\lambda\mathfrak{q}(\Omega)  
        \textrm{\quad and \quad}
        \mathfrak{q}(\Omega \setminus \mathcal{T}_{\lambda})=(1-\lambda)\mathfrak{q}(\Omega)
        .
    \end{equation}
    From Equation \ref{eq:chamon_thesis_A_38}, we construct $\phi_{\lambda}$ as 
    \begin{equation} \label{eq:chamon_thesis_A_39}
        \phi_{\lambda}(\mathbf{x})=
        \begin{cases}
            \phi(\mathbf{x}) & \text{ for } \mathbf{x} \in \mathcal{T}_{\lambda} \\
            \phi^\prime(\mathbf{x}) & \text{ for } \mathbf{x} \in \Omega \setminus \mathcal{T}_{\lambda}.
        \end{cases}
    \end{equation}
    Since $\mathcal{H}$ is decomposable, we have $\phi_{\lambda} \in \mathcal{H}$. 
    
    By Lemma \ref{lemma:proposition1_lemma1}, $\phi_{\lambda}$ can also be shown to satisfy Equations \ref{eq:Lemma8_LDRSA} and \ref{eq:chamon_thesis_A_29_1}.

    First, we will show that $\phi_\lambda$ satisfies Equation \ref{eq:Lemma8_LDRSA} by using Equation \ref{eq:proposition1_lemma2_eq2} from Lemma~\ref{lemma:proposition1_lemma1}.
    \begin{align}
        \mathbb{E}_{(\mathbf{x},t,\delta)\sim \mathfrak{D}_1}\left[\LDRSA[\phi_{\lambda}] \right] = &\lambda \mathbb{E}_{(\mathbf{x},t,\delta) \sim \mathfrak{D}_1}\left[\LDRSA \right]  \nonumber \\
        & +(1-\lambda) \mathbb{E}_{(\mathbf{x},t,\delta )\sim \mathfrak{D}_1}\left[\LDRSA[\phi^\prime] \right] \qquad(\because \text{ Equation~} \ref{eq:proposition1_lemma2_eq2}) \nonumber\\
        \leq & \lambda s_0 + (1-\lambda) s^\prime_0 \nonumber
    \end{align}
        
    Next, we will show that $\phi_\lambda$ also satisfies Equation \ref{eq:chamon_thesis_A_29_1} by using Equation \ref{eq:proposition1_lemma2_eq1} from Lemma~\ref{lemma:proposition1_lemma1}.
    \begin{align}
        &\EDixy{\left[ \left(\frac{1}{N_i} \sum_{j=1}^{N_i} S_\lambda(t|\mathbf{x}_j)-\SKM[t]\right)^2\right]} \nonumber\\
        &=\lambda \constraintterma{S(t|\mathbf{x}_j)} \nonumber\\
        &\quad + (1-\lambda) \constraintterma{S^\prime(t|\mathbf{x}_j)} \quad(\because \text{ Equation~} \ref{eq:proposition1_lemma2_eq1}) \nonumber\\
        &\leq \lambda s_i + (1-\lambda) s^\prime_i \nonumber
    \end{align}
Therefore, there exists $\phi_\lambda \in \mathcal{H}$ such that Equations \ref{eq:Lemma8_LDRSA} and \ref{eq:chamon_thesis_A_29_1} hold for any $\lambda \in [0,1]$. This implies that the set $\mathcal{C}$ is convex.
\endproof{\Halmos}

\begin{lemma}
\label{lemma:proposition1_lemma1}
With $\phi_\lambda$ in Equation \ref{eq:chamon_thesis_A_39}, the following equations hold 
\begin{align}
&\EDixy{\left[ \left(\frac{1}{N_i} \sum_{j=1}^{N_i} S_\lambda(t|\mathbf{x}_j)-\SKM\right)^2\right]} \nonumber\\
&=\lambda \constraintterma{S(t|\mathbf{x}_j)} \nonumber\\
&\quad + (1-\lambda) \constraintterma{S^\prime(t|\mathbf{x}_j)} \label{eq:proposition1_lemma2_eq1}
\end{align}
for $i=1,\ldots,m$,
and 
\begin{align}
\mathbb{E}_{(\mathbf{x},t,\delta)\sim \mathfrak{D}_1}\left[\LDRSA[\phi_{\lambda}] \right] = &\lambda \mathbb{E}_{(\mathbf{x},t,\delta) \sim \mathfrak{D}_1}\left[\LDRSA \right]  \nonumber \\
& +(1-\lambda) \mathbb{E}_{(\mathbf{x},t,\delta) \sim \mathfrak{D}_1}\left[\LDRSA[\phi^\prime] \right] \label{eq:proposition1_lemma2_eq2}
\end{align}
\end{lemma}
\proof{\underline{Proof}:}
Let
\begin{align}
&C_i\defeq  \constraintterma{S_\lambda(t|\mathbf{x}_j)} \nonumber \\
&c_i\defeq
\lambda \constraintterma{S(t|\mathbf{x}_j)} \nonumber \\ 
&\quad\quad +(1-\lambda) \constraintterma{S^\prime(t|\mathbf{x}_j)}
\nonumber 
\end{align}
for $i=1,\ldots,m$, and 
\begin{align}
&D 
\defeq \mathbb{E}_{(\mathbf{x},t,\delta)\sim \mathfrak{D}_1}\left[\LDRSA[\phi_{\lambda}] \right] \nonumber\\
&d \defeq \lambda  \mathbb{E}_{(\mathbf{x},t,\delta)\sim \mathfrak{D}_1}\left[\LDRSA \right] + (1-\lambda)  \mathbb{E}_{(\mathbf{x},t,\delta)\sim \mathfrak{D}_1}\left[\LDRSA[\phi^\prime] \right]. \nonumber
\end{align}
To begin, we will establish Equation \ref{eq:proposition1_lemma2_eq1}.

For $i=1,\ldots,m$,
\begin{align}
    & C_i \defeq  \constraintterma{S_\lambda(t|\mathbf{x}_j)} \nonumber \\
    & \quad\; = \int_{\mathcal{T}_{\lambda}} \left( \frac{1}{N_i}\sum_{j=1}^{N_i}S_\lambda(t|\mathbf{x}_j)-\SKM[t]\right)^2 f_{i} \left( \Samplexy[N_i] \right) d\left(\Samplexy[N_i]\right) \nonumber \\
    &\quad\quad +\int_{\Omega \setminus\mathcal{T}_{\lambda}} \left( \frac{1}{N_i}\sum_{j=1}^{N_i}S_\lambda(t|\mathbf{x}_j)-\SKM[t]\right)^2 f_{i} \left( \Samplexy[N_i] \right)  d\left(\Samplexy[N_i]\right) \nonumber \\
    &\quad\; = \lambda \constraintterma{S(t|\mathbf{x}_j)} \nonumber \\ 
    &\quad\quad +(1-\lambda) \constraintterma{S^\prime(t|\mathbf{x}_j)}.
    \nonumber 
\end{align}
Equation \ref{eq:proposition1_lemma2_eq1} is then derived. Next, we will proceed with the derivation of Equation \ref{eq:proposition1_lemma2_eq2}.
\begin{align}
    D = &\mathbb{E}_{(\mathbf{x},t,\delta)\sim \mathfrak{D}_1}\left[ -\mathbb{I}\{\delta=1\}\log\left(P(T=t\mid \mathbf{x}, \phi_{\lambda})\right) -\mathbb{I}\{\delta=1\}\log\left(P(t \leq \tau\mid \mathbf{x}, \phi_{\lambda})\right)\right. \nonumber \\
&\quad\quad\quad\quad\quad \left.-\mathbb{I}\{\delta=0\}\log\left(P(T > t\mid \mathbf{x}, \phi_{\lambda})\right)\right] \nonumber \\
    =&-\int_{\mathcal{T}_{\lambda}} \mathbb{I}\{\delta=1\}\log\left(P(T=t\mid \mathbf{x}, \phi)\right) f_1(\mathbf{x},t,\delta)d(\mathbf{x},t,\delta)  \nonumber \\
    &-\int_{\Omega \setminus\mathcal{T}_{\lambda}} \mathbb{I}\{\delta=1\}\log\left(P(T=t\mid \mathbf{x}, \phi^\prime)\right) f_1(\mathbf{x},t,\delta)d(\mathbf{x},t,\delta) \nonumber \\
    &-\int_{\mathcal{T}_{\lambda}} \mathbb{I}\{\delta=1\}\log\left(P(T\leq \tau\mid \mathbf{x}, \phi)\right) f_1(\mathbf{x},t,\delta)d(\mathbf{x},t,\delta)  \nonumber \\
    &-\int_{\Omega \setminus\mathcal{T}_{\lambda}} \mathbb{I}\{\delta=1\}\log\left(P(T\leq \tau\mid \mathbf{x}, \phi^\prime)\right) f_1(\mathbf{x},t,\delta)d(\mathbf{x},t,\delta) \nonumber \\
    &-\int_{\mathcal{T}_{\lambda}} \mathbb{I}\{\delta=0\}\log\left(P(T>t\mid \mathbf{x}, \phi)\right) f_1(\mathbf{x},t,\delta)d(\mathbf{x},t,\delta)  \nonumber \\
    &-\int_{\Omega \setminus\mathcal{T}_{\lambda}} \mathbb{I}\{\delta=0\}\log\left(P(T>t\mid \mathbf{x}, \phi^\prime)\right) f_1(\mathbf{x},t,\delta)d(\mathbf{x},t,\delta) \nonumber \\
    = & \lambda\mathbb{E}_{(\mathbf{x},t,\delta)\sim \mathfrak{D}_1}\left[\LDRSA \right] + (1-\lambda)  \mathbb{E}_{(\mathbf{x},t,\delta)\sim \mathfrak{D}_1}\left[\LDRSA[\phi^\prime] \right] \label{eq:proposition1_eq1.5}
\end{align}
Equation \ref{eq:proposition1_lemma2_eq2} follows as a conclusion.
\endproof{\Halmos}
\begin{proposition} \label{Proposition2}
Let $\theta^*$ achieve the optimality in \ref{eq:Dtheta_CSL}. Under Assumptions \ref{assumption_3} - \ref{assumption:non_atomic_0}, $\phi^{\theta^*} \defeq \{S(t|\mathbf{x};\theta^*)\}_{t=1}^\tau$ is a feasible solution of \ref{eq:P_CSO} and 
\begin{align}
    \label{eq:chamon_thesis_4.12}
    P^* \leq D_\theta^* \leq P^*+(M_1+M_2 \tau+M_3 \tau+M_4\tau)\nu + \lVert \Tilde{\mathbf{\mu}}^* \rVert_{1}\; (\tau+1) \nu,
\end{align}
where $P^*$ and $D_\theta^*$ are the solutions from \ref{eq:P_CSO} and \ref{eq:Dtheta_CSL} respectively. $M_1, M_2, M_3, M_4$ are Lipschitz constants. $\mathbf{\Tilde{\mu}^*}$ are the dual variables of modified version of \ref{eq:P_CSO} where its constraints are tightened by $(\tau+1) \nu$ for $i=1,\ldots,m$.
\end{proposition}
\proof{\underline{Proof}:} We first demonstrate that $\phi^{\theta^*}$, defined as $\{S(t|\mathbf{x};\theta^*)\}_{t=1}^\tau$, is a feasible solution for \ref{eq:P_CSO}, and then show the optimality relationship between $D_\theta^*$ and $P^*$ in Equation \ref{eq:chamon_thesis_4.12}.
\paragraph{Feasibility.}
For the purpose of proving by contradiction, let us assume that $\{S(t|\mathbf{x};\theta^*)\}_{t=1}^\tau$ is not a feasible solution. Then there exists at least one index $i$ for which
\[
\frac{1}{\tau}\sum_{t=1}^\tau \EDi\left[\left(\frac{1}{N_i}\sum_{j=1}^{N_i}S(t|\mathbf{x}_j;\theta^*)-\SKM\right)^2\right] > c_i
.\]
Given that $\mathbf{\mu}$ has no upper limit, we can obtain that $D_{\theta}^*$ approaches infinity. However, according to Assumption \ref{assumption_3}, $D_\theta^*$ is finite, thereby obtaining a contradiction. Specifically, consider the dual function $d(\mathbf{\mu})$ where $D_\theta^*$ in \ref{eq:Dtheta_CSL} can be written as $D_\theta^*=\max_{\mathbf{\mu} \in \mathbb{R}^m_{+}} d(\mathbf{\mu})$. \begin{align*}
    d(\mathbf{\mu})&=\min_{\theta \in \mathbb{R}^p} L(\phi^{\theta},\mathbf{\mu}) \\
    &=\min_{\theta \in \mathbb{R}^p}\mathbb{E}_{(\mathbf{x},t,\delta)\sim \mathfrak{D}_1}\left[\LDRSA[\phi^\theta]\right] \\
    &\quad+ \sum_{i=1}^m \mu_i \left(\frac{1}{\tau}\sum_{t=1}^\tau\mathbb{E}_{\{(\mathbf{x}_{j},t_{j},\delta_{j})\}_{j=1}^{N_i}\sim \mathcal{D}_i}\left[\left(\frac{1}{N_i}\sum_{j=1}^{N_i}S(t|\mathbf{x}_j;\theta)-\SKM\right)^2\right] -c_i \right)
\end{align*}
Leveraging the fact that $\mathcal{L}_{DRSA}$ is $B$-bounded and a strictly feasible $\theta^\prime$ exists (per Assumption \ref{assumption_3}), $d(\mu)$ is upper bounded by
\begin{align}
    d(\mathbf{\mu}) & \leq \mathbb{E}_{(\mathbf{x},t,\delta)\sim \mathfrak{D}_1}\left[\LDRSA[\phi^{\theta^\prime}]\right] \nonumber \\
    &\quad+ \sum_{i=1}^m \mu_i \left(\frac{1}{\tau}\sum_{t=1}^\tau\mathbb{E}_{\{(\mathbf{x}_j,t_{j},\delta_{j})\}_{j=1}^{N_i}\sim \mathcal{D}_i}\left[\left(\frac{1}{N_i}\sum_{j=1}^{N_i}S(t|\mathbf{x}_j;\theta^\prime)-\SKM\right)^2\right] -c_i \right) \leq B,
    \label{eq:chamon_thesis_A_47}
\end{align}
where we use the fact that $\mu_i \geq 0$. By contradiction, it can be concluded that $\phi^{\theta^*} \defeq \{S(t|\mathbf{x};\theta^*)\}_{t=1}^\tau$ is feasible for \ref{eq:P_CSO}.

\paragraph{Close to Optimality.}
Recall that Proposition \ref{Proposition1} posits that strong duality holds for \ref{eq:P_CSO} and \ref{eq:D_CSL}, assuming the conditions stipulated in Proposition \ref{Proposition2} are met. Accordingly, the Lagrangian (Equation \ref{eq:Lagrangian}) adheres to the saddle-point relation described by
\begin{align}
    L(\phi^*,\mu^\prime) &\leq \max_{\mathbf{\mu} \in \mathbb{R}_+^{m}} \min_{\phi \in \mathcal{H}} L(\phi,\mathbf{\mu}) = D^*\nonumber \\
    &= P^*  \nonumber \\
    &= \min_{\phi \in \mathcal{H}} \max_{\mathbf{\mu} \in \mathbb{R}_+^{m}}L(\phi,\mathbf{\mu}) \hspace{170pt}
    \text{(Section 5.4 in \cite{boyd2004convex})}\nonumber\\ 
    &\leq L(\phi^\prime,\mathbf{\mu}^*)
    \label{eq:chamon_thesis_A_48}
\end{align}
for any $\phi^\prime \in \mathcal{H}$, $\mu^\prime \in \mathbb{R}^m_{+}$, and $\phi^*$ and $\mu^*$ represents a solution of \ref{eq:P_CSO} and \ref{eq:D_CSL} respectively.

Moreover, based on \ref{eq:Dtheta_CSL}, we have
\[
D_\theta^* \geq \min_{\theta \in \mathbb{R}^P} L(\phi^\theta,\mu)  \textrm{\quad for all $\mu \in \mathbb{R}_+^m$}.
\]
This immediately provides the lower bound presented in Equation \ref{eq:chamon_thesis_4.12}. Specifically,
\[
D_\theta^* \geq \min_{\theta \in \mathbb{R}^P} L(\phi^\theta,\mu) \geq \min_{\phi \in \mathbb{R}^P} L(\phi,\mu^*)=P^*,
\]
where the second inequality stems from $\mathcal{P}$ being a subset of $\mathcal{H}$, as stated in Assumption \ref{assumption_5}.

The upper bound is established by connecting the parameterized dual problem \ref{eq:Dtheta_CSL} with the modified version of the original problem \ref{eq:P_CSO} or \ref{eq:chamon_thesis_PXVIII}. This can be accomplished by adding and subtracting $L(\phi,\mu)$ in \ref{eq:Dtheta_CSL}, which results in
\begin{align}
D^*_\theta = &\max_{\mathbf{\mu} \in \mathbb{R}^m_{+}} \min_{\theta \in \mathbb{R}^b} L(\phi,\mathbf{\mu}) \nonumber \\
    &+\mathbb{E}_{(\mathbf{x},t,\delta)\sim \mathfrak{D}_1}\left[\LDRSA[\phi^{\theta}] -\LDRSA \right] 
    \nonumber \\
    &+\sum_{i=1}^m \mu_i\left(\frac{1}{\tau}\sum_{t=1}^\tau \left( 
    \constraintterma{S(t|\mathbf{x}_j;\theta)} \right.\right. \nonumber \\
    &\quad\quad\quad\quad\quad\quad- \left.\left.\quad\constraintterma{S(t|\mathbf{x}_j)} \right)\right)  \label{eq:chamon_thesis_A_51} \\
    \leq & \max_{\mathbf{\mu} \in \mathbb{R}^m_{+}} L(\phi,\mathbf{\mu}) +(M_1+M_2 \tau+M_3 \tau+M_4\tau)\nu + \sum_{i=1}^m \mu_i(\tau+1) \nu  \quad\quad\quad\quad\quad\quad (\because \textrm{ Lemma \ref{lemma:Proposition2_X_10} and \ref{lemma:Proposition2_X_9}} ) \nonumber \\
    \leq & \max_{\mathbf{\mu} \in \mathbb{R}^m_{+}}  L(\phi,\mathbf{\mu}) +(M_1+M_2 \tau+M_3 \tau+M_4\tau)\nu + \lVert \mu \rVert_{1}\; (\tau+1) \nu . \label{eq:chamon_thesis_A_53}
\end{align}
Since Equation \ref{eq:chamon_thesis_A_53} is valid for all $\phi \in \mathcal{H}$, it is true for the minimizer.
\begin{align}
    D_\theta^* \leq \min_{\phi \in \mathcal{H}} \max_{\mathbf{\mu} \in \mathbb{R}^m_+} L(\phi,\mathbf{\mu}) +(M_1+M_2 \tau+M_3 \tau+M_4\tau)\nu + \lVert \mu \rVert_{1}\; (\tau+1) \nu  \eqdef \Tilde{P}^* \label{eq:chamon_thesis_A_54}
\end{align}
Moreover, the right-hand side of Equation \ref{eq:chamon_thesis_A_54}, specifically $\Tilde{P}^*$, actually represents a modified form of \ref{eq:P_CSO}. Therefore, we deduce a new saddle-point relation, analogous to Equation \ref{eq:chamon_thesis_A_48}, that associates $\Tilde{P}^*$ and consequently $D_\theta^*$ with $P^*$.

Equation \ref{eq:chamon_thesis_A_54} can be rearranged as follows:
\begin{align}
    \Tilde{P}^* = &\min_{\phi \in \mathcal{H}} \max_{\mathbf{\mu} \in \mathbb{R}_{+}^{m}} \mathbb{E}_{(\mathbf{x},t,\delta)\sim \mathfrak{D}_1}\left[\LDRSA \right] +(M_1+M_2 \tau+M_3 \tau+M_4\tau)\nu \nonumber \\
    &+\sum_{i=1}^m \mu_i\left(\frac{1}{\tau}\sum_{t=1}^\tau
    \constraintterma{S(t|\mathbf{x}_j)} - c_i+ (\tau+1) \nu \right) .\label{eq:chamon_thesis_A_55}
\end{align}
This is where we identify the primal optimization problem.
\begin{align} 
\Tilde{P}^*=\min_{\phi \in \mathcal{H}} \quad & \mathbb{E}_{(\mathbf{x},t,\delta)\sim \mathfrak{D}_1}\left[\LDRSA\right]  +(M_1+M_2 \tau+M_3 \tau+M_4\tau)\nu \nonumber\\ 
\textrm{s.t.} \quad & \frac{1}{\tau}\sum_{t=1}^\tau\mathbb{E}_{\{(\mathbf{x}_{j},t_{j},\delta_{j})\}_{j=1}^{N_i}\sim \mathcal{D}_i}\left[\left(\frac{1}{N_i}\sum_{j=1}^{N_i}S(t|\mathbf{x}_j)-\SKM\right)^2\right] \leq c_i - (\tau+1) \nu \label{eq:chamon_thesis_PXVIII}  \tag{ \texttt{Disturbed-Primal}}\\
\textrm{for } \quad & i=1,\ldots,m. \nonumber 
\end{align}
With Proposition \ref{Proposition1}, \ref{eq:chamon_thesis_PXVIII} also exhibits strong duality, i.e.,
\begin{equation}
    \Tilde{P}^* = \min_{\phi \in \mathcal{H}} L(\phi,\Tilde{\mathbf{\mu}}^*)+(M_1+M_2 \tau+M_3 \tau+M_4\tau)\nu + \lVert \Tilde{\mathbf{\mu}}^* \rVert_{1}\; (\tau+1) \nu, \label{eq:chamon_thesis_A_56}
\end{equation}
where $\Tilde{\mathbf{\mu}}^*$ are the dual variables of \ref{eq:chamon_thesis_PXVIII}, i.e., the $\mathbf{\mu}$ that achieve
\begin{align}
    \Tilde{D}^* = &\max_{\mathbf{\mu} \in \mathbb{R}_{+}^{m}} \min_{\phi \in \mathcal{H}}  \mathbb{E}_{(x,t,\delta)\sim \mathfrak{D}_1}\left[\LDRSA \right] +(M_1+M_2 \tau+M_3 \tau+M_4\tau)\nu \nonumber \\
    &+\sum_{i=1}^m \mu_i\left(\frac{1}{\tau}\sum_{t=1}^\tau
    \constraintterma{S(t|\mathbf{x}_j)} - c_i+ (\tau+1) \nu \right). \label{eq:chamon_thesis_A_57}
\end{align}
Returning to Equation \ref{eq:chamon_thesis_A_54}, we are now in a position to finalize the proof. Initially, we employ Equation \ref{eq:chamon_thesis_A_56} to derive the following
\begin{equation}
    D^*_\theta \leq \Tilde{P}^* \leq L(\phi^*,\Tilde{\mathbf{\mu}}^*)+(M_1+M_2 \tau+M_3 \tau+M_4\tau)\nu + \lVert \Tilde{\mathbf{\mu}}^* \rVert_{1}\; (\tau+1) \nu.
    \label{eq:chamon_thesis_A_58}
\end{equation}
We have employed $\phi^*$, the solution of \ref{eq:P_CSO}, as a suboptimal solution in Equation \ref{eq:chamon_thesis_A_56}. The saddle point relationship given by Equation \ref{eq:chamon_thesis_A_48} implies that $L(\phi^*,\Tilde{\mathbf{\mu}}^) \leq P^*$, which allows us to establish the upper bound in Equation \ref{eq:chamon_thesis_4.12} or
\begin{align*}
    D_\theta^* \leq P^*+(M_1+M_2 \tau+M_3 \tau+M_4\tau)\nu + \lVert \Tilde{\mathbf{\mu}}^* \rVert_{1}\; (\tau+1) \nu.
\end{align*}
\endproof{\Halmos}
\begin{lemma}
\label{lemma:Proposition2_X_1}
With $\theta^\prime$ be as in Assumption \ref{assumption_5} where $\lvert h_t(\mathbf{x})-h_{\theta^\prime,t}(\mathbf{x}) \rvert \leq \nu
$ for $t=1,\ldots,\tau$,
    \begin{align}
        \label{eq:Proposition2_X_3}
        \Big|S(t_n|\mathbf{x};\theta^\prime) - S(t_n|\mathbf{x}) \Big|  \leq n\nu  
    \end{align}
where $t_n$ represents the n\textsuperscript{th} discrete time step within the time interval $[1, \tau]$.      
\end{lemma}

\proof{\underline{Proof}:}
We will prove by induction. For $n=1$, Equation \ref{eq:Proposition2_X_3} is true by Assumption \ref{assumption_5}.

Assuming $\Big|S(t_n|\mathbf{x}_j;\theta^\prime) - S(t_n|\mathbf{x}) \Big|  \leq n\nu  $, we will show $\Big|S(t_{n+1}|\mathbf{x};\theta^\prime) - S(t_{n+1}|\mathbf{x}) \Big|  \leq (n+1)\nu$.

We first rewrite the term on the left hand side of Equation \ref{eq:Proposition2_X_3} by using instantaneous hazard rate.
\begin{align}
    \Big|S(t_n|\mathbf{x};\theta^\prime) - S(t_n|\mathbf{x}) \Big| =
    \Big| \prod_{t; 1 \leq t \leq t_n}\left( 1-h_{\theta^\prime,t}(\mathbf{x})\right) - \prod_{t; 1 \leq t \leq t_n}\left( 1-h_{t}(\mathbf{x})\right) \Big|  \label{eq:Proposition2_X_3.5}
\end{align}
Therefore, from inductive assumption
\begin{align}
    \Big| \prod_{t; 1 \leq t \leq t_n}\left( 1-h_{\theta^\prime,t}\left(\mathbf{x}\right)\right) - \prod_{t; 1 \leq t \leq t_n}\left( 1-h_{t}(\mathbf{x})\right) \Big| \left(1-h_{\theta^\prime,t_{n+1}}(\mathbf{x})\right) &\leq n\nu\left(1-h_{\theta^\prime,t_{n+1}}(\mathbf{x})\right) \nonumber \\
    \Big| \prod_{t; 1 \leq t \leq t_{n+1}}\left( 1-h_{\theta^\prime,t}\left(\mathbf{x}\right)\right) - \left(1-h_{\theta^\prime,t_{n+1}}(\mathbf{x})\right) \prod_{t; 1 \leq t \leq t_n}\left( 1-h_{t}(\mathbf{x})\right) \Big| 
    &\leq n\nu\left(1-h_{\theta^\prime,t_{n+1}}(\mathbf{x})\right).
    \label{eq:Proposition2_X_4}
\end{align}
From Assumption \ref{assumption_5}, for $t = 1,\ldots,\tau$
\begin{align}
    \lvert h_{\theta,t}(\mathbf{x})-h_{t}(\mathbf{x})\rvert = \lvert \left(1-h_{\theta,t}(\mathbf{x})\right)-\left(1-h_{t}(\mathbf{x})\right)\rvert &\leq \nu \nonumber \\
    -\nu \leq  \left(1-h_{\theta,t}(\mathbf{x})\right)-\left(1-h_{t}(\mathbf{x})\right) &\leq \nu.
    \label{eq:Proposition2_X_5}
\end{align}
WLOG $\prod_{t; 1 \leq t \leq t_{n}}\left( 1-h_{\theta^\prime,t}\left(\mathbf{x}\right)\right) \geq \prod_{t; 1 \leq t \leq t_n}\left( 1-h_{t}(\mathbf{x})\right)$, we can then use inductive assumption to have
\begin{align}
    n\nu &\geq \prod_{t; 1 \leq t \leq t_{n}}\left( 1-h_{\theta^\prime,t}\left(\mathbf{x}\right)\right) - \prod_{t; 1 \leq t \leq t_n}\left( 1-h_{t}(\mathbf{x})\right) \geq 0 
    \nonumber \\
    n\nu\left(1-h_{\theta^\prime,t_{n+1}}(\mathbf{x})\right) &\geq \prod_{t; 1 \leq t \leq t_{n+1}}\left( 1-h_{\theta^\prime,t}\left(\mathbf{x}\right)\right) - \left(1-h_{\theta^\prime,t_{n+1}}(\mathbf{x})\right)\prod_{t; 1 \leq t \leq t_n}\left( 1-h_{t}(\mathbf{x})\right) \geq 0 
    \nonumber \\
    n\nu\left(1-h_{\theta^\prime,t_{n+1}}(\mathbf{x})\right) &\geq \prod_{t; 1 \leq t \leq t_{n+1}}\left( 1-h_{\theta^\prime,t}\left(\mathbf{x}\right)\right) - \left(1-h_{\theta^\prime,t_{n+1}}(\mathbf{x})\right)\prod_{t; 1 \leq t \leq t_n}\left( 1-h_{t}(\mathbf{x})\right) \nonumber \\
    &\quad + \left(\prod_{t; 1 \leq t \leq t_{n+1}}\left( 1-h_{t}(\mathbf{x})\right) - \prod_{t; 1 \leq t \leq t_{n+1}}\left( 1-h_{t}(\mathbf{x})\right) \right) \geq 0  \nonumber \\
    n\nu\left(1-h_{\theta^\prime,t_{n+1}}(\mathbf{x})\right) &\geq \prod_{t; 1 \leq t \leq t_{n+1}}\left( 1-h_{\theta^\prime,t}\left(\mathbf{x}\right)\right)  - \prod_{t; 1 \leq t \leq t_{n+1}}\left( 1-h_{t}(\mathbf{x})\right) \nonumber \\
    &\quad  +  \prod_{t; 1 \leq t \leq t_n}\left( 1-h_{t}(\mathbf{x})\right) \left(\left(1-h_{\theta^\prime,t_{n+1}}(\mathbf{x})\right) - \left( 1-h_{t_{n+1}}(\mathbf{x}) \right) +\nu -\nu\right)  \geq 0  \nonumber \\
    n\nu\left(1-h_{\theta^\prime,t_{n+1}}(\mathbf{x})\right) &\geq \prod_{t; 1 \leq t \leq t_{n+1}}\left( 1-h_{\theta^\prime,t}\left(\mathbf{x}\right)\right)  - \prod_{t; 1 \leq t \leq t_{n+1}}\left( 1-h_{t}(\mathbf{x})\right) \nonumber \\
    &\quad  +  \prod_{t; 1 \leq t \leq t_n}\left( 1-h_{t}(\mathbf{x})\right) \left(\left(1-h_{\theta^\prime,t_{n+1}}(\mathbf{x})\right) - \left( 1-h_{t_{n+1}}(\mathbf{x}) \right) -\nu \right) \nonumber \\
    & \quad +\prod_{t; 1 \leq t \leq t_n}\left( 1-h_{t}(\mathbf{x})\right)\nu  \geq 0  \nonumber \\
    n\nu\left(1-h_{\theta^\prime,t_{n+1}}(\mathbf{x})\right) &+A 
    \geq \prod_{t; 1 \leq t \leq t_{n+1}}\left( 1-h_{\theta^\prime,t}\left(\mathbf{x}\right)\right)  - \prod_{t; 1 \leq t \leq t_{n+1}}\left( 1-h_{t}(\mathbf{x})\right) 
    \geq A \label{eq:Proposition2_X_6}
\end{align}
where 
\begin{align*}
    A \defeq \prod_{t; 1 \leq t \leq t_n}\left( 1-h_{t}(\mathbf{x})\right) \left( \left( 1-h_{t_{n+1}}(\mathbf{x}) \right)+\nu-\left(1-h_{\theta^\prime,t_{n+1}}(\mathbf{x})\right) \right) - \prod_{t; 1 \leq t \leq t_n}\left( 1-h_{t}(\mathbf{x})\right)\nu .
\end{align*}
We will next show that $A \geq -\nu$. Then, we will show that $n\nu\left(1-h_{\theta^\prime,t_{n+1}}(\mathbf{x})\right) +A\leq (n+1)\nu$. These inequalities will be used with Equation \ref{eq:Proposition2_X_6} to conclude that $\Big|S(t_n|\mathbf{x};\theta^\prime) - S(t_n|\mathbf{x}) \Big| \leq (n+1)\nu$.

Our subsequent step will be to establish that $A \geq -\nu$. Following that, we will demonstrate that $n\nu\left(1-h_{\theta^\prime,t_{n+1}}(\mathbf{x})\right) +A\leq (n+1)\nu$. These inequalities will be employed in conjunction with Equation \ref{eq:Proposition2_X_6} to confirm that the absolute difference between $S(t_n|\mathbf{x};\theta^\prime)$ and $S(t_n|\mathbf{x})$ is less than or equal to $(n+1)\nu$.
\begin{align}
    A &\geq - \prod_{t; 1 \leq t \leq t_n}\left( 1-h_{t}(\mathbf{x})\right)\nu \textrm{$\quad\quad$($\because$ Equation \ref{eq:Proposition2_X_5})} \nonumber \\
    A&\geq -\nu
    \label{eq:Proposition2_X_7}
\end{align}
Next, we will show that $n\nu\left(1-h_{\theta^\prime,t_{n+1}}(\mathbf{x})\right) +A\leq (n+1)\nu$.
\begin{align}
    n\nu\left(1-h_{\theta^\prime,t_{n+1}}(\mathbf{x})\right) +A &\leq n\nu+A \textrm{$\quad\quad$($\because$ $0\leq 1-h_{\theta^\prime,t_{n+1}}(\mathbf{x}) \leq 1$)} \nonumber \\
    &\leq n\nu+ \prod_{t; 1 \leq t \leq t_n}\left( 1-h_{t}(\mathbf{x})\right) \left( \left( 1-h_{t_{n+1}}(\mathbf{x}) \right)-\left(1-h_{\theta^\prime,t_{n+1}}(\mathbf{x})\right) \right) \nonumber \\
    & \leq n\nu+ \prod_{t; 1 \leq t \leq t_n} \left( 1-h_{t}(\mathbf{x})\right) \nu \textrm{$\quad\quad$($\because$ Equation \ref{eq:Proposition2_X_5})} \nonumber \\
    & \leq n\nu+ \nu = (n+1)\nu \label{eq:Proposition2_X_8}
\end{align}
With Equations \ref{eq:Proposition2_X_6}, \ref{eq:Proposition2_X_7}, and \ref{eq:Proposition2_X_8}, 
\begin{align*}
    (n+1) \nu \geq
    \prod_{t; 1 \leq t \leq t_{n+1}}\left( 1-h_{\theta^\prime,t}\left(\mathbf{x}\right)\right)  - \prod_{t; 1 \leq t \leq t_{n+1}}\left( 1-h_{t}(\mathbf{x})\right) \geq -\nu  \\
    \Big | \prod_{t; 1 \leq t \leq t_{n+1}}\left( 1-h_{\theta^\prime,t}\left(\mathbf{x}\right)\right)  - \prod_{t; 1 \leq t \leq t_{n+1}}\left( 1-h_{t}(\mathbf{x})\right) \Big | \leq (n+1)\nu   \\
    \Big | S(t_n|\mathbf{x};\theta^\prime)-S(t_n|\mathbf{x}) \Big | \leq (n+1)\nu.
\end{align*}
\begin{lemma}
    \label{lemma:Proposition2_X_9}
    With $\theta^\prime$ be as in Assumption \ref{assumption_5} for $\phi \in \mathcal{H}$,
    \begin{align}
        \frac{1}{\tau} \sum_{t=1}^\tau &\left( \constraintterma{S(t|\mathbf{x}_j;\theta^\prime)} \right.\nonumber \\ 
        & \left.-\constraintterma{S(t|\mathbf{x}_j)} \right) 
        \leq (\tau+1)\nu.
        \label{lemma:Proposition_2_X_9}
    \end{align}
\end{lemma}
\proof{\underline{Proof}:}  
    \begin{align}
        \frac{1}{\tau} \sum_{t=1}^\tau &\left( \constraintterma{S(t|\mathbf{x}_j;\theta^\prime)} \right.\nonumber \\ 
        & \left.-\constraintterma{S(t|\mathbf{x}_j)} \right) \nonumber \\
        =&\frac{1}{\tau} \sum_{t=1}^\tau 
        \EDixy\left[\left( \frac{1}{N_i}\sum_{j=1}^{N_i}S(t|\mathbf{x}_j;\theta^\prime)-\SKM[t]
        \right.\right.   \nonumber\\ &\quad\quad\quad\quad\quad\quad\quad\quad\quad\quad\quad\left.\left.- \frac{1}{N_i}\sum_{j=1}^{N_i}S(t|\mathbf{x}_j)+\SKM[t]\right) \right.
        \nonumber \\        
        &\left. \quad\quad\quad\quad\quad\quad\quad\quad\quad\times \left(
        \frac{1}{N_i}\sum_{j=1}^{N_i}S(t|\mathbf{x}_j;\theta^\prime)-\SKM[t]  \right.\right.\nonumber \\   
        &\quad\quad\quad\quad\quad\quad\quad\quad\quad\quad\quad\left.\left.+\frac{1}{N_i}\sum_{j=1}^{N_i}S(t|\mathbf{x}_j)-\SKM
        \right) \right]\nonumber \\
        =&\frac{1}{\tau} \sum_{t=1}^\tau 
        \EDixy\left[ \left(\frac{1}{N_i}\sum_{j=1}^{N_i}S(t|\mathbf{x}_j;\theta^\prime) -  \frac{1}{N_i}\sum_{j=1}^{N_i}S(t|\mathbf{x}_j)
        \right)\right. \nonumber \\
        &\quad\quad\quad\quad\quad\quad\quad\quad \left.\times\left(
        \frac{1}{N_i}\sum_{j=1}^{N_i}S(t|\mathbf{x}_j;\theta^\prime)+S(t|\mathbf{x}_j)-2\SKM[t] \right)\right] \nonumber \\
        \leq & \frac{1}{\tau} \sum_{t=1}^\tau \left(
        \EDixy\left[ \frac{1}{N_i}\sum_{j=1}^{N_i}S(t|\mathbf{x}_j;\theta^\prime) -  \frac{1}{N_i}\sum_{j=1}^{N_i}S(t|\mathbf{x}_j)
        \right]\right) \times 2 \nonumber \\
        &\quad (\because \quad S(t|\mathbf{x}_j;\theta^\prime), S(t|\mathbf{x}_j), \SKM \in [0,1]) \nonumber \\
        \leq & \frac{1}{\tau} \sum_{t=1}^\tau \left(
        \EDixy\left[ \frac{1}{N_i}\sum_{j=1}^{N_i} \Big | S(t|\mathbf{x}_j;\theta^\prime) - S(t|\mathbf{x}_j)
        \Big | \right]\right) \times 2 \nonumber \\
        \leq & \frac{1}{\tau} \sum_{t=1}^\tau t \nu  \times 2 = (\tau+1)\nu   \quad\quad\quad(\because \quad \textrm{Lemma } \ref{lemma:Proposition2_X_1})
    \end{align}
\endproof{\Halmos}

\begin{lemma}
    \label{lemma:Proposition2_X_10}
    \begin{align}
        \mathbb{E}_{(\mathbf{x},t,\delta)\sim \mathfrak{D}_1}\left[\LDRSA[\phi^{\theta^\prime}] - \LDRSA\right] \leq (M_1+M_2 \tau+M_3 \tau+M_4\tau)\nu \label{eq:Proposition2_X_14_5}
    \end{align}
for some constant $M_1,M_2,M_3, M_4 \in \mathbb{R}$.
\end{lemma}

\proof{\underline{Proof}:}
\begin{align}
    &\mathbb{E}_{(\mathbf{x},t,\delta)\sim \mathfrak{D}_1}\left[\LDRSA[\phi^{\theta^\prime}] - \LDRSA\right] \nonumber\\
    &=\mathbb{E}_{(\mathbf{x},t,\delta)\sim \mathfrak{D}_1}\left[-\mathbb{I}\{\delta=1\}\log P(T=t|\mathbf{x},\phi^{\theta^\prime})
    -\mathbb{I}\{\delta=1\}\log P(T\leq\tau|\mathbf{x},\phi^{\theta^\prime}) 
    -\mathbb{I}\{\delta=0\}\log P(T>t|\mathbf{x},\phi^{\theta^\prime}) \right. \nonumber \\
    &\quad \quad\quad\quad \left.+\mathbb{I}\{\delta=1\}\log P(T=t|\mathbf{x})
    +\mathbb{I}\{\delta=1\}\log P(t\leq\tau|\mathbf{x},\phi) 
    +\mathbb{I}\{\delta=0\}\log P(T>t|\mathbf{x},\phi)\right] \nonumber \\
    &=\mathbb{E}_{(\mathbf{x},t,\delta)\sim \mathfrak{D}_1}\left[-\mathbb{I}\{\delta=1\}\left(\log h_{\theta^\prime,t}(\mathbf{x})+\sum_{t^\prime| t^\prime<t}\log(1-h_{\theta^\prime,t^\prime}(\mathbf{x})) \right)\right. \nonumber \\
    &\quad\quad\quad\quad\quad\quad -\mathbb{I}\{\delta=1\} \log \left( 1- \prod_{t^\prime|t^\prime< \tau} \left( 1-h_{\theta^\prime,t^\prime}(\mathbf{x})\right)\right) \nonumber \\
    &\quad\quad\quad\quad\quad\quad -\mathbb{I}\{\delta=0\} \sum_{t^\prime|t^\prime < t}\log \left(  1-h_{\theta^\prime,t^\prime}(\mathbf{x})\right) \nonumber \\
    &\quad\quad\quad\quad\quad\quad +\mathbb{I}\{\delta=1\}\left(\log h_{t}(\mathbf{x})+\sum_{t^\prime| t^\prime<t}\log(1-h_{t^\prime}(\mathbf{x})) \right) \nonumber \\
    &\quad\quad\quad\quad\quad\quad +\mathbb{I}\{\delta=1\} \log \left( 1- \prod_{t^\prime|t^\prime< \tau} \left( 1-h_{t^\prime}(\mathbf{x})\right)\right) \nonumber \\
    &\quad\quad\quad\quad\quad\quad \left. +\mathbb{I}\{\delta=0\} \sum_{t^\prime|t^\prime < t}\log \left(  1-h_{t^\prime}(\mathbf{x})\right) \right] \nonumber \\
    &=\mathbb{E}_{(\mathbf{x},t,\delta)\sim \mathfrak{D}_1}\left[\quad\mathbb{I}\{\delta=1\}\left( \log h_{t}(\mathbf{x}) -\log h_{\theta^\prime,t}(\mathbf{x})\right)\right. \nonumber \\
    &\quad\quad\quad\quad\quad\quad+\mathbb{I}\{\delta=1\}\left( \sum_{t^\prime| t^\prime<t}\log(1-h_{t^\prime}(\mathbf{x})) -\log(1-h_{\theta^\prime,t^\prime}(\mathbf{x}))\right) \nonumber \\
    &\quad\quad\quad\quad\quad\quad +\mathbb{I}\{\delta=1\} \left(\log \left( 1- \prod_{t^\prime|t^\prime< \tau} \left( 1-h_{t^\prime}(\mathbf{x})\right)\right)-\log \left( 1- \prod_{t^\prime|t^\prime< \tau} \left( 1-h_{\theta^\prime,t^\prime}(\mathbf{x})\right)\right) \right)\nonumber \\
    &\quad\quad\quad\quad\quad\quad \left. +\mathbb{I}\{\delta=0\} \left( \sum_{t^\prime|t^\prime < t}\log \left(  1-h_{t^\prime}(\mathbf{x})\right)-\log \left(  1-h_{\theta^\prime,t^\prime}(\mathbf{x})\right) \right)\right]
    \label{eq:Proposition2_X_10}
\end{align}
Our next step will be to demonstrate that the magnitude of each term in Equation \ref{eq:Proposition2_X_10} can be constrained based on the distance between the parameterized instantaneous hazard rate, $h_{\theta^\prime,t}(\mathbf{x})$, and the instantaneous hazard rate, $h_t(\mathbf{x})$, or $\nu$.

\paragraph{\underline{1. $\mathbb{I}\{\delta=1\}\left( \log h_{t}(\mathbf{x}) -\log h_{\theta^\prime,t}(\mathbf{x})\right)$}}
\begin{align}
    \mathbb{I}\{\delta=1\}\left( \log h_{t}(\mathbf{x}) -\log h_{\theta^\prime,t}(\mathbf{x})\right) 
    &\leq \Big |\mathbb{I}\{\delta=1\}\left( \log h_{t}(\mathbf{x}) -\log h_{\theta^\prime,t}(\mathbf{x})\right) \Big | \nonumber \\ 
    & \leq M_1 \Big |h_{t}(\mathbf{x})- h_{\theta^\prime,t}(\mathbf{x})\Big |  \quad\quad(\because\textrm{ Assumption }\ref{assumption_4}\textrm{ and Lipschitz function} )\nonumber \\
    & \leq M_1 \nu \quad\quad\quad\quad(\because\textrm{ Assumption \ref{assumption_5}}) \label{eq:Proposition2_X_11}
\end{align}
\paragraph{\underline{2. $\mathbb{I}\{\delta=1\}\left( \sum_{t^\prime| t^\prime<t}\log(1-h_{t^\prime}(\mathbf{x})) -\log(1-h_{\theta^\prime,t^\prime}(\mathbf{x}))\right)$}}
\begin{align}
    &\mathbb{I}\{\delta=1\}\left( \sum_{t^\prime| t^\prime<t}\log(1-h_{t^\prime}(\mathbf{x})) -\log(1-h_{\theta^\prime,t^\prime}(\mathbf{x}))\right) \nonumber \\
    &\leq \Big |\mathbb{I}\{\delta=1\}\left( \sum_{t^\prime| t^\prime<t}\log(1-h_{t^\prime}(\mathbf{x})) -\log(1-h_{\theta^\prime,t^\prime}(\mathbf{x}))\right)\Big | \nonumber \\
    & \leq \tau\max_{t^\prime|t^\prime < t}\Big | \log(1-h_{t^\prime}(\mathbf{x})) -\log(1-h_{\theta^\prime,t^\prime}(\mathbf{x}))\Big | \nonumber \\
    & \leq \tau\max_{t^\prime|t^\prime < t}\Big | h_{t^\prime}(\mathbf{x}) -h_{\theta^\prime,t^\prime}\Big | \nonumber \\
    & \leq M_2 \tau \nu \label{eq:Proposition2_X_12}
\end{align}
\paragraph{\underline{3. $\mathbb{I}\{\delta=1\} \left(\log \left( 1- \prod_{t^\prime|t^\prime< \tau} \left( 1-h_{t^\prime}(\mathbf{x})\right)\right)-\log \left( 1- \prod_{t^\prime|t^\prime< \tau} \left( 1-h_{\theta^\prime,t^\prime}(\mathbf{x})\right)\right) \right)$}}
\begin{align}
    &\mathbb{I}\{\delta=1\} \left(\log \left( 1- \prod_{t^\prime|t^\prime< \tau} \left( 1-h_{t^\prime}(\mathbf{x})\right)\right)-\log \left( 1- \prod_{t^\prime|t^\prime< \tau} \left( 1-h_{\theta^\prime,t^\prime}(\mathbf{x})\right)\right) \right) \nonumber \\
    &\leq \Big|\mathbb{I}\{\delta=1\} \left(\log \left( 1- \prod_{t^\prime|t^\prime< \tau} \left( 1-h_{t^\prime}(\mathbf{x})\right)\right)-\log \left( 1- \prod_{t^\prime|t^\prime< \tau} \left( 1-h_{\theta^\prime,t^\prime}(\mathbf{x})\right)\right) \right) \Big| \nonumber \\
    &\leq M_3 \Big| \prod_{t^\prime|t^\prime< \tau} \left( 1-h_{t^\prime}(\mathbf{x})\right) - \prod_{t^\prime|t^\prime< \tau} \left( 1-h_{\theta^\prime,t^\prime}(\mathbf{x})\right)\Big| \nonumber \\
    & \leq M_3 \tau \nu \quad\quad\quad\quad\quad\quad(\because\textrm{ Lemma }\ref{lemma:Proposition2_X_1}) \label{eq:Proposition2_X_13}
\end{align}
\paragraph{\underline{4. $\mathbb{I}\{\delta=0\} \left( \sum_{t^\prime|t^\prime < t}\log \left(  1-h_{t^\prime}(\mathbf{x})\right)-\log \left(  1-h_{\theta^\prime,t^\prime}(\mathbf{x})\right) \right)$}}
\begin{align}
    &\mathbb{I}\{\delta=0\} \left( \sum_{t^\prime|t^\prime < t}\log \left(  1-h_{t^\prime}(\mathbf{x})\right)-\log \left(  1-h_{\theta^\prime,t^\prime}(\mathbf{x})\right) \right) \nonumber \\
    &\leq \Big| \mathbb{I}\{\delta=0\} \left( \sum_{t^\prime|t^\prime < t}\log \left(  1-h_{t^\prime}(\mathbf{x})\right)-\log \left(  1-h_{\theta^\prime,t^\prime}(\mathbf{x})\right) \right) \Big| \nonumber \\
    &\leq \sum_{t^\prime|t^\prime < t}\Big|\log \left(  1-h_{t^\prime}(\mathbf{x})\right)-\log \left(  1-h_{\theta^\prime,t^\prime}(\mathbf{x})\right) \Big| \nonumber \\
    &\leq \sum_{t^\prime|t^\prime < t}\Big|h_{t^\prime}(\mathbf{x})-h_{\theta^\prime,t^\prime}(\mathbf{x}) \Big| \nonumber \\
    & \leq M_4 \tau \nu \label{eq:Proposition2_X_14}
\end{align}
From Equations \ref{eq:Proposition2_X_10}, \ref{eq:Proposition2_X_11}, \ref{eq:Proposition2_X_12}, \ref{eq:Proposition2_X_13}, and \ref{eq:Proposition2_X_14}, we have 
\begin{align}
    \mathbb{E}_{(\mathbf{x},t,\delta)\sim \mathfrak{D}_1}\left[\LDRSA[\phi^{\theta^\prime}] - \LDRSA\right] \leq (M_1+M_2 \tau+M_3 \tau+M_4\tau)\nu .
\end{align}
\endproof{\Halmos}

\begin{proposition}
    \label{Proposition3}
    Suppose $\Hat{\theta}^*$ attains the saddle-point in equation \ref{eq:Hat_D_CSL}. Under Assumptions \ref{assumption_3}-\ref{assumption_5}, it is guaranteed with a probability of $1-(3m+2)\delta^\prime$ over the samples drawn from the distribution $\mathcal{D}_i$ that
    \begin{align}
        \mid D^*_\theta - \Hat{D}^*_\theta\mid \leq (1+\kappa) \zeta  \label{eq:chamon_thesis_4.13}\quad\quad&\textrm{and}\\
        \frac{1}{\tau} \sum_{t=1}^\tau \constraintterma{S(t|\mathbf{x}_j;{\Hat{\theta}^*})}  &\leq c_i + \rho_{\mathcal{N}_i},
        \label{eq:chamon_thesis_4.14}
    \end{align}
    for $i=1,\ldots,m$ where $\zeta \defeq \max_{i} \rho_{\mathcal{N}_i}$, $\rho_{\mathcal{N}_i}\defeq B\sqrt{\frac{1}{\mathcal{N}_i}\left( 1+ \log\left( \frac{4\left( 2\mathcal{N}_i\right)^{d_{vc}}}{\delta^\prime/\tau}\right)\right)}$, and \linebreak$\kappa=\max\left(\|\mathbf{\mu}^*\|_1, \|\Hat{\mathbf{\mu}}^*\|_1 \right)$ for $\mathbf{\mu}^*$ and $\Hat{\mathbf{\mu}}^*$ being solutions of \ref{eq:Dtheta_CSL} and \ref{eq:Hat_D_CSL} that achieve $D_\theta^*$ and $\Hat{D}^*_\theta$, respectively.
\end{proposition}
\proof{\underline{Proof}:}
Following a similar approach as in Proposition \ref{Proposition2}, we will initially demonstrate the feasibility of $\Hat{\theta}^*$ for \ref{eq:PIV}. Subsequently, we will establish the proximity between the solution obtained from \ref{eq:Hat_D_CSL} or $\Hat{D}_\theta^*$ and the solution of \ref{eq:Dtheta_CSL} or $D_\theta^*$, which had been established as being close to the solution of \ref{eq:P_CSO} or $P^*$ in Proposition \ref{Proposition2}.
\paragraph{Feasibility.}
Utilizing the PAC learnability of $\mathcal{P}$, as assumed in Assumption \ref{assumption_5}, we can apply the generalization bounds from classical learning theory.

Suppose that there exists at least one $i>0$ such that
\[
\frac{1}{\tau}\sum_{t=1}^\tau\left(\frac{1}{\mathcal{N}_i}\sum_{n_i=1}^{\mathcal{N}_i}\left(\frac{1}{N_i}\sum_{j=1}^{N_i}S(t|\mathbf{x}^i_{n_i,j};\theta)-\SKMProp\right)^2 \right) > c_i
\]
for the samples $\{(\mathbf{x}^i_{n_i,j},t^i_{n_i,j},\delta^i_{n_i,j})\}_{j=1}^{N_i}$ from distribution $\mathcal{D}_i$. In this case, since $\mathbf{\mu}$ is unbounded above, we deduce that $\Hat{D}^*_\theta \rightarrow +\infty$. However, Assumption \ref{assumption_3} implies that $\Hat{D}^*_\theta < +\infty$. Specifically, consider the empirical dual function given by
\begin{equation}
\Hat{d}(\mu)=\min_{\theta \in \mathbb{R}^p} \Hat{L}(\phi^\theta, \mu)
\label{eq:chamon_thesis_A.59}
\end{equation}
where $\Hat{L}$ is the empirical Lagrangian from Equation \ref{eq:chamon_thesis_4.4}. By leveraging the fact that $\mathcal{L}_{DRSA}$ is $B$-bounded and there exists a strictly feasible $\Hat{\theta}$ (Assumption \ref{assumption_3}), we can utilize the same argument leading to Equation 
 \ref{eq:chamon_thesis_A_47} to establish that $\Hat{d}(\mathbf{\mu})<B$. Consequently, it must hold that
\begin{align}
\frac{1}{\tau}\sum_{t=1}^\tau\left(\frac{1}{\mathcal{N}_i}\sum_{n_i=1}^{\mathcal{N}_i}\left(\frac{1}{N_i}\sum_{j=1}^{N_i}S(t|\mathbf{x}_{n_i,j}^{i};\Hat{\theta}^*)-\SKMProp[t]\right)^2 \right) \leq c_i
\label{eq:chamon_thesis_A_60}
\end{align}
for $i=1,\ldots,m$. We can now proceed to employ Lemma \ref{lemma:chamon_paper_propositionIII.1} to demonstrate that $\phi^{\Hat{\theta}^*}$ must approximately satisfy each constraint of \ref{eq:P_CSO} with a high probability $1-\delta^\prime$ over the samples $\{(\mathbf{x}^i_{n_i,j},t^i_{n_i,j},\delta^i_{n_i,j})\}_{j=1}^{N_i}$ that
\begin{align}
    &\frac{1}{\tau} \sum_{t=1}^\tau \constraintterma{S(t|\mathbf{x}_j;\Hat{\theta}^*)} \nonumber \\ 
    &\leq \frac{1}{\tau}\sum_{t=1}^\tau\left(\frac{1}{\mathcal{N}_i}\sum_{n_i=1}^{\mathcal{N}_i}\left(\frac{1}{N_i}\sum_{j=1}^{N_i}S(t|\mathbf{x}_j^{n_i};\Hat{\theta}^*)-y_t^{n_i}\right)^2 \right)  + \rho_{\mathcal{N}_i}
    \label{eq:chamon_thesis_A_61}
\end{align}
for each $i=1,\ldots,m$, where $\rho_N$ is as follows
\begin{equation}
    \rho_N= B\sqrt{\frac{1}{N}\left( 1+ \log\left( \frac{4\left( 2 N\right)^{d_{vc}}}{\delta^\prime / \tau}\right)\right)}.
    \label{eq:chamon_thesis_4.6}
\end{equation} 

Moreover, Equations \ref{eq:chamon_thesis_A_60} and \ref{eq:chamon_thesis_A_61} result in 
\begin{equation}
    \frac{1}{\tau} \sum_{t=1}^\tau\constraintterma{S(t|\mathbf{x}_j;\Hat{\theta}^*)}  \leq c_i + \rho_{\mathcal{N}_i} \leq c_i + \zeta
    \label{eq:Proposition3_eq3}
\end{equation}
for $i=1,\ldots,m$, $\zeta\defeq \max_i \rho_{\mathcal{N}_i}$, and by Boole's inequality\footnote{Boole's inequality states that for a set of events $A_1, \ldots, A_n$, $\mathbb{P}\left(\bigcup_{i=1}^n A_i\right) \leq \sum_{i=1}^n \mathbb{P}\left(A_i\right)$.}, these equations will simultaneously hold with probability of at least $1-m\delta^\prime$.
\paragraph{Close to Optimality.} Suppose $\mathbf{\mu}^*$ and $\Hat{\mathbf{\mu}}^*$ are solutions of equations \ref{eq:Dtheta_CSL} and \ref{eq:Hat_D_CSL} respectively. The set of minimizers for the Lagrangians defined in Equations \ref{eq:Lagrangian} and \ref{eq:chamon_thesis_4.4} are as follows
\[
\Theta^\dagger(\mathbf{\mu})=\argmin_\theta L(\phi^\theta,\mathbf{\mu}) \quad\textrm{ and }\quad
\Hat{\Theta}^\dagger(\mathbf{\mu})=\argmin_\theta \Hat{L}(\phi^\theta,\mathbf{\mu}).
\]
Using the solution of \ref{eq:Dtheta_CSL}, or $\mathbf{\mu}^*$, we have
\begin{align*}
    D_\theta^*- \Hat{D}^*_\theta &= \min_{\theta \in \mathbb{R}^p} L(\phi^\theta,\mathbf{\mu}^*) - \min_{\theta \in \mathbb{R}^p} \Hat{L}(\phi^\theta,\Hat{\mathbf{\mu}}^*) \\
    &\leq \min_{\theta \in \mathbb{R}^p} L(\phi^\theta,\mu^*) - \min_{\theta \in \mathbb{R}^p} \Hat{L}(\phi^\theta,\mathbf{\mu}^*).
\end{align*}
As $\Hat{\theta}^\dagger \in \Hat{\Theta}^\dagger(\mathbf{\mu}^*)$ is a suboptimal solution for $L(\phi^\theta,\mathbf{\mu}^*)$, we obtain
\begin{equation}
    \label{eq:chamon_paper_64}
    D_\theta^*-\Hat{D}^*_\theta \leq L(\phi^{\Hat{\theta}^\dagger},\mathbf{\mu}^*)-\Hat{L}(\phi^{\Hat{\theta}^\dagger},\mathbf{\mu}^*).
\end{equation}
Furthermore, by employing a similar line of reasoning, we can demonstrate that \linebreak$D_\theta^*-\Hat{D}^*_\theta \geq L(\phi^{\theta^\dagger},\Hat{\mathbf{\mu}}^*)-\Hat{L}(\phi^{\theta^\dagger},\Hat{\mathbf{\mu}}^*)$. Given that $\Hat{\mathbf{\mu}}^*$ is the solution to equation \ref{eq:Hat_D_CSL}, we have
\begin{align*}
    D_\theta^*- \Hat{D}^*_\theta &= \min_{\theta \in \mathbb{R}^p} L(\phi^\theta,\mathbf{\mu}^*) - \min_{\theta \in \mathbb{R}^p} \Hat{L}(\phi^\theta,\Hat{\mathbf{\mu}}^*) \\
    &\geq \min_{\theta \in \mathbb{R}^p} L(\phi^\theta,\Hat{\mathbf{\mu}}^*) - \min_{\theta \in \mathbb{R}^p} \Hat{L}(\phi^\theta,\Hat{\mathbf{\mu}}^*)
\end{align*}
and then
\begin{equation}
    \label{eq:chamon_paper_65}
    D_\theta^*-\Hat{D}^*_\theta \geq L(\phi^{\theta^\dagger},\Hat{\mathbf{\mu}}^*)-\Hat{L}(\phi^{\theta^\dagger},\Hat{\mathbf{\mu}}^*)
\end{equation}
for $\theta^\dagger \in \Theta^\dagger(\Hat{\mathbf{\mathbf{\mu}}}^*)$, since it is suboptimal for $L(\phi^\theta,\Hat{\mathbf{\mu}}^*)$.

With Equations \ref{eq:chamon_paper_64} and \ref{eq:chamon_paper_65}, we have
\begin{align}
    | D_\theta^*-\Hat{D}^*_\theta | \leq \max\left( \Big |  L(\phi^{\Hat{\theta}^\dagger},\mathbf{\mu}^*)-\Hat{L}(\phi^{\Hat{\theta}^\dagger},\mathbf{\mu}^*)\Big |, 
     \Big | L(\phi^{\theta^\dagger},\Hat{\mathbf{\mu}}^*)-\Hat{L}(\phi^{\theta^\dagger},\Hat{\mathbf{\mu}}^*)\Big |\right) . 
     \label{eq:chamon_paper_66}
\end{align}
By applying Lemma \ref{lemma:chamon_paper_propositionIII.1}, we can derive
\begin{align}
    |L(\phi^\theta,\mathbf{\mu}) - \Hat{L}(\phi^\theta,\mathbf{\mu})| &\leq \rho_{\mathcal{N}_0} + \sum_{i=1}^m \mu_i \rho_{\mathcal{N}_i} \nonumber\\
    &\leq (1+\|\mathbf{\mu}\|_1)\max_i \rho_{\mathcal{N}_i} \nonumber\\
    & = (1+\|\mathbf{\mu}\|_1)\zeta  \quad\quad\quad\quad\quad\textrm{($\because\;\zeta\defeq \max_i \rho_{\mathcal{N}_i}$)} \label{eq:chamon_paper_67}
\end{align}
which holds uniformly over $\theta$ with probability $1- (m+1)\delta^\prime$.

To demonstrate that $| D^*_\theta-\Hat{D}^*_\theta | \leq (1+\kappa) \zeta$ with a probability of at least $1- (2m+2)\delta^\prime$, we will employ Boole's inequality together with Equations \ref{eq:chamon_paper_66} and \ref{eq:chamon_paper_67}. 

With Equation \ref{eq:chamon_paper_67}, we have
\begin{align}
    \Big |  L(\phi^{\Hat{\theta}^\dagger},\mathbf{\mu}^*)-\Hat{L}(\phi^{\Hat{\theta}^\dagger},\mathbf{\mu}^*)\Big | \leq (1+\|\mathbf{\mu}^*\|_1)\zeta \leq (1+\kappa)\zeta  \label{eq:Proposition3_eq1} \\
    \Big | L(\phi^{\theta^\dagger},\Hat{\mathbf{\mu}}^*)-\Hat{L}(\phi^{\theta^\dagger},\Hat{\mathbf{\mu}}^*)\Big | \leq (1+\|\Hat{\mathbf{\mu}}^*\|_1)\zeta\leq (1+\kappa)\zeta
    \label{eq:Proposition3_eq2}   
\end{align}
where the second inequality of each equation is from the definition of $\kappa$ which is $\max\left(\|\mathbf{\mu}^*\|_1, \|\Hat{\mathbf{\mu}}^*\|_1 \right)$. Each equation will hold with a probability of at least $1-(m+1)\delta^\prime$. As a result, both equations will hold with a probability of at least $1- (2m+2)\delta^\prime$ based on Boole's inequality. Consequently, we can conclude that $| D^*_\theta-\Hat{D}^*_\theta | \leq (1+\kappa) \zeta$ with a probability of at least $1- (2m+2)\delta^\prime$.

Given the fact that $| D^*-\Hat{D}^*_\theta | \leq (1+\kappa) \zeta$ with a probability of at least $1- (2m+2)\delta^\prime$, and Equation \ref{eq:Proposition3_eq3} holds for all $i=1,\ldots,m$ with a probability of at least $1-m\delta^\prime$, we can infer that both of these facts will hold simultaneously with a probability of at least $1- (3m+2)\delta^\prime$.
\endproof{\Halmos} 
\begin{lemma}
    \label{lemma:chamon_paper_propositionIII.1}
    Let $d_{vc}$ be the upper bounds on the VC dimension with respect to $G_0$ and $G_{i,t}$ where
    \begin{align}
        G_0(\phi,\,\mathbf{x},t,\delta) \defeq \LDRSA\quad \nonumber \\
        G_{i,t}(\phi,\{\mathbf{x}_j,t_j,\delta_j\}_{j=1}^{N_i}) \defeq  \left(\frac{1}{N_i} \sum_{j=1}^{N_i} S(t|\mathbf{x}_j)-\SKM\right)^2 \nonumber
    \end{align}
    for $i=1,\ldots,m$ and $t=1,\ldots,\tau$. 
    
    Equation \ref{eq:chamon_paper_14.1} holds with probability at least $1-\delta^\prime$
    \begin{align}
        \Big | \mathbb{E}_{(\mathbf{x},t,\delta)\sim \mathfrak{D}_1}\left[ \LDRSA \right] - \frac{1}{N_1}\sum_{j=1}^{N_1} \mathcal{L}_{DRSA}(\phi,\,\mathbf{x}_j,t_j,\delta_j) \Big |&\leq \rho_{\mathcal{N}_0} \label{eq:chamon_paper_14.1}
    \end{align}
    where $\rho_{\mathcal{N}_0}\defeq B\sqrt{\frac{1}{N_1}\left( 1+ \log\left( \frac{4\left( 2 N_1\right)^{d_{vc}}}{\delta^\prime}\right)\right)}$ and for each $i=1,\ldots,m$, Equation \ref{eq:chamon_paper_14.2} holds with probability at least $1-\delta^\prime$
    \begin{align}
        \Bigg | &\frac{1}{\tau}\sum_{t=1}^\tau\constraintterma{S(t|\mathbf{x}_j)} \nonumber \\
        &- \frac{1}{\tau}\sum_{t=1}^\tau\frac{1}{\mathcal{N}_i}\sum_{n_i=1}^{\mathcal{N}_i}\left(\frac{1}{N_i}\sum_{i=1}^{N_i} S(t|\mathbf{x}_{n_ij}^{i})-\SKM\right)^2\Bigg | \leq \rho_{\mathcal{N}_i} 
        \label{eq:chamon_paper_14.2} 
    \end{align}
\end{lemma}
where $\rho_{\mathcal{N}_i}\defeq B\sqrt{\frac{1}{\mathcal{N}_i}\left( 1+ \log\left( \frac{4\left( 2\mathcal{N}_i\right)^{d_{vc}}}{\delta^\prime /\tau}\right)\right)}$
\proof{\underline{Proof}:}
Referring to Equation 3.26 of \cite{vapnik1999nature}, we have 
    \begin{align}
        \Big | \mathbb{E}_{(\mathbf{x},t,\delta)\sim \mathfrak{D}_1}\left[ \LDRSA \right] - \frac{1}{N_1}\sum_{j=1}^{N_1} \mathcal{L}_{DRSA}(\phi,\,\mathbf{x}^{1}_{1,j},t^{1}_{1,j},\delta^{1}_{1,j}) \Big | &\leq \frac{B}{2}\sqrt{\varepsilon_1} \label{eq:chamon_paper_14.1a} 
    \end{align}
where $\varepsilon_1\defeq 4\frac{d_{vc}\left(\ln\frac{2\mathcal{N}_0}{d_{vc}}+1\right)-\ln\left( \frac{\Tilde{\delta}_1}{4} \right)}{\mathcal{N}_0}$, $\mathcal{N}_0 \defeq N_1$, $\Tilde{\delta}_1\defeq\delta^\prime$ and for each $i=1,\ldots,m$,
    \begin{align}
        \Bigg | \constraintterma{S(t|\mathbf{x}_j)} \nonumber\\ -\frac{1}{\mathcal{N}_i}\sum_{n_i=1}^{\mathcal{N}_i}\left(\frac{1}{N_i}\sum_{i=1}^{N_i} S(t|\mathbf{x}_{n_i,j}^{i})-\SKMProp\right)^2\Bigg | &\leq \frac{B}{2} \sqrt{\varepsilon_2} \label{eq:chamon_paper_14.2a} 
    \end{align}
where $\varepsilon_2\defeq 4\frac{d_{vc}\left(\ln\frac{2\mathcal{N}_i}{d_{vc}}+1\right)-\ln\left( \frac{\Tilde{\delta}_2}{4} \right)}{\mathcal{N}_i}$, $\Tilde{\delta}_2\defeq\delta^\prime/\tau$.
Equations \ref{eq:chamon_paper_14.1a} and \ref{eq:chamon_paper_14.2a} hold with probability at least $1-\Tilde{\delta}_1$ and $1-\Tilde{\delta}_2$ respectively.  

Consider the term on the right hand side of Equations \ref{eq:chamon_paper_14.1a} and \ref{eq:chamon_paper_14.2a}. For $k=1 \text{ or } 2$,  
\begin{align}
    \frac{B}{2} \sqrt{\varepsilon_k} &= B\sqrt{\frac{d_{vc}\left(\log\frac{2\mathcal{N}_i}{d_{vc}}+1\right)-\log\left( \frac{\Tilde{\delta}_k}{4} \right)}{\mathcal{N}_i}}\nonumber \\
    &=B\sqrt{\frac{d_{vc}\left(\log(2\mathcal{N}_i)-\log(d_{vc})+1\right)-\log\left( \frac{\Tilde{\delta}_k}{4} \right)}{\mathcal{N}_i}}\nonumber \\
    &=B\sqrt{\frac{d_{vc}\ln(2\mathcal{N}_i)+d_{vc}\left(1-\log(d_{vc})\right)-\log\left( \frac{\Tilde{\delta}_k}{4} \right)}{\mathcal{N}_i}}\nonumber \\
    &\leq B\sqrt{\frac{d_{vc}\log(2\mathcal{N}_i)+d_{vc}\left(\frac{1}{d_{vc}}\right)-\log\left( \frac{\Tilde{\delta}_k}{4} \right)}{\mathcal{N}_i}} \quad\quad(\because \log(x) \geq 1-\frac{1}{x})\nonumber \\
    &= B\sqrt{\frac{1}{\mathcal{N}_i}\left( 1+ \log\left( \frac{4\left( 2\mathcal{N}_i\right)^{d_{vc}}}{\Tilde{\delta}_k}\right)\right)}. \label{eq:Proposition3_1}
\end{align}
Using Equations \ref{eq:chamon_paper_14.1a} and \ref{eq:Proposition3_1}, we can derive Equation \ref{eq:chamon_paper_14.1}. Similarly, to obtain Equation \ref{eq:chamon_paper_14.2}, we can apply Equations \ref{eq:chamon_paper_14.2a} and \ref{eq:Proposition3_1} in the following manner.
\begin{align}
    &\Bigg |\frac{1}{\tau} \sum_{t=1}^\tau \constraintterma{S(t|\mathbf{x}_j)} \nonumber\\
    &-\frac{1}{\tau} \sum_{t=1}^\tau\frac{1}{\mathcal{N}_i}\sum_{n_i=1}^{\mathcal{N}_i}\left(\frac{1}{N_i}\sum_{i=1}^{N_i} S(t|\mathbf{x}^i_{n_i,j})-\SKMProp\right)^2\Bigg | \nonumber \\
    &=\Bigg |\frac{1}{\tau}\sum_{t=1}^\tau \left( \constraintterma{S(t|\mathbf{x}_j)} \right. \nonumber \\
    &\quad \left.-\frac{1}{\mathcal{N}_i}\sum_{n_i=1}^{\mathcal{N}_i}\left(\frac{1}{N_i}\sum_{i=1}^{N_i} S(t|\mathbf{x}^i_{n_i,j})-\SKMProp\right)^2\right)\Bigg | \nonumber \\
    &\leq \frac{1}{\tau}\sum_{t=1}^\tau\Bigg | \constraintterma{S(t|\mathbf{x}_j)} \nonumber \\
    & \quad-\frac{1}{\mathcal{N}_i}\sum_{n_i=1}^{\mathcal{N}_i}\left(\frac{1}{N_i}\sum_{i=1}^{N_i} S(t|\mathbf{x}^i_{n_i,j})-\SKMProp\right)^2\Bigg | \nonumber \\ 
    & \leq \frac{1}{\tau}\sum_{t=1}^\tau \frac{B}{2} \sqrt{\varepsilon_2} \nonumber \\ 
    & = B\sqrt{\frac{1}{\mathcal{N}_i}\left( 1+ \log\left( \frac{4\left( 2\mathcal{N}_i\right)^{d_{vc}}}{\Tilde{\delta}_2}\right)\right)} = \rho_{\mathcal{N}_i}  \quad\quad (\because \textrm{ Equation } \ref{eq:Proposition3_1})
    \label{eq:Proposition3_2}
\end{align}
Equation \ref{eq:Proposition3_2} is valid as long as Equation \ref{eq:chamon_paper_14.2a} simultaneously holds for all $i=1,\ldots,m$. Consequently, applying Boole's inequality, we can conclude that Equation \ref{eq:Proposition3_2} holds with a probability of at least $1-\delta^\prime$.
\endproof{\Halmos}

The main result of this section is Theorem \ref{thm:chamon_thesis_3} establishing that \ref{eq:Hat_D_CSL} is a PACC learner (Definition \ref{definition:chamon_thesis_3}).
\begin{theorem}
    \label{thm:chamon_thesis_3}
    Under Assumptions \ref{assumption_3}-\ref{assumption:non_atomic_0} , there exists $\Hat{\theta}^* = \argmin_{\theta} \Hat{L}(\phi^\theta,\Hat{\mu}^*)$ such that, with probability $1-(3m+2)\delta$
    \begin{align}
        \label{eq:chamon_thesis_theorem3_eq0}
        \Big | P^* - \Hat{D}^*_\theta \Big |  \leq (M_1+M_2 \tau+M_3 \tau+M_4\tau)\nu + \lVert \Tilde{\mathbf{\mu}}^* \rVert_{1}\; (\tau+1) \nu + (1+\kappa)\zeta  \quad\quad&\textrm{and}\\
        \frac{1}{\tau} \sum_{t=1}^\tau \constraintterma{S_{\Hat{\theta}^*}(t|\mathbf{x}_j)}  &\leq c_i + \rho_{\mathcal{N}_i} \label{eq:chamon_thesis_theorem3_eq0.15}
    \end{align}
for $i=1,\ldots,m$ where $P^*$ is the value of \ref{eq:P_CSO}, and $\Hat{D}^*_\theta$ is the solution of the dual problem \ref{eq:Hat_D_CSL}. $M_1, M_2, M_3, M_4, \mathbf{\Tilde{\mu}^*}$ are as defined in Proposition \ref{Proposition2}. $\rho_{\mathcal{N}_i}, \zeta, \kappa$ are as defined in Proposition \ref{Proposition3}. 
\end{theorem}
\proof{\underline{Proof}:} To prove Equation \ref{eq:chamon_thesis_theorem3_eq0}, we will first demonstrate that 
\begin{equation}
    \Hat{D}^*_\theta \leq P^*+(M_1+M_2 \tau+M_3 \tau+M_4\tau)\nu + \lVert \Tilde{\mathbf{\mu}}^* \rVert_{1}\; (\tau+1) \nu + (1+\kappa) \zeta.
    \label{eq:chamon_thesis_theorem3_eq0.3}
\end{equation}
Then, it suffices to establish that 
\begin{equation}
    \Hat{D}^*_\theta \geq P^*-(M_1+M_2 \tau+M_3 \tau+M_4\tau)\nu - \lVert \Tilde{\mathbf{\mu}}^* \rVert_{1}\; (\tau+1) \nu - (1+\kappa) \zeta .
    \label{eq:chamon_thesis_theorem3_eq0.6}
\end{equation}
These inequalities can be established by using Propositions \ref{Proposition2} and \ref{Proposition3}. From Proposition \ref{Proposition2}, we can restate Equation \ref{eq:chamon_thesis_4.12} as follows:
\begin{align}
    P^* \leq D_\theta^* \leq P^*+(M_1+M_2 \tau+M_3 \tau+M_4\tau)\nu + \lVert \Tilde{\mathbf{\mu}}^* \rVert_{1}\; (\tau+1) \nu .\tag{\ref{eq:chamon_thesis_4.12}}
\end{align}
Furthermore, from Proposition \ref{Proposition3}, Equation \ref{eq:chamon_thesis_4.13} can be reformulated as
\begin{equation}
    \Hat{D}^*_\theta - (1+\kappa) \zeta   \leq D^*_\theta \leq \Hat{D}^*_\theta + (1+\kappa) \zeta  .\label{eq:chamon_thesis_theorem3_eq1}
\end{equation}

Firstly, we will show Equation \ref{eq:chamon_thesis_theorem3_eq0.3}. by using Equations \ref{eq:chamon_thesis_theorem3_eq1} (the first inequality) and \ref{eq:chamon_thesis_4.12} (the second inequality), which leads to the following relation.
\begin{align}
    \Hat{D}^*_\theta - (1+\kappa) \zeta &\leq P^*+(M_1+M_2 \tau+M_3 \tau+M_4\tau)\nu + \lVert \Tilde{\mathbf{\mu}}^* \rVert_{1}\; (\tau+1) \nu \nonumber \\
    \Hat{D}^*_\theta &\leq P^*+(M_1+M_2 \tau+M_3 \tau+M_4\tau)\nu + \lVert \Tilde{\mathbf{\mu}}^* \rVert_{1}\; (\tau+1) \nu + (1+\kappa) \zeta 
    \label{eq:chamon_thesis_theorem3_eq2}
\end{align}
Lastly, we will show Equation \ref{eq:chamon_thesis_theorem3_eq0.6} by utilizing Equations \ref{eq:chamon_thesis_theorem3_eq1} (the second inequality) and \ref{eq:chamon_thesis_4.12} (the first inequality), which imply the following relation.
\begin{align}
     P^* &\leq \Hat{D}^*_\theta + (1+\kappa) \zeta  \nonumber \\
     P^* - (1+\kappa) \zeta &\leq \Hat{D}^*_\theta  \nonumber \\
     P^* -(M_1+M_2 \tau+M_3 \tau+M_4\tau)\nu - \lVert \Tilde{\mathbf{\mu}}^* \rVert_{1}\; (\tau+1) \nu - (1+\kappa)\zeta &\leq \Hat{D}^*_\theta
     \label{eq:chamon_thesis_theorem3_eq3}
\end{align}
From Equations \ref{eq:chamon_thesis_theorem3_eq2} and \ref{eq:chamon_thesis_theorem3_eq3}, 
\begin{align*}
    \Big | P^* - \Hat{D}^*_\theta \Big |  \leq (M_1+M_2 \tau+M_3 \tau+M_4\tau)\nu + \lVert \Tilde{\mathbf{\mu}}^* \rVert_{1}\; (\tau+1) \nu + (1+\kappa)\zeta 
\end{align*}
Equation \ref{eq:chamon_thesis_theorem3_eq0.15} follows directly from Proposition \ref{Proposition3}.
\endproof{\Halmos}
\section{Experiments} \label{sec:experiments} 

\subsection{Datasets}
We use three real-world clinical survival datasets (two public and one private).

\textbf{nwtco}.
This public dataset contains data from the National Wilm’s Tumor Study \citep{breslow1999design}. It contains 4028 instances (individuals), of which 86\% are censored. Each individual has 7 features such as disease stage and histology information. For this dataset, we are interested in the distribution of time until relapse. The number of examples in the training, validation, and test sets are respectively 2416, 806, and 806.

\textbf{flchain}.
This publicly available dataset is from \cite{dispenzieri2012use}. It contains half of the data collected during a study of the relationship between serum free light chains\footnote{Proteins found in the blood that are part of the immune system} and mortality. The dataset has 7874 examples, of which 73\% are censored. Each example has 9 variables such as age, sex, serum creatinine concentration, and the diagnosis of monoclonal gammopathy. For this dataset, we model the distribution of time until death. The number of examples in the training, validation, and test sets are respectively 4724, 1575, and 1575.

\textbf{Eye Complication}.
This is a proprietary dataset from a government hospital containing the electronic health records of patients who are afflicted with type-2 diabetes, a leading cause of eye complications (such as diabetic retinopathy and macularedema) that could potentially result in permanent blindness. For this dataset, we model the distribution of time until the onset of an eye complication. This dataset has 4374 patients, of which 80\% are censored. Each patient is associated with 38 demographic and longitudinal clinical features such as age, race, serum creatinine level, fasting plasma glucose level, and smoking status. The number of examples in the training, validation, and test sets are respectively 1873, 1251, and 1250.

\subsection{Evaluation Metrics}

To evaluate our GRADUATE model and the baseline systems in our experiments, we use four metrics, viz., C-index, logrank test, expected calibration error (ECE), and a composite score using both C-index and ECE that we term total score.

C-index~\citep{harrell1982cindex} measures a model's discriminative performance in terms of its ability to correctly rank individuals by their probabilities of event onset. For individuals $i$ and $j$ associated respectively with data examples $(\mathbf{x}_i, t_i, \delta_i=1)$ and $(\mathbf{x}_j, t_j, \delta_j)$, let the event of interest occur at (uncensored) time $t_i$ for individual $i$. Assume that at that time $t_i$, the event has not occurred for individual $j$ nor has the event been censored (i.e., $t_i < t_j$ with $\delta_j \in \{0,1\}$). Then, one would intuitively expect individual $i$'s to have a higher probability of event onset at time $t_i$ than individual $j$ at the same time, or equivalently, $F(t_i|\mathbf{x}_i) > F(t_i|\mathbf{x}_j)$. C-index calculates the proportion of such unique pairs ($i$,$j$) of individuals whose relative ranking is correct according to their cumulative distribution functions. C-index ranges between 0 to 1 (inclusive). Mathematically, it can be written as $C-Index=\frac{\sum_{i,j}\mathbb{I}[t_i<t_j]\mathbb{I}[F(t_i|\mathbf{x}_i)> F(t_i|\mathbf{x}_j)]\delta_i}{\sum_{i,j}\mathbb{I}[t_i<t_j]\delta_i}$.

The logrank test is typically administered to test whether two survival curves are identical (under the null hypothesis that there is no difference between the curves) in medical domains (e.g., \cite{faries2017completion} and \cite{liao2012aspirin}). In our experiments, we compute the ground-truth survival curve for a (sub)population by applying the Kaplan-Meier estimator to test data, and compare it to a predicted survival curve (from our GRADUATE model or a comparison system). If the null hypothesis is not rejected, then we have support that the predicted curve is a good estimate of the ground-truth curve, and we give this metric a score of 1. Conversely, if the null hypothesis is rejected, the predicted curve is a bad estimate of the ground-truth curve, and we assign the metric a score of 0. Admittedly, the burden is placed on disproving the null hypothesis in the logrank test.

ECE~\citep{guo2017calibration} is originally used to measure how well calibrated predictions are in the (binary) classification setting. 
Let $p$ be the proportion of examples that are actually true (i.e., have ground-truth labels $Y=1$) among those for which a binary classifier has assigned a predicted probability $\hat{p}$ of being true. The binary classifier is said to be well calibrated if $\hat{p} \approx p$. More formally, the binary classifier is perfectly calibrated when $\mathbb{P}(Y=1|\hat{P}=p) = p, \forall p \in [0,1]$ , where the left-hand side of the equality is the probability that an example is actually true given that the classifier has assigned it a probability $p$ of being true.
To compute ECE from finite samples, ~\cite{guo2017calibration} group the prediction probabilities of test examples into $M$ bins with the $m^{th}$ bin containing examples with predictions $p \in (\frac{m-1}{M}, \frac{m}{M}]$. They then calculate the absolute difference between the average predicted probability in a bin and the proportion of examples in the bin that is true according to ground-truth. After that, they take a weighted average across all bins to arrive at the final ECE score. If the ECE score is 0, then the binary classifier is perfectly calibrated; the further the ECE score is from 0 (i.e., the closer it is to 1), the worse the calibration. 

In the survival analysis setting, we can view the survival probability, for (sub)population $D$, $S_{\theta,D}(t|\cdot)\defeq\mathbb{E}_{(\mathbf{x},\cdot,\cdot)\sim D} \left[\, S_{\theta}(t|\mathbf{x}) \,\right]$ as the predicted probability for the binary classification task of classifying whether the event of interest would occur after time $t$. By relating the survival probability at time $t$ to a binary classification task, we can now compute an ECE score as before for time $t$. Conceptually, we can obtain a single ECE score by averaging over all $t$.
In practice, like for binary classification with finite samples, we divide the $[0,1]$ probability scale into $M$ equally-sized bins, and assign survival probabilities $S_{\theta,D}(t|\cdot)$ to the bin $m$ to which they belong. (NB: Because the survival curve is a monotonically decreasing function, the discrete times $\{t\}$ associated with the probabilities $\{S_{\theta,D}(t|\cdot)\}$ in a bin are necessarily consecutive, and can be considered to be grouped into a bin $B_m$ of their own.)
Then we average over all bins to obtain the final ECE score. More formally, let $B_m$ be the set of discrete times $t$ whose corresponding survival probabilities $S_{\theta,D}(t|\cdot)$ falls in the interval $I_m=(\frac{m-1}{M},\frac{m}{M}]$. The ECE score is calculated as follows.
\begin{equation*}
S_{KM,D}(B_m)\!=\!\frac{1}{|B_m|}\sum_{t \in B_m} S_{KM,D}(t); \;\;\;\;\;\;\;\;
S_{\theta,D}(B_m)\!=\!\frac{1}{|B_m|}\sum_{t \in B_m} S_{\theta,D}(t|\cdot);
\end{equation*}
\begin{equation*}
ECE \defeq \sum_{m=1}^M \frac{|B_m|}{\tau+1}\left| S_{KM,D}(B_m)-S_{\theta,D}(B_m)\right|
\end{equation*}

where $S_{KM,D}(t)$ is the survival curve obtained by applying the KM estimator to a test set, and provides the ground-truth probabilities;
$S_{\theta,D}(t|\cdot)$ is the survival curve that we are evaluating; and $\tau$ is the final timestep (recall from the Background section that time is discretized as $\{0,1,\ldots,\tau\}$). Note that because the KM survival curve is constructed from the test set, its probability $S_{KM,D}(t)$ gives the proportion of individual in the test set for whom event onset has not occurred at time $t$. Intuitively, the ECE score measures the difference between the gold-standard Kaplan-Meier survival curve and the survival curve under evaluation. The smaller the difference, the better the calibration of the latter curve. Like C-index, the ECE score ranges from 0 to 1 (inclusive).

Because discrimination and calibration are both important aspects of a survival model, we would like a single metric to measure its overall performance on both aspects. To this end, we compute the harmonic mean between C-index and ($1-$ECE), and term this metric \emph{total score}.

\subsection{Baseline Systems and Our GRADUATE Model}

We compare our GRADUATE model against four baselines, viz., 
\linebreak DRSA~\citep{ren2019deep}, X-Cal~\citep{goldstein2020x}, McBoostSurv~\citep{becker2021multicalibration}, and RPS~\citep{kamran2021estimating}. We choose DRSA as a baseline because of its state-of-the-art discriminative performance. This allows us to compare our GRADUATE model against a system that excels in discrimination, and investigate their relative calibration performances.

The X-Cal and RPS baselines use specialized objective functions to train pre-specified deep-learning architectures in order to achieve good calibration performance (albeit at the population level only). They are model-agnostic in the sense that they work with any pre-specified deep-learning architectures. $\mbox{MCBoostSurv}$ is a model-agnostic system that post-processes a pre-specified survival model to ensure multicalibration, i.e., the model's predictions are calibrated for multiple pre-specified subpopulations. (See the Introduction and Related Work sections for more information about these baselines.) Like the three baselines, our GRADUATE model is model-agnostic, and can be used for any learning system such as deep learning ones. To ensure that these models are fairly compared, we use DRSA as the underlying deep-learning architecture for all the baselines and our GRADUATE model. We choose DRSA as the underlying architecture because of its stellar discrimination results. Specifically, DRSA uses an RNN consisting of a one-layer chain of LSTM units, one for each timestep. The hidden state of each LSTM unit is set to have 50 dimensions, and the output from an LSTM unit is fed into a corresponding fully-connected neural network with a sigmoid activation output.

For our GRADUATE model, we specify the subpopulations over which we require it to be well-calibrated. We employ two methods to determine the subpopulations or subgroups. (We use the terms \emph{subpopulation} and \emph{subgroup} interchangeably.) First, we manually select demographic features that are typically considered protected traits, e.g., gender, race, age, and housing type (as a reflection of income). We divided the possible values of a trait into bins, and created subgroups defined by membership in the bins. By defining subgroups based on protected attributes, we can investigate how well our model is calibrated for the subgroups in the interest of \emph{fairness}. For the datasets \texttt{nwtco}, \texttt{flchain} and \texttt{eye complication}, we define 3, 6 and 5 demographic-based subgroups respectively.

Second, to ensure that we have not cherry-picked the aforementioned demographic subgroups and to investigate how well our GRADUATE model performs on subgroups that are not based on demographics, we use an automatic selection procedure. We take the cross-product of the possible values of one categorical feature, and then two categorical features, and then three categorical features, and so forth. Next we form subgroups based on the combinations of values formed from the cross-product. We select those subgroups one at a time in order of decreasing size, disregarding those that are too small (each selected subgroup should have at least 100 examples from the training set) or that overlaps too much with previously selected subgroups (a chosen subgroup cannot have more than 80\% of its instances in common with a previously selected subgroup). In this manner, we choose 8, 18, and 18 subgroups for the \texttt{nwtco}, \texttt{fchain} and \texttt{eye complication} datasets respectively. For each dataset, we also included the trivial ``sub''-group of \emph{all} examples, which allows our model to be calibrated for the entire population.
In total, there are 12, 25, and 24 subgroups for the \texttt{nwtco}, \texttt{flchain} and \texttt{eye complication} datasets respectively. Note that the subgroups can overlap (e.g., an individual can concurrently belong to the ``\texttt{male}'' subgroup and the  ``\texttt{age-35-to-40 and is-smoker}'' subgroup).

The MCBoostSurv baseline also requires subgroups to be specified so that it can post-process and calibrate the predictions from its underlying DRSA model for each of the subgroups. We use the same subgroups for MCBoostSurv as those selected above.

For our GRADUATE model using the $L^2$-norm method, we set the right-hand side of the inequality constraints (the $c_i$'s in Problem \ref{eq:model_3}) to 2\%, 1\% and 1\% for \texttt{nwtco}, \texttt{flchain}, and \texttt{eye complication} respectively (the numbers are picked using validation sets). For our GRADUATE model using the variance-adjusted method, we set the right-hand side of the inequality constraints (the $c_i$'s in Problem \ref{eq:model_3}) to 1.96 for all datasets (corresponding to the Z score at 95\% confidence level). 

For all datasets and models, we set $\tau = 102$, i.e., we discretize time as $\{0,1,\ldots,102\}$. In Algorithm~\ref{alg:bg_primal_dual_1}, we set $\eta=0.01$ and $T=3000$. The model is selected using validation set by considering the number of constraints and C-Index at each iteration. We select the model satisfying the highest number of constraints or highest C-Index (if equal number of constraints satisfied). We have implemented early stopping with a patience parameter of 500. Early stopping terminates the algorithm if the selected model remains the same for 500 consecutive iterations. It is worth noting that the hyperparameters, including $\eta$, $T$, and the patience parameter, are also chosen based on their performance on the validation set, similar to the model selection process, where we consider both the number of constraints satisfied and C-Index.

\subsection{Discrimination and Calibration Results over Entire Population}
\bgroup
\def\arraystretch{1.3}
\footnotesize
\begin{table}[ht]
    \centering
    \footnotesize
\begin{tabular}{|c|>{\centering\arraybackslash}p{1.85cm}|>{\centering\arraybackslash}p{1.85cm}|>{\centering\arraybackslash}p{1.85cm}|>{\centering\arraybackslash}p{1.85cm}|} 
\cline{1-5}
 
  & \multicolumn{4}{c|}{\normalsize{\textbf{nwtco}}}\rule{0pt}{3ex} \\ [2pt] \cline{2-5}
 & 
  C-Index $\uparrow$ &
  \begin{tabular}[c]{@{}c@{}}Logrank \\Test\end{tabular} $\uparrow$ &
  ECE $\downarrow$ &
  Total Score $\uparrow$ \\
  \hline
\multicolumn{1}{|c|}{DRSA}        
            & \begin{tabular}[c]{@{}c@{}} \underline{0.684}\\\scriptsize($\pm$0.008) \end{tabular} 
            & 0 
            & \begin{tabular}[c]{@{}c@{}}0.621 \\ \scriptsize($\pm$ 0.038)	\end{tabular}
            & \begin{tabular}[c]{@{}c@{}}0.486 \\ \scriptsize($\pm$ 0.028)	\end{tabular}
            \\ \hline
\multicolumn{1}{|c|}{X-Cal}   
            & \begin{tabular}[c]{@{}c@{}} \textbf{\textbf{0.688**}}\\\scriptsize($\pm$0.011) \end{tabular} 
            & \underline{4} 
            & \begin{tabular}[c]{@{}c@{}} 0.469 \\ \scriptsize($\pm$ 0.111)	\end{tabular}
            & \begin{tabular}[c]{@{}c@{}} 0.594 \\ \scriptsize($\pm$ 0.065)	\end{tabular}
            \\ \hline  
\multicolumn{1}{|c|}{MCBoostSurv}
            & \begin{tabular}[c]{@{}c@{}} 0.650\\\scriptsize($\pm$0.018) \end{tabular} 
            & 0 
            & \begin{tabular}[c]{@{}c@{}} 0.551\\\scriptsize($\pm$0.023) \end{tabular} 
            & \begin{tabular}[c]{@{}c@{}} 0.530\\\scriptsize($\pm$0.017) \end{tabular} 
            \\ \hline
\multicolumn{1}{|c|}{RPS} 
            & \begin{tabular}[c]{@{}c@{}} 0.653\\\scriptsize($\pm$0.015) \end{tabular} 
            & 0 
            & \begin{tabular}[c]{@{}c@{}} 0.315\\\scriptsize($\pm$0.128) \end{tabular} 
            & \begin{tabular}[c]{@{}c@{}} 0.665\\\scriptsize($\pm$0.060) \end{tabular}  
            \\ \hline
\multicolumn{1}{|c|}{\begin{tabular}[c]{@{}c@{}}GRADUATE\\($L^2\text{-Norm}$)\end{tabular}}    
            & \begin{tabular}[c]{@{}c@{}} 0.683\\\scriptsize($\pm$0.005) \end{tabular} 
            & \textbf{21**} 
            & \begin{tabular}[c]{@{}c@{}} \underline{0.142}\\\scriptsize($\pm$0.153) \end{tabular}
            & \begin{tabular}[c]{@{}c@{}} \underline{0.755}\\\scriptsize($\pm$0.065) \end{tabular} 
            \\ \hline
\multicolumn{1}{|c|}{\begin{tabular}[c]{@{}c@{}}GRADUATE\\(Var-Adjusted) \end{tabular} }   
            & \begin{tabular}[c]{@{}c@{}} 0.678\\\scriptsize($\pm$0.008) \end{tabular} 
            & \textbf{21**} 
            & \begin{tabular}[c]{@{}c@{}} \textbf{0.059**}\\\scriptsize($\pm$0.124) \end{tabular} 
            & \begin{tabular}[c]{@{}c@{}} \textbf{0.784**}\\\scriptsize($\pm$ 0.055) \end{tabular}  
            \\ \hline
\multicolumn{5}{|c|}{\footnotesize{{* and ** indicate statistical significance at the 95\% and 99\% level respectively.}}} \rule{0pt}{3ex} \\ [1pt]
\hline
\end{tabular}
\vspace{0.3cm}
    \caption{C-Index, Logrank Test, ECE, and Total Score (with standard deviations where applicable).}
	\label{tab:GRADUATE_pop_result_nwtco}
\end{table}
\egroup
\bgroup
\def\arraystretch{1.3}
\footnotesize
\begin{table}[ht]
    \centering
    \footnotesize
\begin{tabular}{|c|>{\centering\arraybackslash}p{1.85cm}|>{\centering\arraybackslash}p{1.85cm}|>{\centering\arraybackslash}p{1.85cm}|>{\centering\arraybackslash}p{1.85cm}|} 
\cline{1-5}
 
  & \multicolumn{4}{c|}{\normalsize{\textbf{flchain}}}\rule{0pt}{3ex} \\ [2pt] \cline{2-5}
 & 
  C-Index $\uparrow$ &
  \begin{tabular}[c]{@{}c@{}}Logrank \\Test\end{tabular} $\uparrow$ &
  ECE $\downarrow$ &
  Total Score $\uparrow$ \\
  \hline
\multicolumn{1}{|c|}{DRSA}        
            & \begin{tabular}[c]{@{}c@{}} \textbf{0.934**}\\\scriptsize($\pm$0.005) \end{tabular} 
            & \underline{7} 
            & \begin{tabular}[c]{@{}c@{}} 0.0048 \\ \scriptsize($\pm$ 0.0004)	\end{tabular}
            & \begin{tabular}[c]{@{}c@{}} \textbf{0.961} \\ \scriptsize($\pm$ 0.003)	\end{tabular}
            \\ \hline
\multicolumn{1}{|c|}{X-Cal}   
            & \begin{tabular}[c]{@{}c@{}} 0.928\\\scriptsize($\pm$0.004) \end{tabular} 
            & 0 
            & \begin{tabular}[c]{@{}c@{}} 0.0035 \\ \scriptsize($\pm$ 0.0011)	\end{tabular}
            & \begin{tabular}[c]{@{}c@{}} 0.960 \\ \scriptsize($\pm$ 0.000)	\end{tabular}
            \\ \hline  
\multicolumn{1}{|c|}{MCBoostSurv}
            & \begin{tabular}[c]{@{}c@{}} \underline{0.931}\\\scriptsize($\pm$0.003) \end{tabular} 
            & 2 
            & \begin{tabular}[c]{@{}c@{}} 0.0051\\\scriptsize($\pm$0.0003) \end{tabular} 
            & \begin{tabular}[c]{@{}c@{}} 0.960\\\scriptsize($\pm$0.000) \end{tabular} 
            \\ \hline
\multicolumn{1}{|c|}{RPS} 
            & \begin{tabular}[c]{@{}c@{}} 0.930\\\scriptsize($\pm$0.000) \end{tabular} 
            & 0 
            & \begin{tabular}[c]{@{}c@{}} 0.0066\\\scriptsize($\pm$0.0035) \end{tabular} 
            & \begin{tabular}[c]{@{}c@{}} 0.960 \\\scriptsize($\pm$0.000) \end{tabular}  
            \\ \hline
\multicolumn{1}{|c|}{\begin{tabular}[c]{@{}c@{}}GRADUATE\\($L^2\text{-Norm}$)\end{tabular}}    
            & \begin{tabular}[c]{@{}c@{}} 0.926\\\scriptsize($\pm$0.005) \end{tabular} 
            & \textbf{10} 
            & \begin{tabular}[c]{@{}c@{}} \textbf{0.0018**}\\\scriptsize($\pm$0.0004) \end{tabular} 
            & \begin{tabular}[c]{@{}c@{}} 0.960\\\scriptsize($\pm$0.000) \end{tabular} 
            \\ \hline
\multicolumn{1}{|c|}{\begin{tabular}[c]{@{}c@{}}GRADUATE\\(Var-Adjusted) \end{tabular} }   
            & \begin{tabular}[c]{@{}c@{}} 0.920\\\scriptsize($\pm$0.000) \end{tabular} 
            & \textbf{10} 
            & \begin{tabular}[c]{@{}c@{}} \underline{0.0030}\\\scriptsize($\pm$0.0000) \end{tabular} 
            & \begin{tabular}[c]{@{}c@{}} 0.960\\\scriptsize($\pm$0.000) \end{tabular} 
            \\ \hline
\multicolumn{5}{|c|}{\footnotesize{{* and ** indicate statistical significance at the 95\% and 99\% level respectively.}}} \rule{0pt}{3ex} \\ [1pt]
\hline
\end{tabular}
\vspace{0.3cm}
    \caption{C-Index, Logrank Test, ECE, and Total Score (with standard deviations where applicable).}
	\label{tab:GRADUATE_pop_result_flchain}
\end{table}
\egroup
\bgroup
\def\arraystretch{1.3}
\footnotesize
\begin{table}[ht]
    \centering
    \footnotesize
\begin{tabular}{|c|>{\centering\arraybackslash}p{1.85cm}|>{\centering\arraybackslash}p{1.85cm}|>{\centering\arraybackslash}p{1.85cm}|>{\centering\arraybackslash}p{1.85cm}|} 
\cline{1-5}
 
  & \multicolumn{4}{c|}{\normalsize{\textbf{Eye Complication}}}\rule{0pt}{3ex} \\ [2pt] \cline{2-5}
 & 
  C-Index $\uparrow$ &
  \begin{tabular}[c]{@{}c@{}}Logrank \\Test\end{tabular} $\uparrow$ &
  ECE $\downarrow$ &
  Total Score $\uparrow$ \\
  \hline
\multicolumn{1}{|c|}{DRSA}        
            & \begin{tabular}[c]{@{}c@{}} \textbf{0.715**}\\\scriptsize($\pm$0.006) \end{tabular} 
            & 0 
            & \begin{tabular}[c]{@{}c@{}}0.358 \\ \scriptsize($\pm$ 0.076)	\end{tabular}
            & \begin{tabular}[c]{@{}c@{}}0.674 \\ \scriptsize($\pm$ 0.042)	\end{tabular}
            \\ \hline
\multicolumn{1}{|c|}{X-Cal}   
            & \begin{tabular}[c]{@{}c@{}} 0.671\\\scriptsize($\pm$0.022) \end{tabular} 
            & 2 
            & \begin{tabular}[c]{@{}c@{}} 0.233 \\ \scriptsize($\pm$ 0.123)	\end{tabular}
            & \begin{tabular}[c]{@{}c@{}} 0.712 \\ \scriptsize($\pm$ 0.065)	\end{tabular}
            \\ \hline  
\multicolumn{1}{|c|}{MCBoostSurv}
            & \begin{tabular}[c]{@{}c@{}} \underline{0.677}\\\scriptsize($\pm$0.011) \end{tabular} 
            & 0 
            & \begin{tabular}[c]{@{}c@{}} 0.42\\\scriptsize($\pm$0.024) \end{tabular} 
            & \begin{tabular}[c]{@{}c@{}} 0.624\\\scriptsize($\pm$ 0.015) \end{tabular} 
            \\ \hline
\multicolumn{1}{|c|}{RPS} 
            & \begin{tabular}[c]{@{}c@{}} 0.639\\\scriptsize($\pm$0.011) \end{tabular} 
            & 0 
            & \begin{tabular}[c]{@{}c@{}} 0.353\\\scriptsize($\pm$0.045) \end{tabular} 
            & \begin{tabular}[c]{@{}c@{}} 0.641\\\scriptsize($\pm$ 0.019) \end{tabular} 
            \\ \hline
\multicolumn{1}{|c|}{\begin{tabular}[c]{@{}c@{}}GRADUATE\\($L^2\text{-Norm}$)\end{tabular}}    
                & \begin{tabular}[c]{@{}c@{}} 0.664\\\scriptsize($\pm$0.018) \end{tabular} 
                & \textbf{15**} 
                & \begin{tabular}[c]{@{}c@{}} \underline{0.082}\\\scriptsize($\pm$0.064) \end{tabular} 
                & \begin{tabular}[c]{@{}c@{}} \textbf{0.769**}\\\scriptsize($\pm$ 0.026) \end{tabular} 
                \\ \hline
\multicolumn{1}{|c|}{\begin{tabular}[c]{@{}c@{}}GRADUATE\\(Var-Adjusted) \end{tabular} }   
                & \begin{tabular}[c]{@{}c@{}} 0.624\\\scriptsize($\pm$0.015) \end{tabular} 
                & \textbf{15**} 
                & \begin{tabular}[c]{@{}c@{}} \textbf{0.057**}\\\scriptsize($\pm$0.076) \end{tabular} 
                & \begin{tabular}[c]{@{}c@{}} \underline{0.749}\\\scriptsize($\pm$ 0.03) \end{tabular} 
                \\ \hline
\multicolumn{5}{|c|}{\footnotesize{{* and ** indicate statistical significance at the 95\% and 99\% level respectively.}}} \rule{0pt}{3ex} \\ [1pt]
\hline
\end{tabular}
\vspace{0.3cm}
    \caption{C-Index, Logrank Test, ECE, and Total Score (with standard deviations where applicable).}
	\label{tab:GRADUATE_pop_result_eyecomp}
\end{table}
\egroup

Tables \ref{tab:GRADUATE_pop_result_nwtco}, \ref{tab:GRADUATE_pop_result_flchain}, and \ref{tab:GRADUATE_pop_result_eyecomp} report the performances of our GRADUATE model and the baselines on three datasets. The results are for the entire population in each respective test set. 
We repeat our experiments 21, 10, and 15 times for the \texttt{nwtco}, \texttt{flchain}, and \texttt{eye complication} datasets respectively, each time starting from a different random initialization of parameters when training the models. The results in the tables are the average scores over the experiments (with the corresponding standard deviations in parentheses where applicable). However, for the logrank test, each reported result is the total number of times a model passes the test in the repeated experiments. For each metric, we  compare the best performing model (bolded in Tables \ref{tab:GRADUATE_pop_result_nwtco}, \ref{tab:GRADUATE_pop_result_flchain}, and \ref{tab:GRADUATE_pop_result_eyecomp}) with the second best model (underlined in tables) to determine whether their differences are statistically significant (using paired t-tests).  

In terms of C-index (which measures discrimination performance), DRSA is consistently the best or (close) second best performer, and is the best system on average. This is expected because DRSA's objective function is primed for good discriminative performance. Our GRADUATE models have comparable C-index scores as DRSA on \texttt{nwtco} and \texttt{flchain}, but loses to DRSA on \texttt{eye complication} by about 8\%-15\%. In terms of calibration however, DRSA does poorly, losing to our GRADUATE models across all datasets in terms of logrank test and ECE scores.

In terms of calibration performance, as measured by the logrank test and ECE scores, we see that the two variants of our GRADUATE model have stellar results, consistently being both the best and second best systems across all datasets. On the \texttt{nwtco} and \texttt{eye complication} datasets, our GRADUATE (variance-adjusted) model outperforms the best baseline system by more than 400\% and 500\% respectively in terms of ECE scores. On \texttt{flchain}, our GRADUATE model, though still the best, outperforms the best baseline more modestly by 9\%. It is also noteworthy that both our GRADUATE models attain the maximum scores possible for the logrank test across all datasets. These good results attest to the strength of our GRADUATE system for calibration at the population level. 

Next we examine the total score (i.e., the harmonic mean of C-index and (1-ECE)) to determine the overall performance of the systems in terms of both discrimination and calibration. On the \texttt{nwtco} and \texttt{eye complication} datasets, our GRADUATE models are both the best and second best performers. The better of our two GRADUATE models outperforms the best baseline by about 7\% and 15\% on the \texttt{nwtco} and \texttt{eye complication} datasets respectively. On the \texttt{flchain} dataset, all systems perform similarly in terms of total score. On average across the datasets, our GRADUATE models are the best performers in terms of total score, showing that they can achieve a good balance between discrimination and calibration.

\begin{figure}[h]
	\begin{center}
	\includegraphics[scale = 0.7]{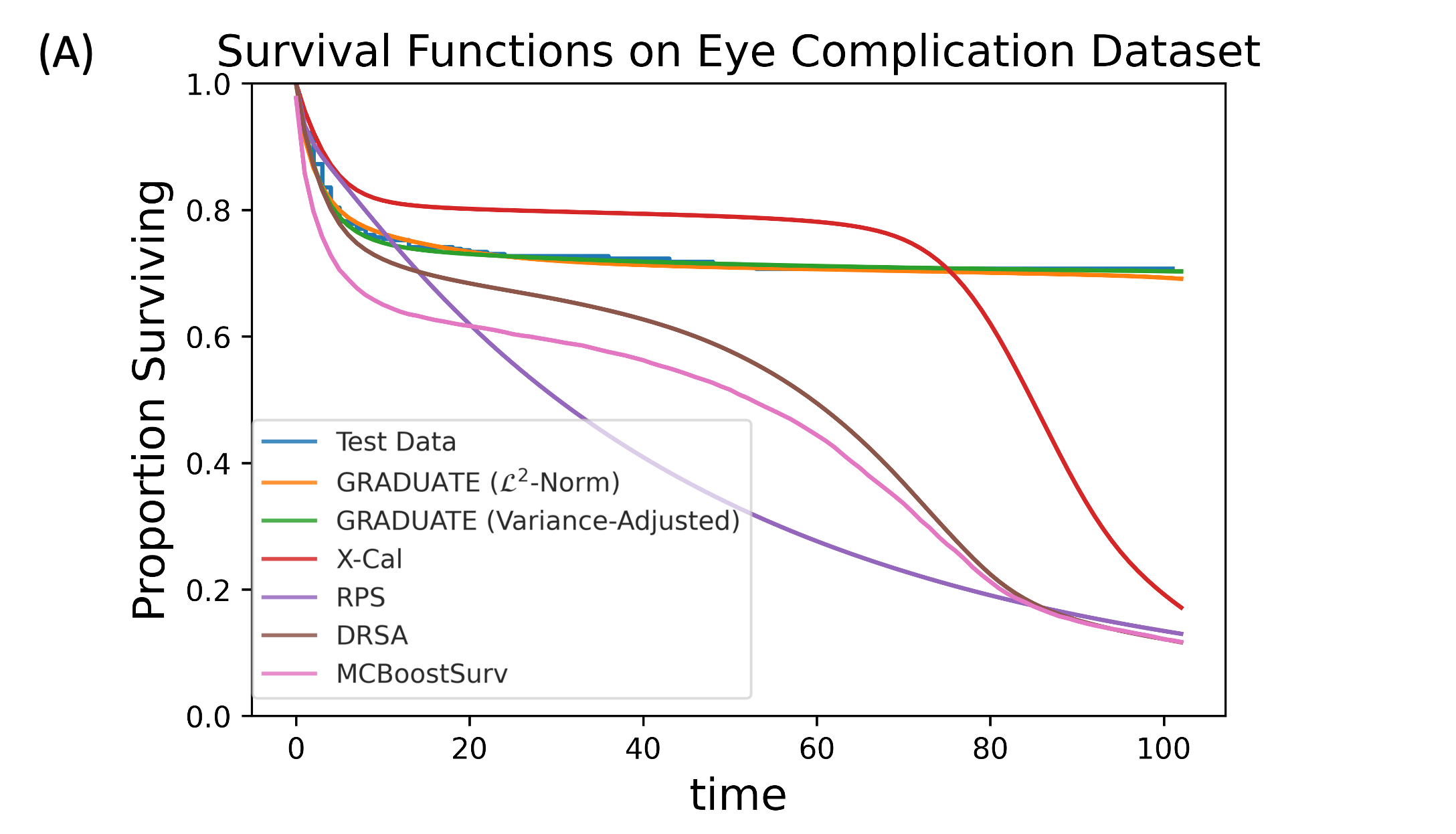} \\ [-4pt]
	\includegraphics[scale = 0.7]{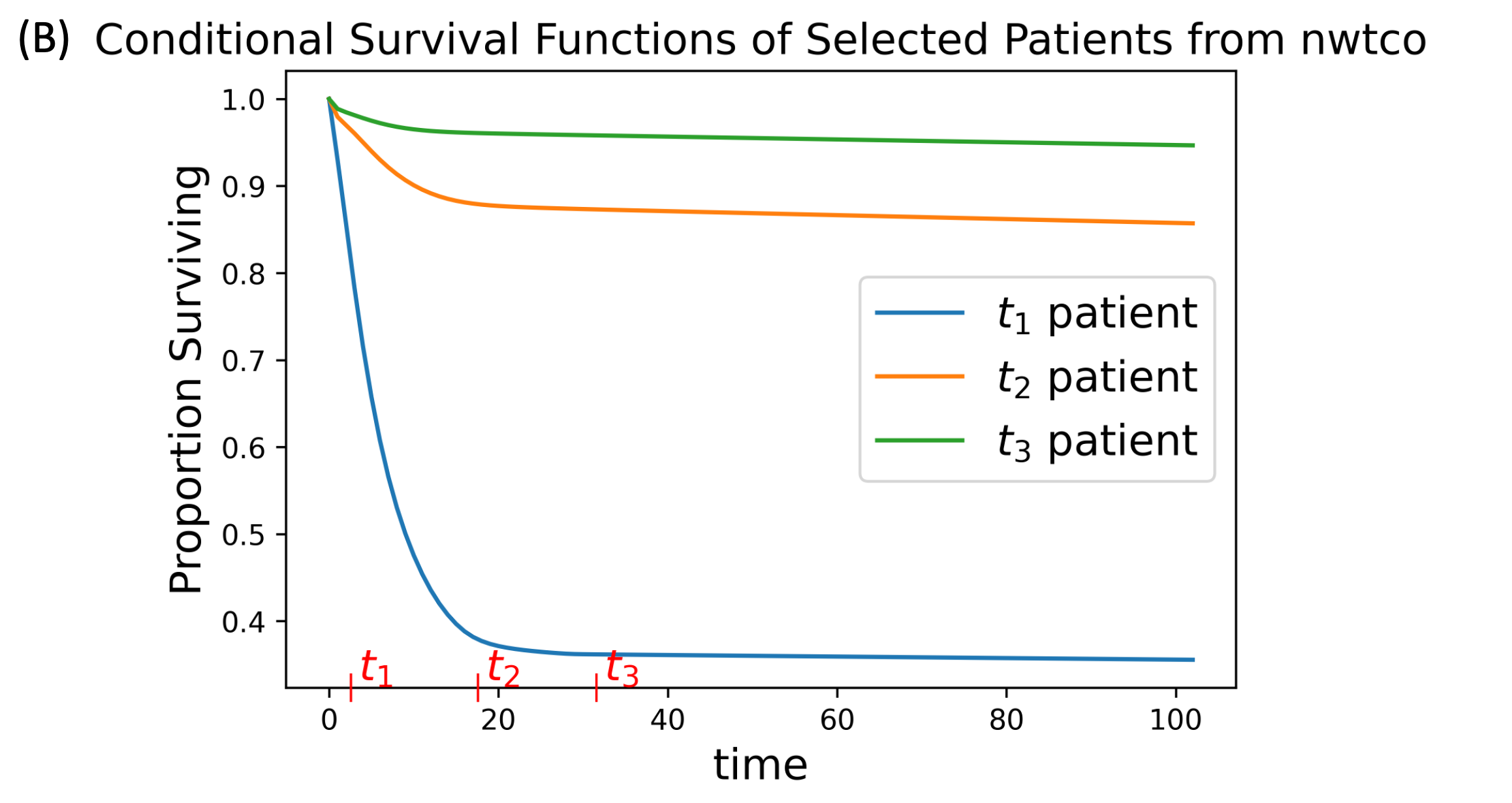}
	\end{center}
	\caption{Illustration of Predicted Survival Functions.}\label{fig:eyecomp_pop_cond_survprobs} 
\end{figure}

As a sanity check of the models' calibration results and as a qualitative evaluation, we visualize the results by plotting the ground-truth Kaplan-Meier (KM) survival curve on the test set, and the marginal survival curves produced by the baseline models and our GRADUATE models. To obtain the marginal survival curve from a model, we plot the individualized survival curve $S(t|\mathbf{x}_i)$ for each individual $i$ with feature $\mathbf{x}_i$, and average the curves for all individuals at every timepoint.
(Figure \ref{fig:eyecomp_pop_cond_survprobs}(B) shows the individualized survival curves from our GRADUATE (variance-adjusted) model for individuals 1, 2, and 3 with ground-truth event onset times $t_1 \!<\! t_2 \!<\! t_3$ before they are averaged together to give the marginal survival curve. From that figure, we also see that our model has good discrimination in correctly ranking the three individuals, e.g., $S(t_2|\mathbf{x_1}) < S(t_2|\mathbf{x_2}) < S(t_2|\mathbf{x_3})$.) 

\begin{figure}[h]
	\begin{center}
	\includegraphics[scale = 0.7]{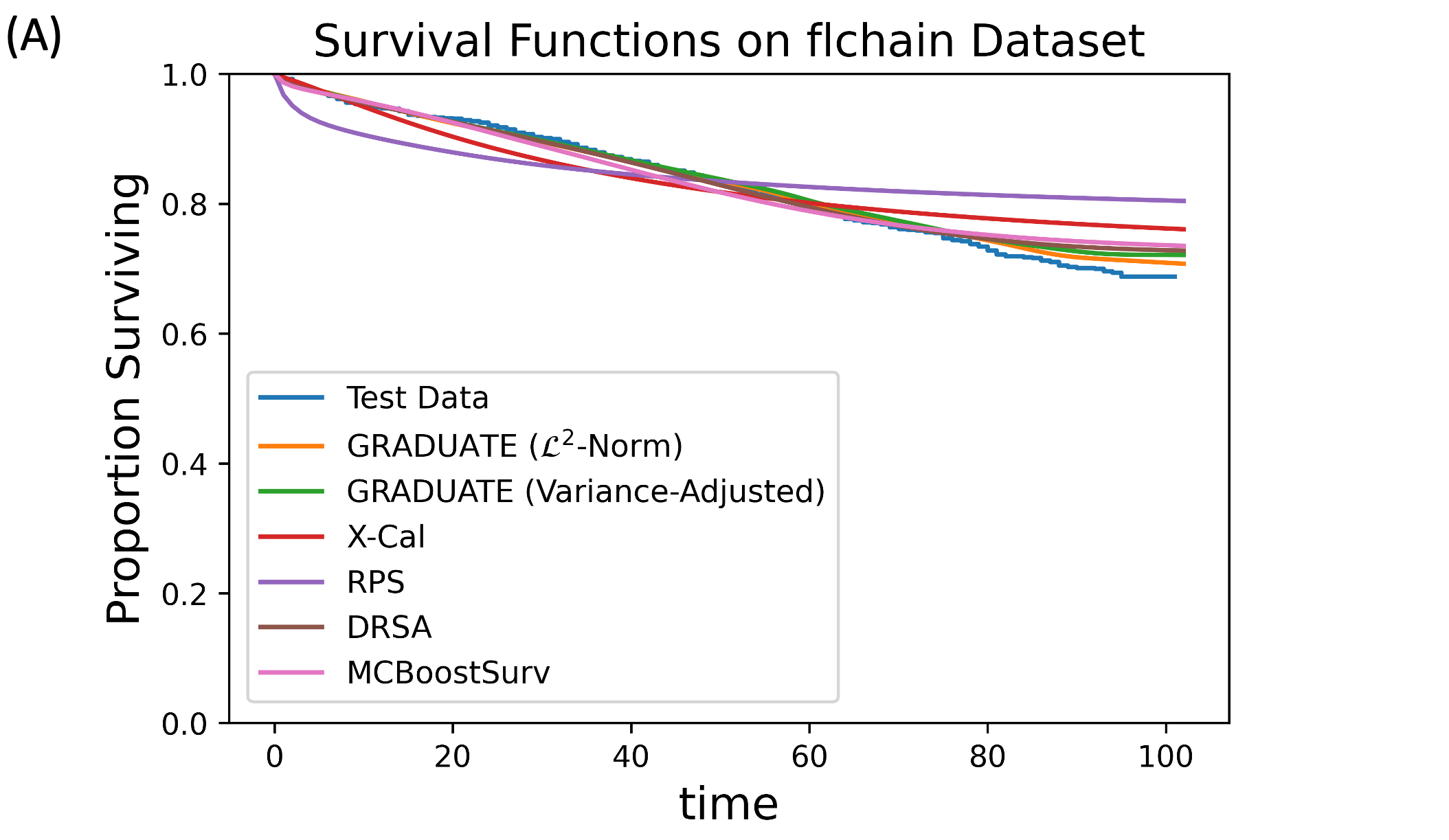} \\ [-4pt]
	\includegraphics[scale = 0.7]{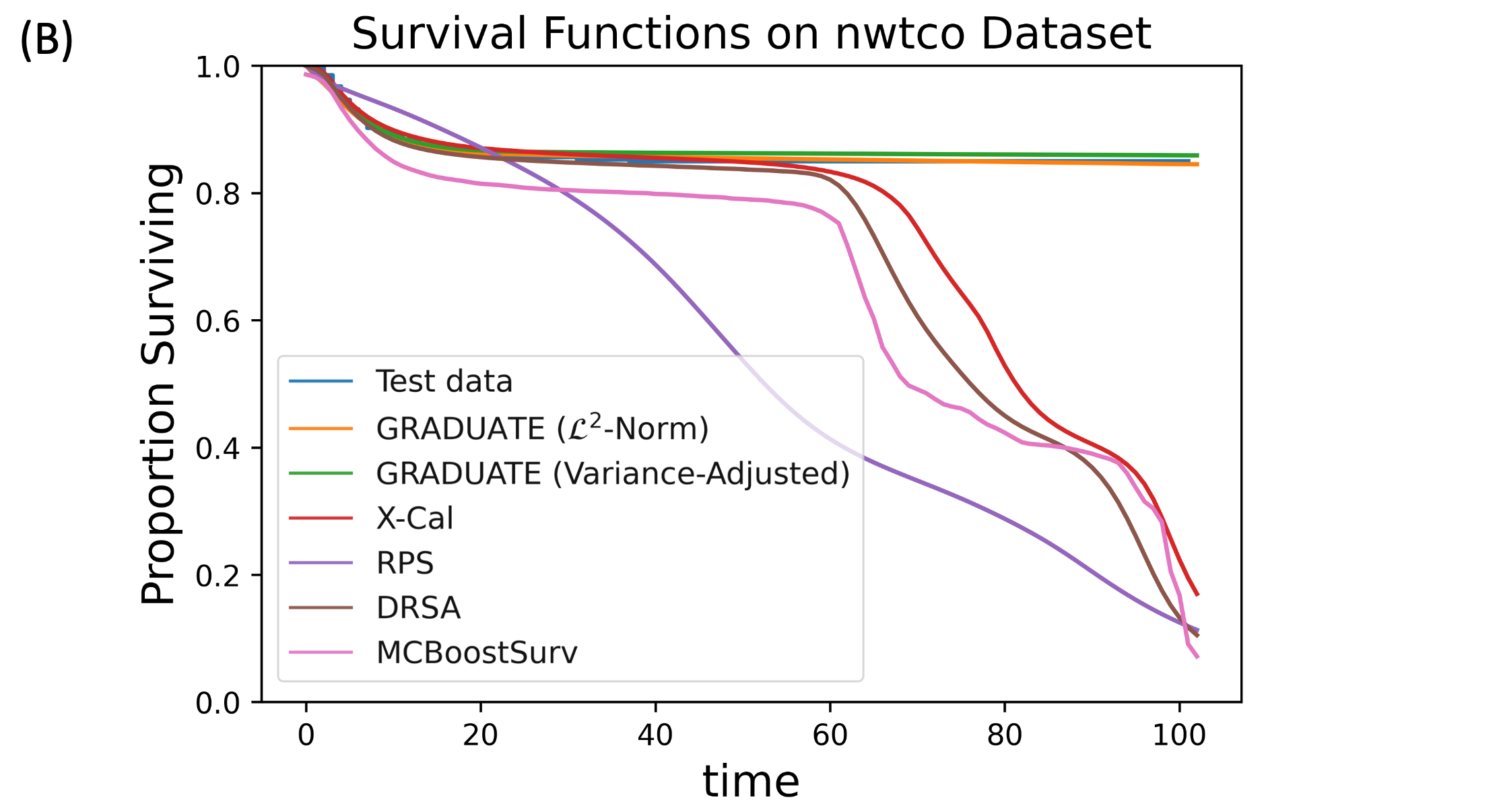}
	\end{center}
	\caption{Illustration of Predicted Survival Functions.}\label{fig:flchain_pop_nwtco_pop}
\end{figure}

In Figure \ref{fig:eyecomp_pop_cond_survprobs}(A) for the \texttt{eye complication} dataset, we see that the gold and green marginal survival curves due to our GRADUATE models closely trace the blue ground-truth KM curve (they are so close that the blue curve is partially hidden beneath the gold and green ones). In contrast, the marginal survival curves of the baseline systems differ significantly from the blue KM curve. These visualizations agree with the results in Table \ref{tab:GRADUATE_pop_result_eyecomp} in which our GRADUATE models achieve the best calibration results (in terms of logrank test and ECE scores). In Figure \ref{fig:flchain_pop_nwtco_pop}, the plots of the marginal survival curves for the \texttt{flchain} and \texttt{nwtco} also agree with the calibration results in Tables \ref{tab:GRADUATE_pop_result_nwtco}, \ref{tab:GRADUATE_pop_result_flchain}, and \ref{tab:GRADUATE_pop_result_eyecomp}.

\subsection{Calibration and Discrimination Performances over Subpopulations}

Through Tables \ref{tab:GRADUATE_pop_result_nwtco} to \ref{tab:GRADUATE_pop_result_eyecomp}, we have ascertained that our GRADUATE models attain the best calibration results and the best balance between discrimination and calibration over the \emph{entire population}. The question we seek to answer now is whether our GRADUATE models have obtained such stellar marginal calibration and balance by compromising their calibrations on subpopulations, or whether they have preserved good calibrations at the subpopulation level too.

To answer this question, we choose the $L^2\text{-norm}$ variant from the two variants of our GRADUATE model, and only evaluate it against baselines to simplify the comparisons (note that the results for both variants are very close in Tables \ref{tab:GRADUATE_pop_result_nwtco} to \ref{tab:GRADUATE_pop_result_eyecomp}). We compare this GRADUATE model against each of the four baselines on all test sets in terms of the calibration metrics, logrank test and ECE.
We pair the GRADUATE model with each baseline, and conduct a paired t-test on each subpopulation of a dataset to decide which system in the pair performs statistically significantly better. We report the outcome as three numbers $X$-$Y$-$Z$, where $X$ is number of times GRADUATE statistically significantly outperforms the baseline, $Y$ is similarly the number of times the baseline outperforms GRADUATE, and $Z$ is the number of times when neither is statistically significantly better than the other. For our GRADUATE model to perform better than the baseline, $X$ has to be larger than $Y$. (Recall that we repeat our experiments 21, 10, and 15 times for the \texttt{nwtco}, \texttt{flchain}, and \texttt{eye complication} datasets respectively so that we can perform the paired t-tests.)

Tables \ref{tab:GRADUATE_subpop_result_nwtco}, \ref{tab:GRADUATE_subpop_result_flchain}, and \ref{tab:GRADUATE_subpop_result_eyecomp} present the outcomes of the comparisons. From the logrank test and ECE results, we see that our GRADUATE model has successfully achieve good calibrations in subpopulations too, and unequivocally outperforms all baselines, frequently beating the baselines on a (large) majority of subpopulations.  For example, on the ECE score for the \texttt{nwtco} dataset, GRADUATE outperforms the RPS baseline on 10 out of 12 subpopulations; in contrast, RPS only outperforms GRADUATE on only one subpopulation. An outlier to this trend occurs when comparing GRADUATE against DRSA on the \texttt{flchain} dataset in terms of logrank test score, but even in this case GRADUATE wins slightly 5 to 3. 

In terms of discrimination performance as measured by C-index, GRADUATE does not do as well, frequently losing against the baselines. However, when considering the aggregate of both discrimination and calibration performances as reflected by the total score, we see that GRADUATE again beats its comparison baselines by a large margin for the \texttt{nwtco} and \texttt{eye complication} datasets. This attests to GRADUATE's ability to achieve an optimal balance between discrimination and calibration, both of which are important aspects for survival analysis. On the \texttt{flchain} dataset, GRADUATE does not perform as well in terms of total score. We observe that even though GRADUATE does relative better on the ECE score on this dataset, the scores are typically very small for all systems (like the ECE scores for \texttt{flchain} in Table \ref{tab:GRADUATE_pop_result_flchain}). Consequently, the (1-ECE) component of total score is similarly close to 1.0 for all systems, and thus the discriminative C-index component has a dominating effect on the total score. This results in poorer performance in terms of total score for GRADUATE.   

\begin{table}[h]
    \centering
\begin{tabular}{|c|>{\centering\arraybackslash}p{2cm}|>{\centering\arraybackslash}p{2cm}|>{\centering\arraybackslash}p{2cm}|>{\centering\arraybackslash}p{2cm}|} 
\cline{1-5}
  \multicolumn{5}{|c|}{\normalsize{\textbf{nwtco}}}\rule{0pt}{3ex} \\ [2pt] \cline{1-5}
\hline
& DRSA & X-Cal & \footnotesize{MCBoostSurv} & RPS\rule{0pt}{3ex}   \\
[2pt] \hline
Logrank Test \rule{0pt}{2.5ex} & 10-1-1 & 12-0-0 & 10-1-1 & 10-0-2  \\[2pt] \hline
ECE \rule{0pt}{2.5ex} & 12-0-0 & 12-0-0 & 10-0-2 & 10-1-1  \\[2pt] \hline
C-Index \rule{0pt}{2.5ex} & 2-4-6 & 0-5-7 & 11-0-1 &  7-2-3 \\[2pt] \hline
Total Score \rule{0pt}{2.5ex} & 12-0-0 & 11-0-1  & 12-0-0 & 10-1-1  \\[2pt] \hline
\multicolumn{5}{|c|}{\footnotesize{{Results denote ``\#wins-\#losses-\#draws" by GRADUATE ($L^2$-Norm) against}}} \rule{0pt}{3ex} \\ [0.001pt]
\multicolumn{5}{|c|}{\footnotesize{{other models (using paired t-tests).}}} \rule{0pt}{3ex} \\ [1pt] \hline
\end{tabular}
\vspace{0.3cm}
    \caption{Comparisons of Logrank Test, ECE, C-Index, and Total Score for 12 Subpopulations.}
	\label{tab:GRADUATE_subpop_result_nwtco}
\end{table}
\begin{table}[h]
    \centering
\begin{tabular}{|c|>{\centering\arraybackslash}p{2cm}|>{\centering\arraybackslash}p{2cm}|>{\centering\arraybackslash}p{2cm}|>{\centering\arraybackslash}p{2cm}|} 
\cline{1-5}
  \multicolumn{5}{|c|}{\normalsize{\textbf{flchain}}}\rule{0pt}{3ex} \\ [2pt] \cline{1-5}
\hline
& DRSA & X-Cal & \footnotesize{MCBoostSurv} & RPS\rule{0pt}{3ex}   \\
[2pt] \hline
Logrank Test \rule{0pt}{2.5ex} & 5-3-17  & 10-2-13	 & 11-8-6 & 14-2-9  \\[2pt] \hline
ECE \rule{0pt}{2.5ex} & 15-6-4 & 16-4-5& 16-6-3 & 17-3-5 \\[2pt] \hline
C-Index \rule{0pt}{2.5ex} & 1-18-6  & 3-11-11  & 1-17-7  &  0-16-9\\[2pt] \hline
Total Score \rule{0pt}{2.5ex}& 1-18-6 & 4-8-13 & 1-15-9  &  2-10-13  \\[2pt] \hline
\multicolumn{5}{|c|}{\footnotesize{{Results denote ``\#wins-\#losses-\#draws" by GRADUATE ($L^2$-Norm) against}}} \rule{0pt}{3ex} \\ [0.001pt]
\multicolumn{5}{|c|}{\footnotesize{{other models (using paired t-tests).}}} \rule{0pt}{3ex} \\ [1pt] \hline
\end{tabular}
\vspace{0.3cm}
    \caption{Comparisons of Logrank Test, ECE, C-Index, and Total Score for 25 Subpopulations.}
	\label{tab:GRADUATE_subpop_result_flchain}
\end{table}
\begin{table}[h]
    \centering
\begin{tabular}{|c|>{\centering\arraybackslash}p{2cm}|>{\centering\arraybackslash}p{2cm}|>{\centering\arraybackslash}p{2cm}|>{\centering\arraybackslash}p{2cm}|} 
\cline{1-5}
  \multicolumn{5}{|c|}{\normalsize{\textbf{Eye Complication}}}\rule{0pt}{3ex} \\ [2pt] \cline{1-5}
\hline
& DRSA & X-Cal & \footnotesize{MCBoostSurv} & RPS\rule{0pt}{3ex}   \\
[2pt] \hline
Logrank Test \rule{0pt}{2.5ex} & 18-0-6  & 18-1-5	 & 16-0-8 & 22-0-2  \\[2pt] \hline
ECE \rule{0pt}{2.5ex} & 17-4-1 & 15-4-5& 17-3-4 & 16-3-5 \\[2pt] \hline
C-Index \rule{0pt}{2.5ex} & 3-17-4  & 2-8-14  & 5-11-8 & 15-1-8 \\[2pt] \hline
Total Score \rule{0pt}{2.5ex} & 11-5-8 & 14-5-5 & 17-4-3  & 16-4-4  \\[2pt] \hline
\multicolumn{5}{|c|}{\footnotesize{{Results denote ``\#wins-\#losses-\#draws" by GRADUATE ($L^2$-Norm) against}}} \rule{0pt}{3ex} \\ [0.001pt]
\multicolumn{5}{|c|}{\footnotesize{{other models (using paired t-tests).}}} \rule{0pt}{3ex} \\ [1pt] \hline
\end{tabular}
\vspace{0.3cm}
    \caption{Comparisons of Logrank Test, ECE, C-Index, and Total Score for 24 Subpopulations.}
	\label{tab:GRADUATE_subpop_result_eyecomp}
\end{table}
\subsection{Discussion of Results} \label{sec:discussion}

From Tables \ref{tab:GRADUATE_pop_result_nwtco}-\ref{tab:GRADUATE_subpop_result_eyecomp}, we see that our GRADUATE model achieves better calibration than the baselines over both population and subpopulations. Now we delve into details about the baselines to explicate their shortcomings, and to elucidate our GRADUATE model's strengths.

Because the baseline X-Cal~\citep{goldstein2020x} builds upon D-Calibration (D-Cal;~\cite{haider2020effective}) by creating a differentiable form of D-Cal's loss function, X-Cal's drawback can be traced to that loss function. Using the probability integral transform~\citep{angus1994pit}, D-Cal knows that when a learned survival function fits a dataset $D=\{x, t_i, \delta_i\}_{i=1}^n$ well (i.e., is well-calibrated), the function will project the samples in $D$ uniformly into the probability range [0,1]. At their cruxes, D-Cal and X-Cal try to capture this notion via the loss function
\begin{equation}\label{eq:dcal}
    DCal_\theta =\sum_{I \in \mathcal{I}}
    \left( \, 
    \mathbb{E}_{(\mathbf{x}_i,t_i,\delta_i) \sim D}\,
    \left[\mathbb{I}[ F_\theta(t_i|\mathbf{x}_i)\in I]\right] \,\,-\, |I| 
    \,\right)^2, 
\end{equation}
for the data containing only uncensored individuals and $\theta$ represents the parameters of D-Cal's model; 
$\mathbb{I}[\cdot]$ is the indicator function; 
the collection $\mathcal{I}$ is selected to contain disjoint, equally-sized, contiguous intervals $I\defeq[a_I,b_I] \subseteq [0, 1]$ that cover the whole interval $[0, 1]$; and 
$F_{\theta}(t_i|\mathbf{x}_i)$ is the individualized cumulative distribution function for an individual $i$ with features $\mathbf{x}_i$. (Recall that $F_{\theta}(t|\mathbf{x}) = 1 - S_{\theta}(t|\mathbf{x})$.) For the data containing both uncensored and censored individuals, Equation \ref{eq:dcal} will be generalized to 
\begin{align*}
    & DCal_\theta =\sum_{I \in \mathcal{I}}
    \left( \, 
    \mathbb{E}_{(\mathbf{x}_i,t_i,\delta_i) \sim D}\,
    [G(\mathbf{x}_i,t_i,\delta_i,I)] \,\,-\, |I| 
    \,\right)^2,
\end{align*}
\begin{align*}
    &\textrm{where}\quad G(\mathbf{x}_i,t_i,\delta_i,I)  \defeq\delta_i\mathbb{I}\left[ F_\theta(t_i|\mathbf{x}_i)\in I\right] \\
    &+(1-\delta_i)\left( \frac{\left(b_I-F_\theta(t_i|\mathbf{x}_i)\right)\mathbb{I}\left[ F_\theta(t_i|\mathbf{x}_i)\in I\right]}{1-F_\theta(t_i|\mathbf{x}_i)}+\frac{\left(b_I-a_I\right)\mathbb{I}\left[ F_\theta(t_i|\mathbf{x}_i)<a_I\right]}{F_\theta(t_i|\mathbf{x}_i)}\right)
\end{align*}
Upon inspecting Equation~\ref{eq:dcal}, we see that each individual $i$ enters the sum exactly once via $F_{\theta}(t_i|\mathbf{x}_i) \in I$. Thus its individualized cumulative distribution function $F_{\theta}(t_i|\mathbf{x}_i)$ (or equivalently its survival function $S_{\theta}(t_i|\mathbf{x}_i)$) is only modeled for time $t_i$. This means that for all $t \neq t_i$, $F_{\theta}(t|\mathbf{x}_i)$ is not factored into the loss function, and can thus take on values that are detrimental to calibration. 


The baseline MCBoostSurv~\citep{becker2021multicalibration} is a post-hoc system that post-processes a pre-trained survival analysis model to calibrate it. Because of its post-hoc nature, MCBoostSurv decouples the learning of a good discriminative model from the search for a well-calibrated one. This prevents it from simultaneously finding a good balance between both discrimination and calibration. In addition, by being constrained to using a (relatively small) validation dataset for post-hoc calibration, it fails to utilize the entire training dataset to better inform its calibration decisions. Aside from these shortcomings, MCBoostSurv suffers from a weakness in its loss function that is built upon the Brier score~\citep{brier1951verification}. The Brier score at time $t$ is given by 
\begin{align}
    Brier_\theta(t)&=\frac{1}{|D|}\sum_{(\mathbf{x}_i,t_i,\delta_i) \in D}
    \mathbb{I}[t_i\!\leq\! t,\delta_i\!=\!1] \,
    \frac{\left(0-S_\theta(t|\mathbf{x}_i)\right)^2}{\hat{G}(t_i)} \nonumber\\
    &\,+\, \mathbb{I}[t_i\!>\!t]
    \,\frac{\left( 1-S_\theta(t|\mathbf{x}_i) \right)^2}{\hat{G}(t)}, \label{eq:brier}
\end{align}

where $\theta$ represents the parameters of the MCBoostSurv model and $\hat{G}(t)$ is the estimation of the survival function of the censoring times using the Kaplan-Meier method or $\hat{G}(t)=P(C>t)$ where $C$ is a random variable representing the censoring time occurred. To minimize Equation~\ref{eq:brier}, MCBoostSurv ``pushes" $S_\theta(t|\mathbf{x}_i)$ closer to probability 0 for an uncensored ($\delta_i\!=\!1$) individual $i$ whose event onset time has occurred (i.e., $t_i\!\leq\!t$; first term in the sum of Equation~\ref{eq:brier}). Similarly, MCBoostSurv ``pushes" $S_\theta(t|\mathbf{x}_i)$ closer to probability 1 for an individual $i$ whose event onset time or censoring time $t_i$ comes after $t$ (second term in the sum of Equation~\ref{eq:brier}). The problem occurs for censored individuals $(\delta_i\!=\!0)$ whose censoring time $t_i$ occurs before or at $t$ (i.e., $t_i\leq t$) -- observe that these individuals are not considered in Equation~\ref{eq:brier} at all. This means that MCBoostSurv  neglects to use the information embodied in these individuals to inform its calibration, potentially resulting in poorer calibration performance. To obtain an alternative view of the problem, we replace the first term in the sum of Equation~\ref{eq:brier} with $\mathbb{I}[t_i\!\leq\! t] \left(0-S_\theta(t|\mathbf{x}_i)\right)^2$, and see that in the resulting equivalent loss function, MCBoostSurv has assumed, for censored individuals ($\delta_i=0$) whose $t_i \leq t$, that $S_\theta(t|\mathbf{x}_i) = 0$! This is a strong assumption that could lead to bad calibration if it is violated in real-world data (as is usually the case).

In actuality, MCBoostSurv uses the \emph{integrated} Brier score (IBS; ~\cite{graf1999ibs}) that integrates the Brier score over all time points, and weights the terms in the sum of Equation~\ref{eq:brier} with inverse probability censoring weights. But even so, it still suffers from the aforementioned problem, but now at every time point. 

The baseline RPS~\citep{kamran2021estimating} uses the following loss function
\begin{equation} \label{eq:rps}
    RPS_\theta =\sum_{(\mathbf{x}_i,t_i,\delta_i) \in D}
    \left[ \delta_i \sum_{t=0}^\tau
    \left(S_\theta(t|\mathbf{x}_i)-\mathbb{I}[t<t_i]\right)^2
    + \, (1-\delta_i)\sum_{t=0}^{t_i}\left(S_\theta(t|\mathbf{x}_i)-1 \right)^2 \right],
\end{equation}

where $\theta$ represents the parameters of the RPS model, and $\tau$ is the maximum discrete timestep. RPS shares a similar problem with MCBoostSurv for censored individuals. To explicate the problem, we replace the second term in the outer sum of Equation~\ref{eq:rps} with $(1-\delta_i)\sum_{t=0}^{\tau}\left(S_\theta(t|\mathbf{x}_i)-1 \right)^2$ (note the sum is up to $\tau$ now). In the resultant equivalent loss function, RPS is assuming, for censored individuals with $t \in \{t_i,\ldots,\tau\}$, that $S_\theta(t|\mathbf{x}_i) = 1$!  (NB: This is diametrically opposite to MCBoostSurv's assumption that $S_\theta(t|\mathbf{x}_i) = 0$.) Like for MCBoostSurv, this is a strong assumption that could lead to bad calibration if it is violated in real-world data (as is often the case).

In contrast to the baselines, our GRADUATE system models every individualized survival function $S(t|\mathbf{x})$ at every time point, and avoids making strong limiting assumptions about the individualized survival functions that could impair calibration. To do so, our GRADUATE system requires that, at every timestep $t \in \{0, \ldots, \tau\}$, the marginalization of the predicted individualized survival functions is close to the survival function from the Kaplan-Meier (KM) estimator. This requirement is enforced at both population level and subpopulation level via the constraints in our constrained programming formulation (Equation~\ref{eq:model_3}). These constraints limit each individualized survival function to only choose values at each time point that when aggregated over all individualized survival functions (approximately) yield the KM survival curve. Consequently, our GRADUATE model attains good multicalibration.

\subsubsection{Examples.}
\begin{figure}[h]
	\begin{center}
	\includegraphics[scale = 0.7]{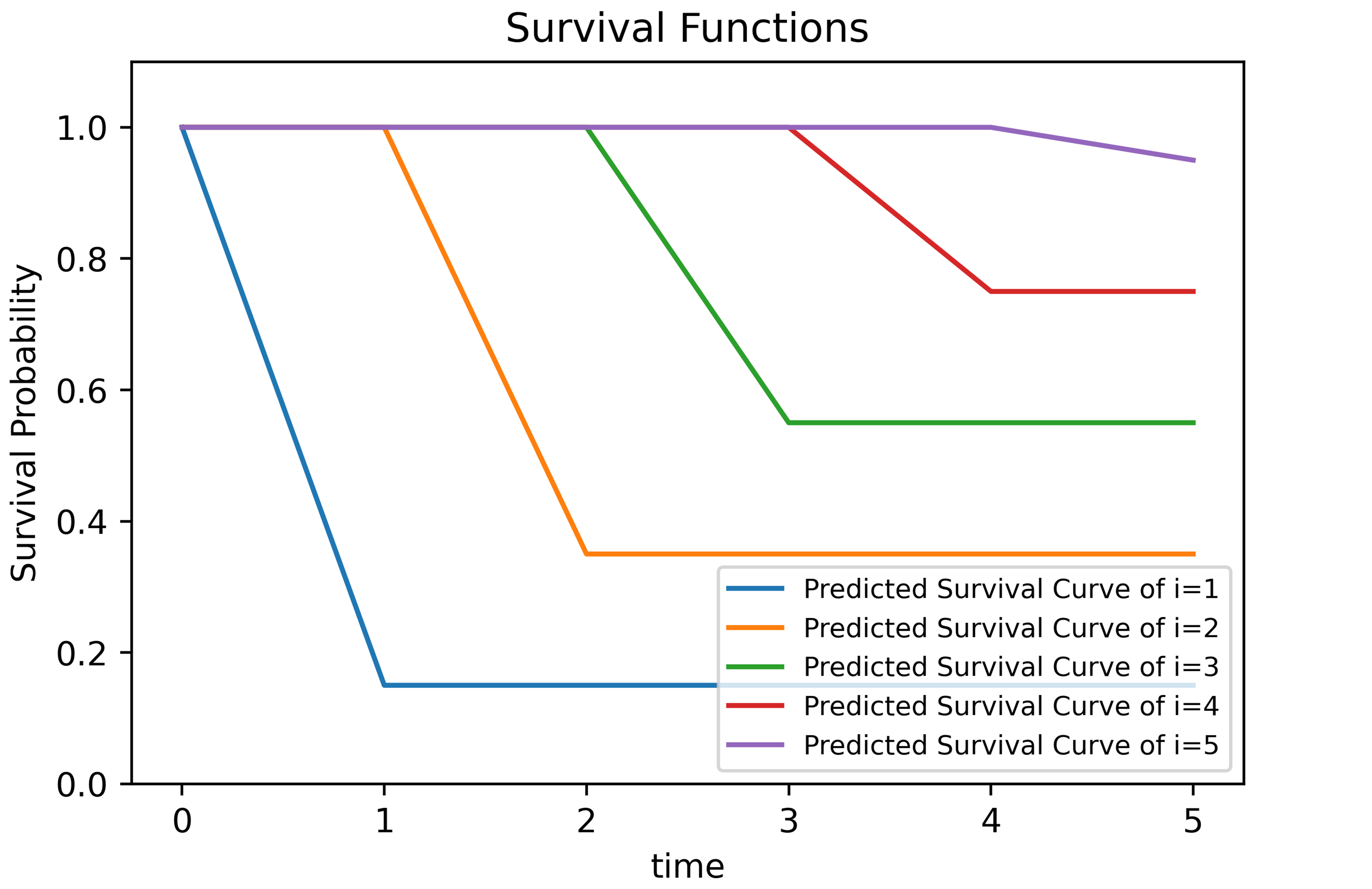}
	\end{center}
	\caption{llustration of Predicted Survival Functions.}\label{fig:dcal_issue_1}
\end{figure}

In this section we use examples to elaborate more on the disadvantage of existing calibration metrics employed by the previous state-of-the-art models including X-Cal~\citep{goldstein2020x}, RPS~\citep{kamran2021estimating}, and MCBoostSurv~\citep{becker2021multicalibration} discussed previously.
\paragraph{Example 1: X-Cal.} \label{subsec:ex1}
Assume that we have the data of the form $D=\{(\mathbf{x}_i,t_i,\delta_i)\}_{i=1}^n$ as in Table \ref{tab:dcal_issue_data} which contains 5 uncensored individuals with the event times (e.g., death time) 1 to 5.
\begin{table}[!htb]
    \centering
    \begin{tabular}{|c|c|c|c|}
    \hline
    $i$&$\mathbf{x}_i$ &$t_i$ & $\delta_i$\\ 
    \hline
    	\begin{tabular}[c]{@{}c@{}} 
    	1\\ 2\\ 3\\ 4\\ 5
    	\end{tabular} 
    	&
    	\begin{tabular}[c]{@{}c@{}} 
    	$\mathbf{x}_1$\\ $\mathbf{x}_2$\\ $\mathbf{x}_3$\\ $\mathbf{x}_4$\\ $\mathbf{x}_5$
    	\end{tabular} 
    	&
    	\begin{tabular}[c]{@{}c@{}} 
    	1\\ 2\\ 3\\ 4\\ 5
    	\end{tabular} 
    	&
    	\begin{tabular}[c]{@{}c@{}} 
    	1\\ 1\\ 1\\ 1\\ 1
    	\end{tabular}\\
	\hline
	\end{tabular}
	\caption{Synthetic Data}
    \label{tab:dcal_issue_data}
\end{table}
Given the data (in Table \ref{tab:dcal_issue_data}), a predictive model could predict survival functions shown in Figure \ref{fig:dcal_issue_1} where $S_{\theta}(t_1|\mathbf{x}_1)=0.15$ or $F_{\theta}(t_1|\mathbf{x}_1)=0.85$. Similarly $F_{\theta}(t_2|\mathbf{x}_2)=0.65$, $F_{\theta}(t_3|\mathbf{x}_3)=0.45$,
$F_{\theta}(t_4|\mathbf{x}_4)=0.25$, and $F_{\theta}(t_5|\mathbf{x}_5)=0.05$.

\begin{figure}[h]
	\begin{center}
	\includegraphics[scale = 0.85]{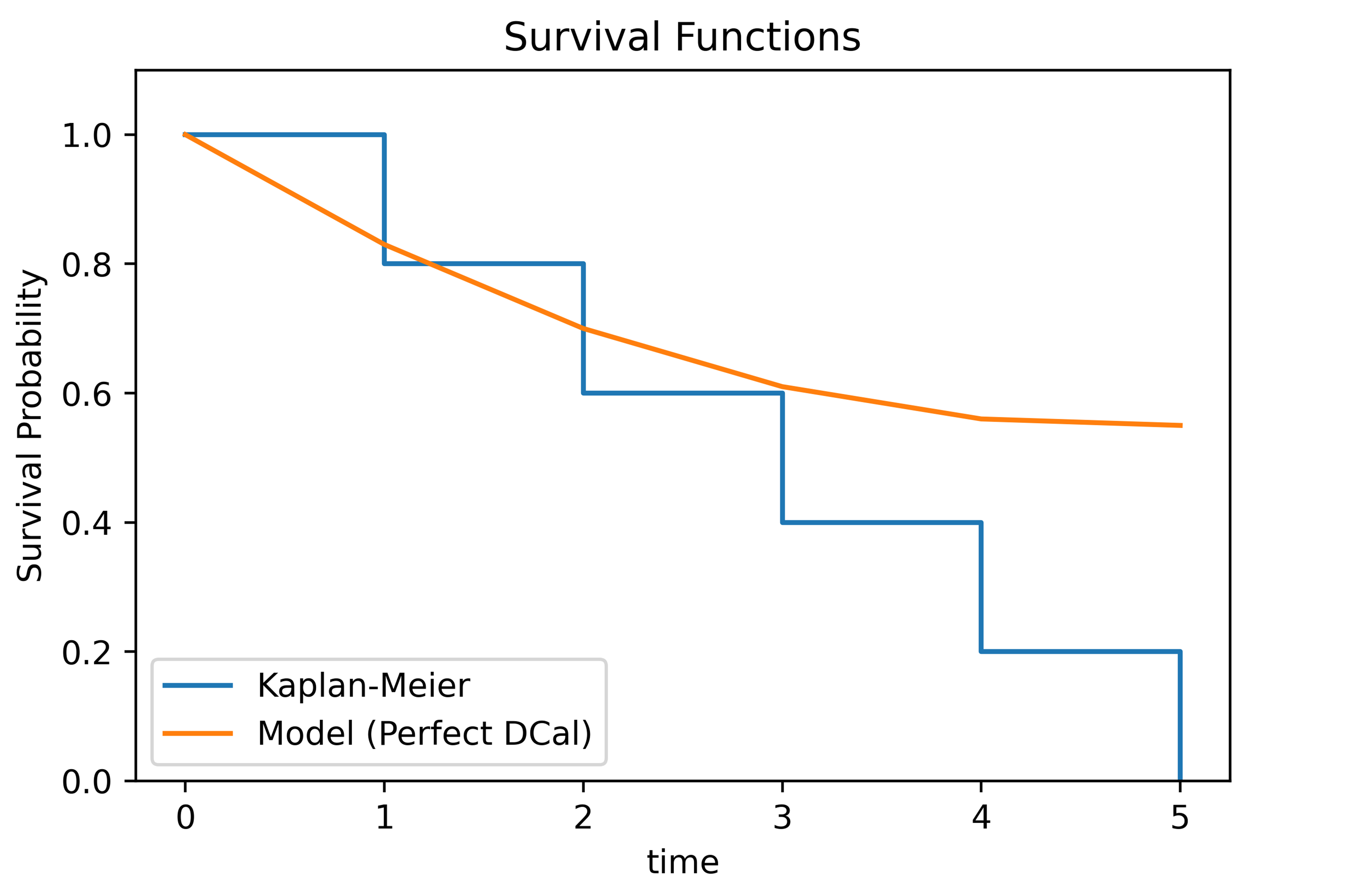}
	\end{center}
	\caption{llustration of Population-Level Predicted Survival Functions.}\label{fig:dcal_issue_2}
\end{figure}

The evaluation of these survival functions using D-Cal metric can be performed using Equation \ref{eq:dcal} (reshown in the following for convenience).
\begin{align*} 
    DCal_\theta =\sum_{I \in \mathcal{I}}
    \left( \, 
    \mathbb{E}_{(\mathbf{x}_i,t_i,\cdot) \sim D}\,
    \mathbb{I}[F_{\theta}(t_i|\mathbf{x}_i) \in I] \,\,-\, |I| 
    \,\right)^2, \tag{\ref{eq:dcal}}  
\end{align*}
By picking $\mathcal{I}$ to be a set of equal-width bins of 0.2 or \[\{[0,0.2],[0.2,0.4],[0.4,0.6],[0.6,0.8],[0.8,1]\}\], for each bin in $\mathcal{I}$ (e.g., $[0,0.2]$), there is only 1 individual's $F_{\theta}(t_i|\mathbf{x}_i)$ out of all 5 individuals' that is in the bin's range, i.e., $\mathbb{E}_{(\mathbf{x}_i,t_i,\cdot) \sim D}\,
    \mathbb{I}[F_{\theta}(t_i|\mathbf{x}_i) \in I]=\frac{1}{5}$. Therefore, $DCal_\theta =0$ since $|I|=0.2$ from having 0.2 width bin; D-Cal then suggests that the predicted survival functions are well-calibrated. 
However, given the predicted survival functions, we can have the population-level survival function as shown in the gold curve in the Figure \ref{fig:dcal_issue_2} which predicts 55\% of population would survive after time 5. The curve can be contrast with the actual survival function (the blue curve of the Figure \ref{fig:dcal_issue_2}) which suggests no one would survive after time 5.

\begin{figure}[h]
	\begin{center}
	\includegraphics[scale = 0.7]{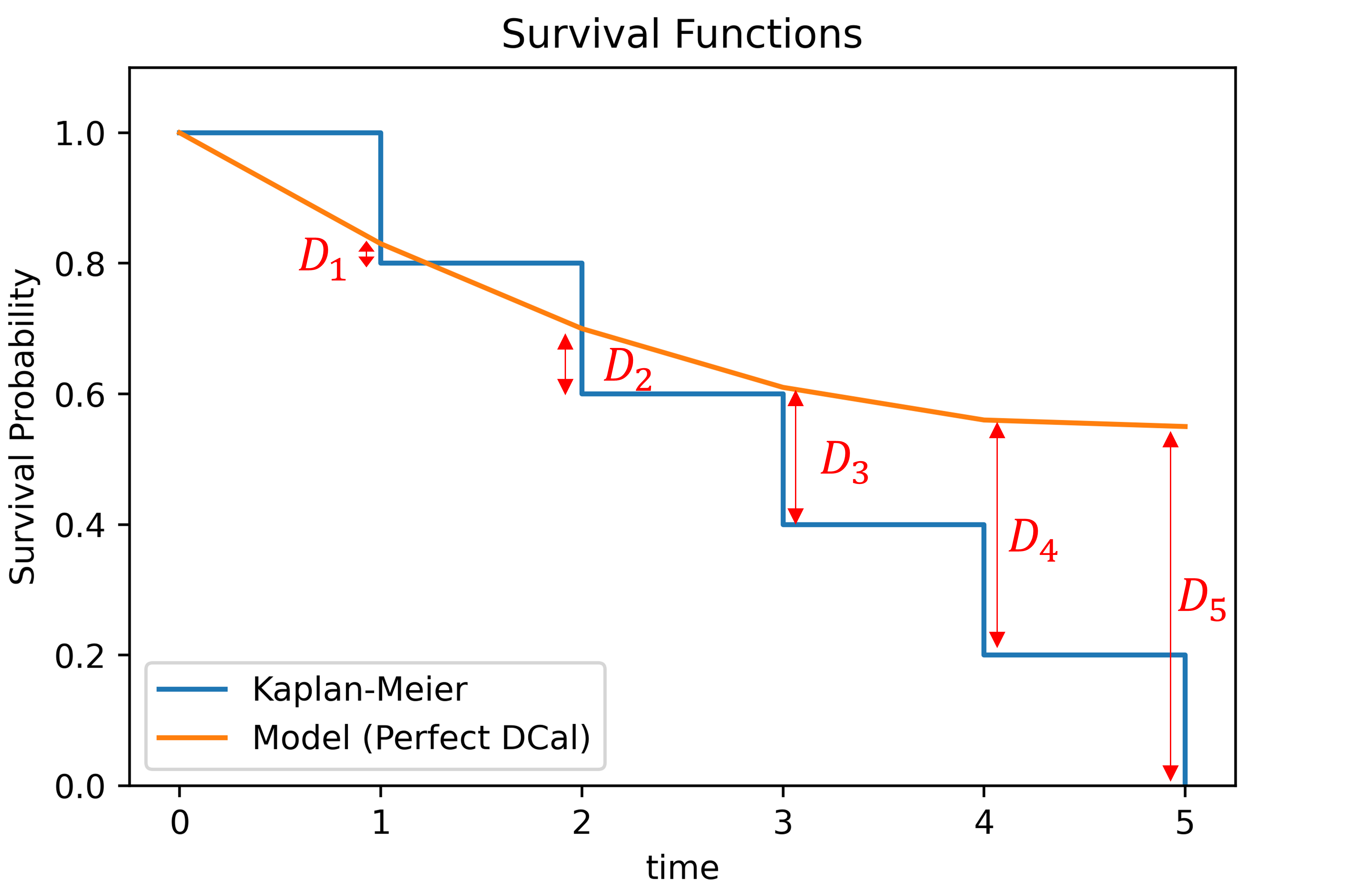}
	\end{center}
	\caption{GRADUATE System's Constraint.}\label{fig:GRADUATE_advantage_Dcal1}
\end{figure}

On the other hand, our GRADUATE system requires that at every timestep $t \in \{0,\ldots,\tau \}$ and subpopulation $D'_i$ where $i \in \{0,\ldots,m-1\}$, the marginalization of the predicted individualized survival functions is close to the survival function from the Kaplan-Meier estimator with the constraints in Problem \ref{eq:model_3} (separately shown in Equation \ref{eq:GRADUATE_advantage}).
\begin{equation} \label{eq:GRADUATE_advantage}
    f\left[ \left(\mathbb{E}_{(\mathbf{x},\cdot,\cdot)\sim D'_i} \left[\, S_{\theta}(t|\mathbf{x}) \,\right]\right)_{t=0}^\tau, \left(S_{KM,D'_i}(t)\right)_{t=0}^{\tau} \right] \leq c_i \quad i=0,\ldots,m-1.
\end{equation}
Function $f\left[ \left(\mathbb{E}_{(\mathbf{x},\cdot,\cdot)\sim D'_i} \left[\, S_{\theta}(t|\mathbf{x}) \,\right]\right)_{t=0}^\tau, \left(S_{KM,D'_i}(t)\right)_{t=0}^{\tau} \right]$ accumulates all the discrepancy between the marginalization of the predicted individualized survival functions and the survival function from the Kaplan-Meier estimator at every timestep (as in Figure \ref{fig:GRADUATE_advantage_Dcal1}). 
By using $L^2$-Norm method for our GRADUATE system, referring to Figure \ref{fig:GRADUATE_advantage_Dcal1} we constrain the average of these discrepancies (i.e, $\frac{1}{6}\left(D_0^2+D_1^2+\ldots+D_5^2\right)$) to be less than a specified constant $c_i$. The mismatch between the predicted and actual survival curves for all time points will be reflected through such averaged quantity. (e.g., large magnitude of any $D_t$ will cause the average to be large.) Therefore, our GRADUATE can avoid the problem faced by the other calibration methods by reducing the mismatch between the predicted and actual survival curves for all time points.   

Similarly, for GRADUATE with variance-adjusted norm method, the discrepancies between the marginalization of the predicted individualized survival functions and the survival function from the Kaplan-Meier estimator
at every timestep (i.e., $D_0,\ldots,D_5$ in \ref{fig:GRADUATE_advantage_Dcal1}) are adjusted with the variance of survival function from Kaplan-Meier estimator before constraining the maximum value among all the time points to be less than a specified constant $c_i$ (i.e., $\max_{t:0\leq t \leq 5}\frac{\left|D_t\right|}{V\hat{a}r\left[S_{KM,D}(t)\right]} \leq c_i$). Consequently, the mismatches of the survival curves for all the timesteps will be controlled by limiting their maximum value to be less than $c_i$.  

\begin{figure}[h]
	\begin{center}
	\includegraphics[scale = 0.5]{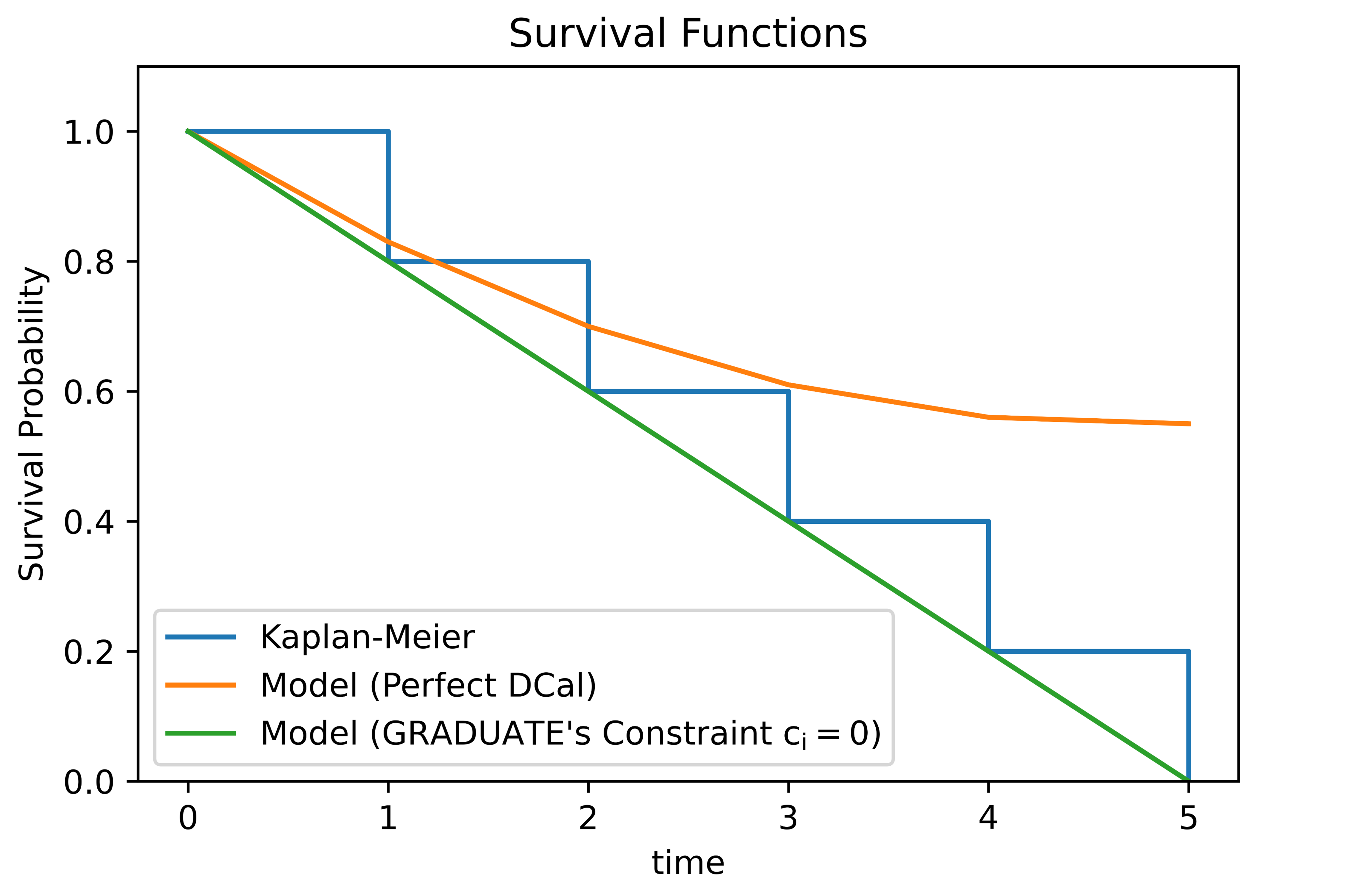}
	\end{center}
	\caption{GRADUATE System's Constraint $c_i=0$ will make the predicted survival curve to be the same as curve from Kaplan-Meier}\label{fig:GRADUATE_advantage_Dcal2}
\end{figure}

\begin{figure}[h]
	\begin{center}
	\includegraphics[scale = 0.7]{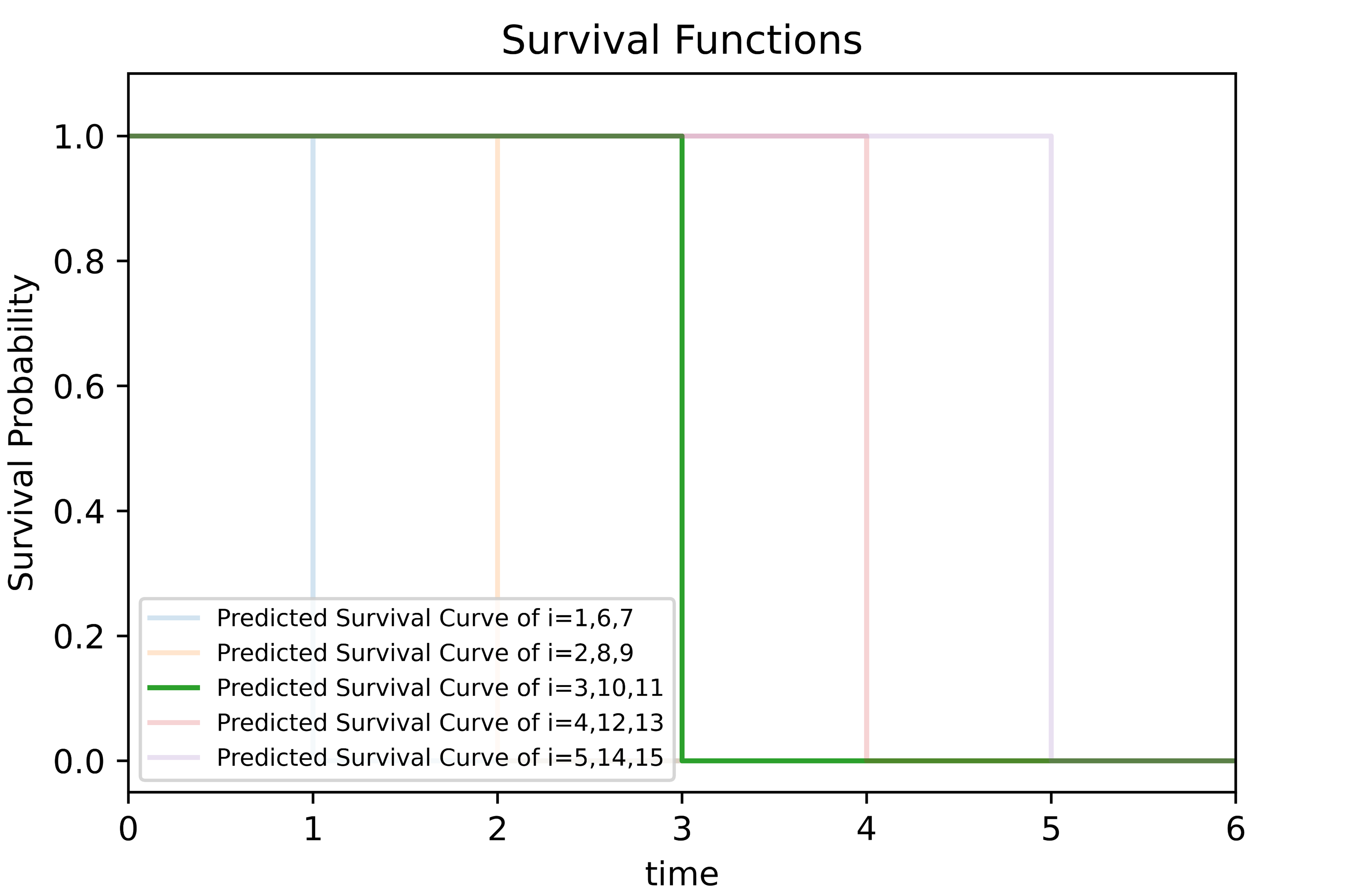}
	\end{center}
	\caption{llustration of Predicted Survival Functions.}\label{fig:rps_brier_issue_1}
\end{figure}

\begin{figure}[h]
	\begin{center}
	\includegraphics[scale = 0.85]{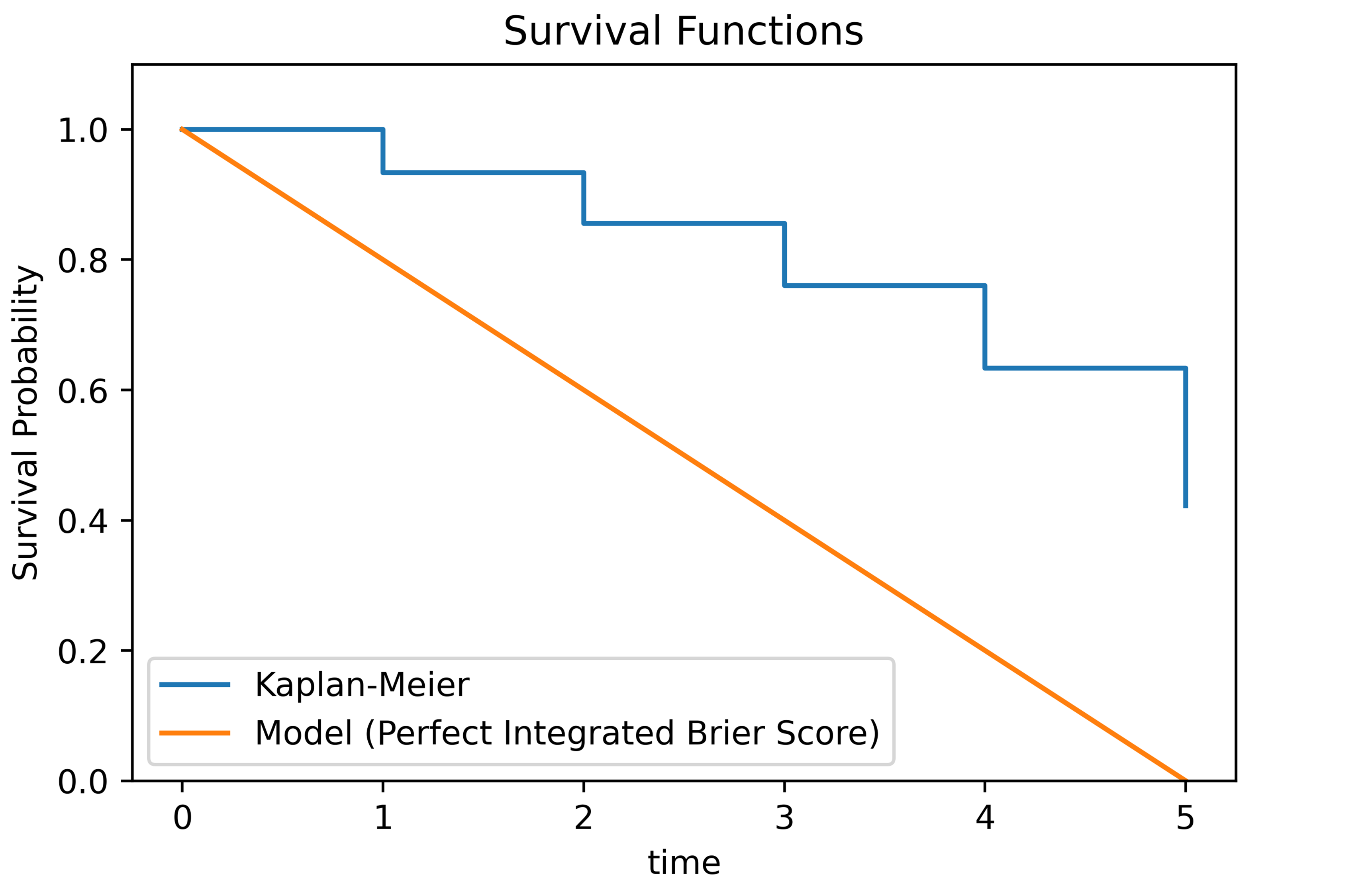}
	\end{center}
	\caption{llustration of Predicted Survival Functions.}\label{fig:rps_brier_issue_2}
\end{figure}

\begin{figure}[h]
	\begin{center}
	\includegraphics[scale = 0.65]{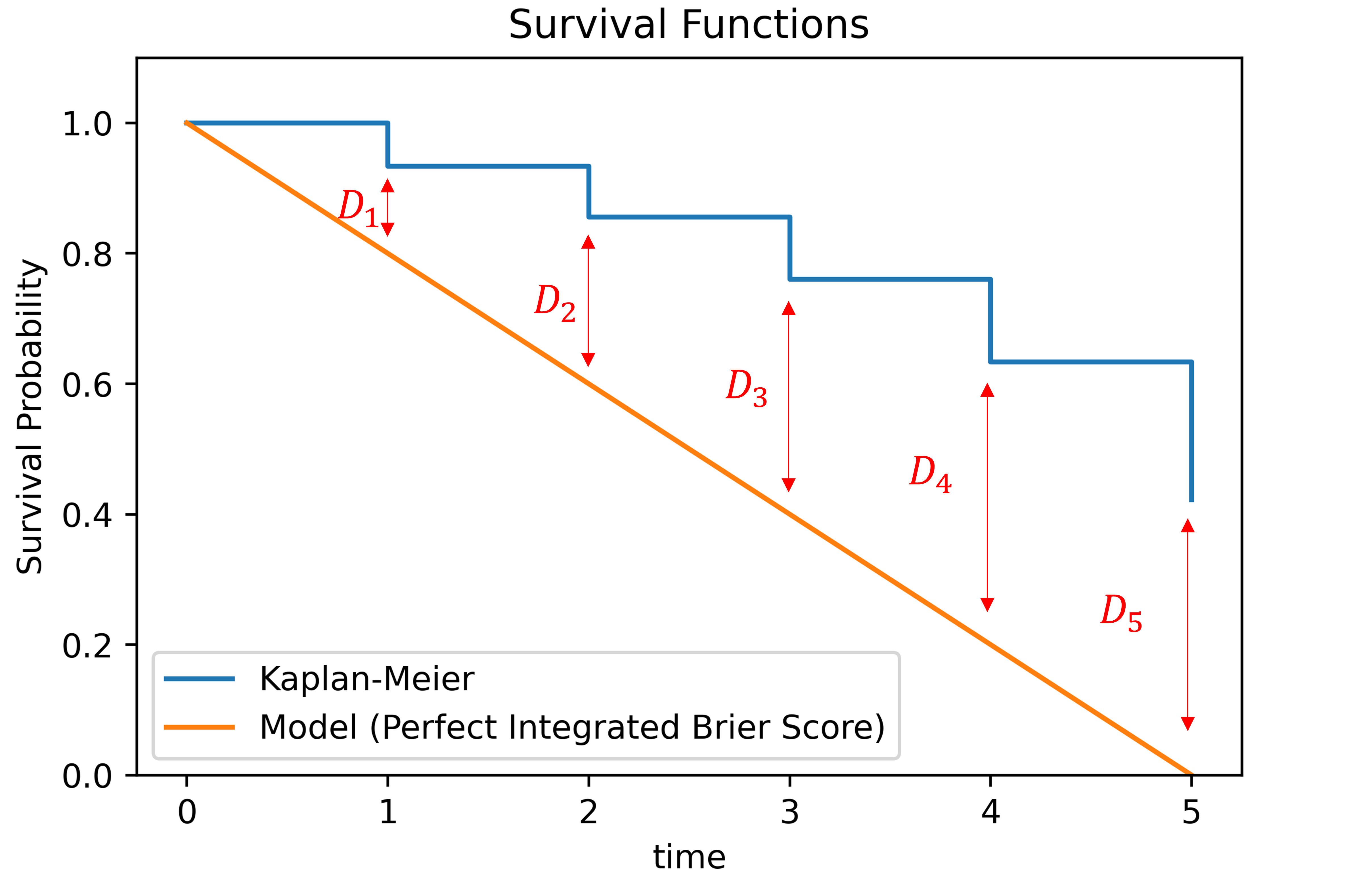}
	\end{center}
	\caption{GRADUATE System's Constraint.}\label{fig:GRADUATE_advantage_ibs1}
\end{figure}

\begin{figure}[h]
	\begin{center}
	\includegraphics[scale = 0.6]{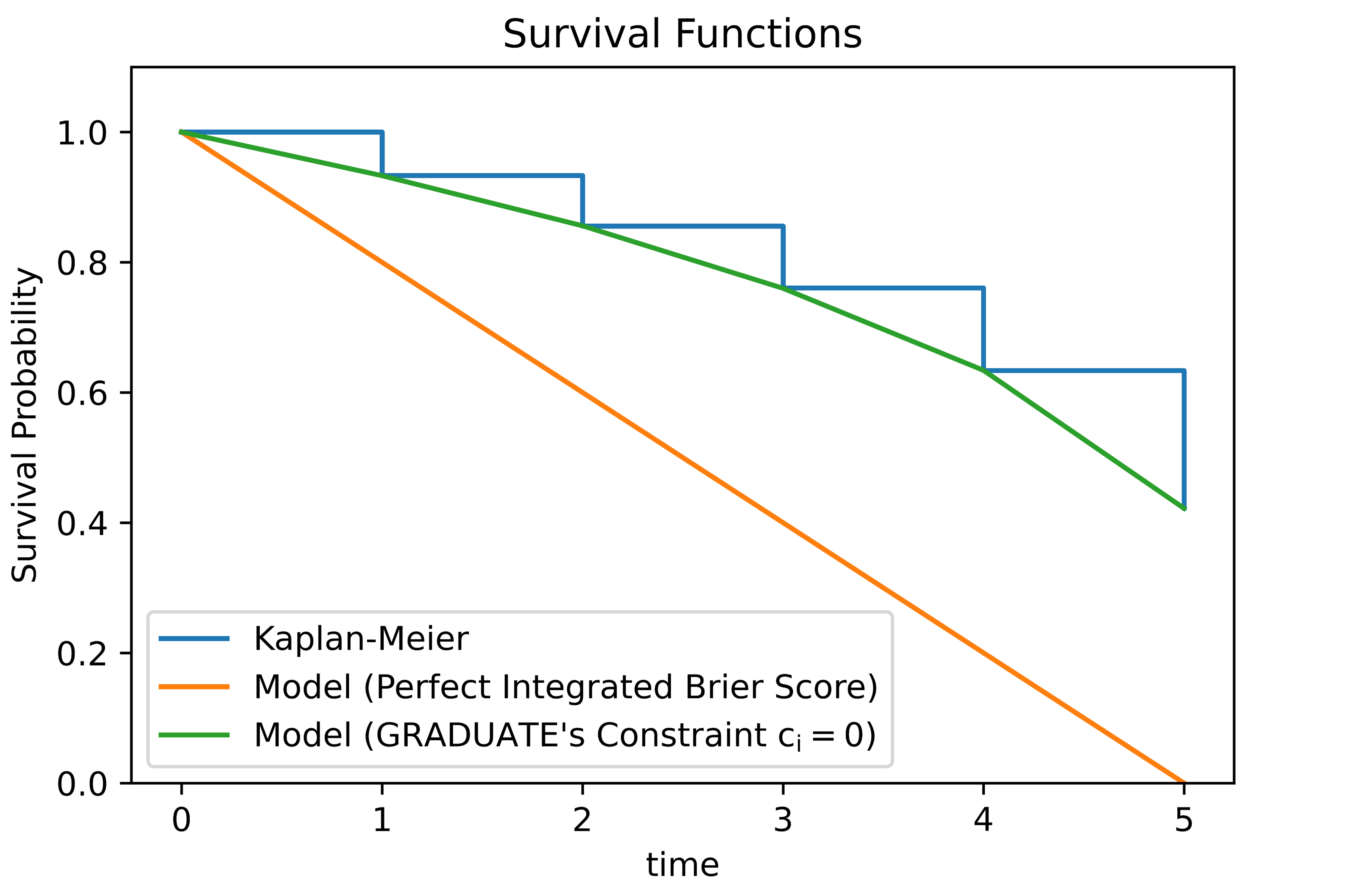}
	\end{center}
	\caption{GRADUATE System's Constraint $c_i=0$ will make the predicted survival curve to be the same as curve from Kaplan-Meier}\label{fig:GRADUATE_advantage_ibs2}
\end{figure}

\begin{figure}[h]
	\begin{center}
	\includegraphics[scale = 0.6]{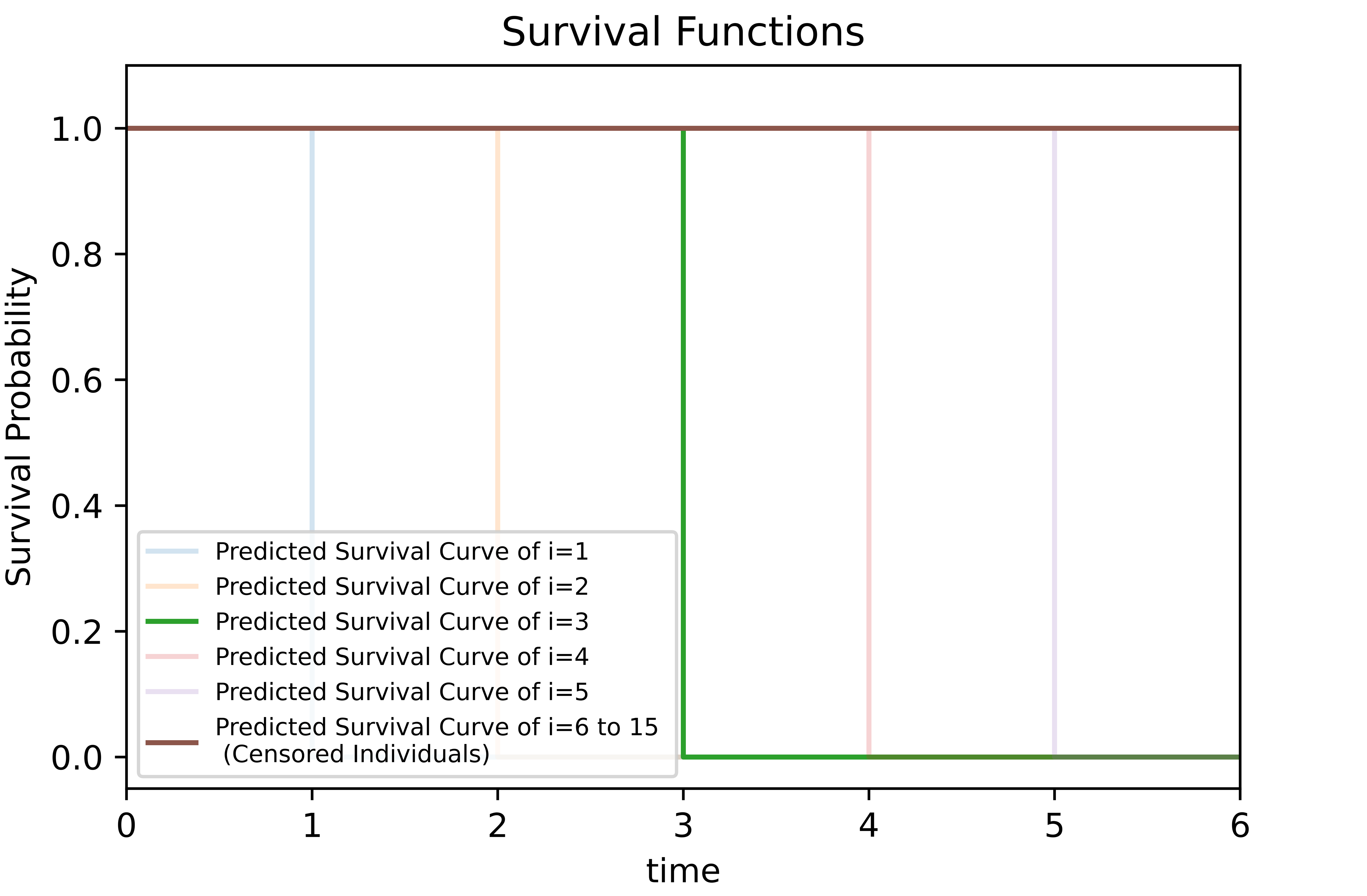}
	\end{center}
	\caption{llustration of Predicted Survival Functions.}\label{fig:rps_issue_1}
\end{figure}

\begin{figure}[h]
	\begin{center}
	\includegraphics[scale = 0.55]{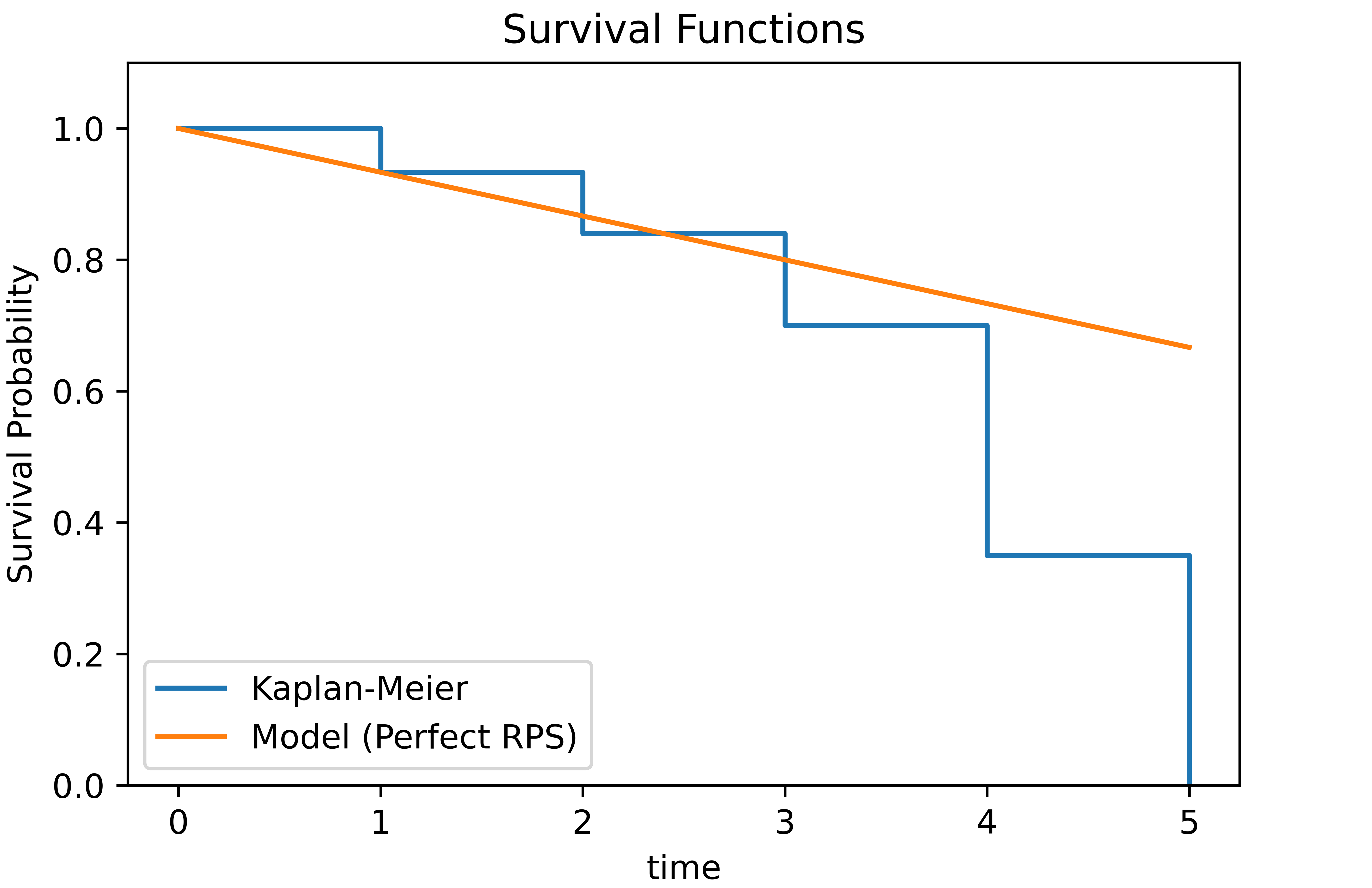}
	\end{center}
	\caption{llustration of Predicted Survival Functions.}\label{fig:rps_issue_2}
\end{figure}

\begin{figure}[H]
	\begin{center}
	\includegraphics[scale = 0.55]{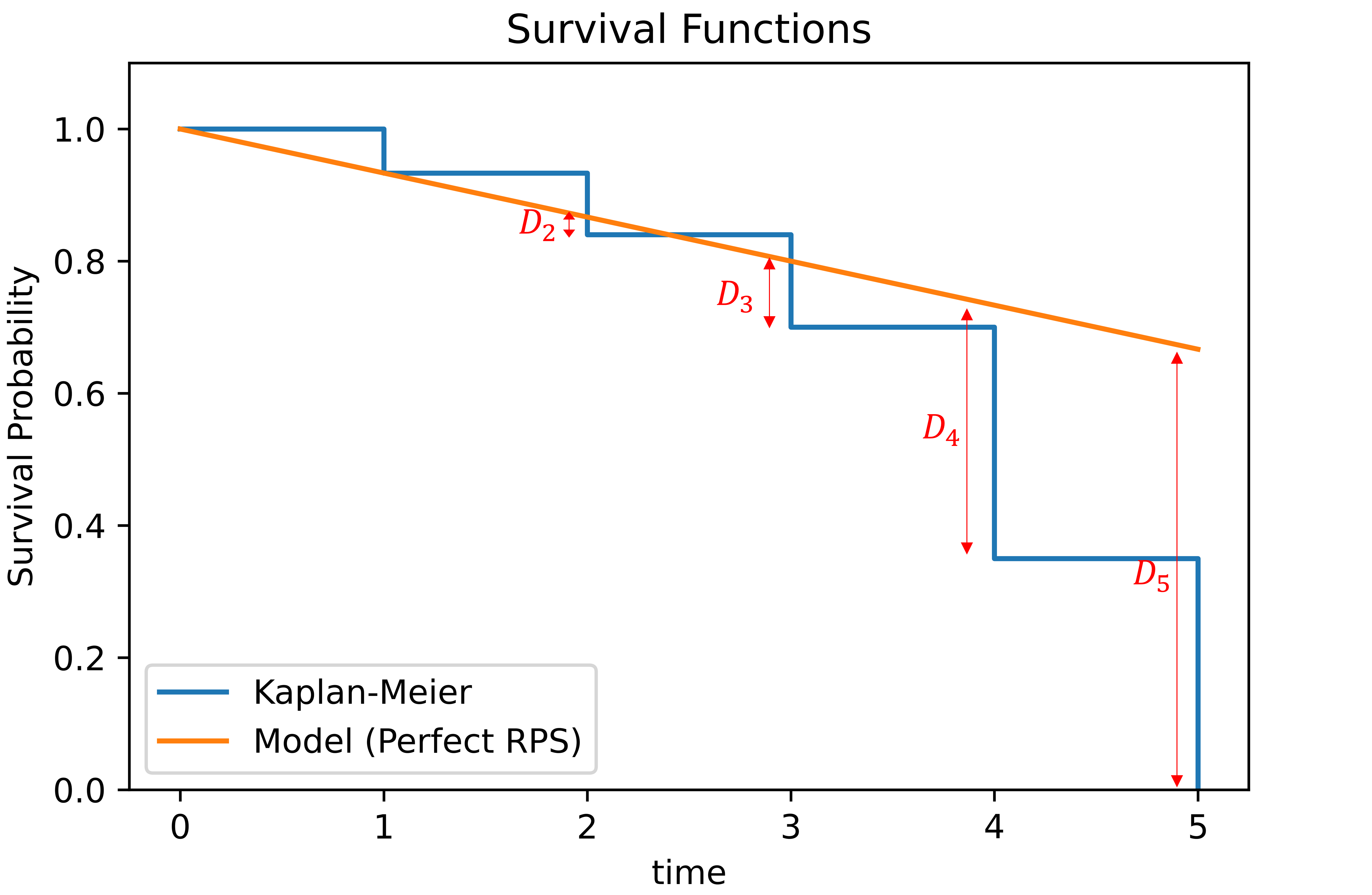}
	\end{center}
	\caption{GRADUATE System's Constraint.}\label{fig:GRADUATE_advantage_rps1}
\end{figure}

\begin{figure}[H]
	\begin{center}
	\includegraphics[scale = 0.55]{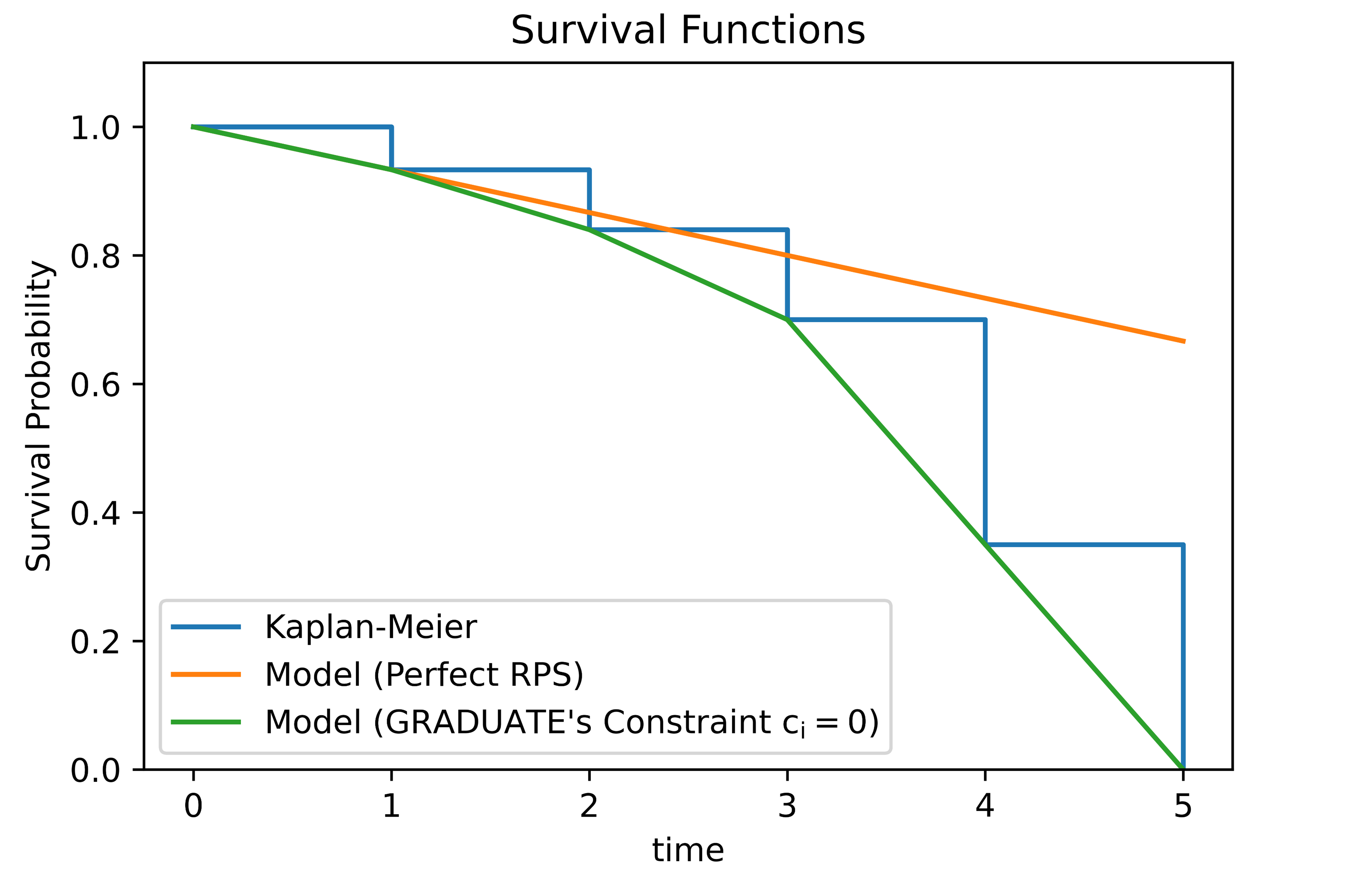}
	\end{center}
	\caption{GRADUATE System's Constraint $c_i=0$ will make the predicted survival curve to be the same as curve from Kaplan-Meier}\label{fig:GRADUATE_advantage_rps2}
\end{figure}

With constraints being 0 ($c_i=0$) , either $L^2$-Norm or variance-adjusted norm methods will enforce the marginalized predicted survival function to be the same as the survival function from Kaplan-Meier estimator as shown in Figure \ref{fig:GRADUATE_advantage_Dcal2}.   
\paragraph{Example 2: MCBoostSurv.}
To show the weakness of the calibration metric, Brier score, employed by MCBoostSurv, we use the synthetic data in Table \ref{tab:rps_brier_issue_data} which contains both censored and uncensored data of 15 individuals.
\begin{table}[!htb]
    \centering
    \begin{tabular}{|c|c|c|c|c|c|c|c|c|c|c|c|c|c|}
    \cline{1-4} \cline{6-9} \cline{11-14}
    $i$&$\mathbf{x}_i$ &$t_i$ & $\delta_i$ & & $i$&$\mathbf{x}_i$ &$t_i$ & $\delta_i$ & & $i$&$\mathbf{x}_i$ &$t_i$ & $\delta_i$\\ 
    \cline{1-4} \cline{6-9} \cline{11-14}
    	\begin{tabular}[c]{@{}c@{}} 
    	1\\ 2\\ 3\\ 4\\ 5
    	\end{tabular} 
    	&
    	\begin{tabular}[c]{@{}c@{}} 
    	$\mathbf{x}_1$\\ $\mathbf{x}_2$\\ $\mathbf{x}_3$\\ $\mathbf{x}_4$\\ $\mathbf{x}_5$
    	\end{tabular} 
    	&
    	\begin{tabular}[c]{@{}c@{}} 
    	1\\ 2\\ 3\\ 4\\ 5
    	\end{tabular} 
    	&
    	\begin{tabular}[c]{@{}c@{}} 
    	1\\ 1\\ 1\\ 1\\ 1
    	\end{tabular}
    	&
    	&
    	\begin{tabular}[c]{@{}c@{}} 
    	6\\ 7\\ 8\\ 9\\ 10
    	\end{tabular} 
    	&
    	\begin{tabular}[c]{@{}c@{}} 
    	$\mathbf{x}_6$\\ $\mathbf{x}_7$\\ $\mathbf{x}_8$\\ $\mathbf{x}_9$\\ $\mathbf{x}_{10}$
    	\end{tabular} 
    	&
    	\begin{tabular}[c]{@{}c@{}} 
    	1\\ 2\\ 3\\ 4\\ 5
    	\end{tabular} 
    	&
    	\begin{tabular}[c]{@{}c@{}} 
    	0\\ 0\\ 0\\ 0\\ 0
    	\end{tabular}
    	&
    	&
    	\begin{tabular}[c]{@{}c@{}} 
    	11\\ 12\\ 13\\ 14\\ 15
    	\end{tabular} 
    	&
    	\begin{tabular}[c]{@{}c@{}} 
    	$\mathbf{x}_{11}$\\ $\mathbf{x}_{12}$\\ $\mathbf{x}_{13}$\\ $\mathbf{x}_{14}$\\ $\mathbf{x}_{15}$
    	\end{tabular} 
    	&
    	\begin{tabular}[c]{@{}c@{}} 
    	1\\ 2\\ 3\\ 4\\ 5
    	\end{tabular} 
    	&
    	\begin{tabular}[c]{@{}c@{}} 
    	0\\ 0\\ 0\\ 0\\ 0
    	\end{tabular}
    	 \\
	\cline{1-4} \cline{6-9} \cline{11-14}
	\end{tabular}
	\caption{Synthetic Data}
    \label{tab:rps_brier_issue_data}
\end{table}
Suppose a predictive model yields predicted survival functions in Figure \ref{fig:rps_brier_issue_1}. Please notice that they are step functions taking value of 1 before the event or censoring time and 0 afterwards. (NB: MCBoostSurv has assumed, for censored individuals ($\delta_i=0$) whose $t_i \leq t$, that $S_\theta(t|\mathbf{x}_i)=0$.)    
Therefore, with Brier Score (Equation \ref{eq:brier} repeated here), both of the terms $\mathbb{I}[t_i\!\leq\! t,\delta_i\!=\!1] \,
    \frac{\left(0-S_\theta(t|\mathbf{x}_i)\right)^2}{\hat{G}(t_i)} $ and $\mathbb{I}[t_i\!>\!t]
    \,\frac{\left( 1-S_\theta(t|\mathbf{x}_i) \right)^2}{\hat{G}(t)}$ will be zero.
\begin{equation} 
    Brier_\theta(t)=\frac{1}{|D|}\sum_{(\mathbf{x}_i,t_i,\delta_i) \in D}
    \mathbb{I}[t_i\!\leq\! t,\delta_i\!=\!1] \,
    \frac{\left(0-S_\theta(t|\mathbf{x}_i)\right)^2}{\hat{G}(t_i)} 
    \,+\, \mathbb{I}[t_i\!>\!t]
    \,\frac{\left( 1-S_\theta(t|\mathbf{x}_i) \right)^2}{\hat{G}(t)} \tag{\ref{eq:brier}}
\end{equation}
Consequently, Brier score suggest that the predicted survival curves are well-calibrated.
As before in Section \ref{subsec:ex1}, the marginalized survival function for population can be shown as the gold curve in Figure \ref{fig:rps_brier_issue_2} which predicts 0\% of population would survive after time 5. On the other hand, the actual survival curve from Kaplan-Meier estimator (blue curve in Figure \ref{fig:rps_brier_issue_2}) suggests 42\% of population survives after time 5. 
As before in Section \ref{subsec:ex1}, Figure \ref{fig:GRADUATE_advantage_ibs1} shows all the discrepancies between the marginalization of the predicted individualized survival functions and the survival function from the Kaplan-Meier estimator at every timestep which would be constrained either by $L^2$-Norm or variance-adjusted norm methods in our GRADUATE system. 
Therefore, with $c_i=0$ constraints, the marginalized predicted survival function will be the same as the survival function from Kaplan-Meier estimator as shown in Figure \ref{fig:GRADUATE_advantage_ibs2}.      
\paragraph{Example 3: RPS.}
Again, we use the synthetic data in Table \ref{tab:rps_issue_data} which contains both censored and uncensored data of 15 individuals to show the weakness of RPS. 
\begin{table}[!htb]
    \centering
    \begin{tabular}{|c|c|c|c|c|c|c|c|c|c|c|c|c|c|}
    \cline{1-4} \cline{6-9} \cline{11-14}
    $i$&$\mathbf{x}_i$ &$t_i$ & $\delta_i$ & & $i$&$\mathbf{x}_i$ &$t_i$ & $\delta_i$ & & $i$&$\mathbf{x}_i$ &$t_i$ & $\delta_i$\\ 
    \cline{1-4} \cline{6-9} \cline{11-14}
    	\begin{tabular}[c]{@{}c@{}} 
    	1\\ 2\\ 3\\ 4\\ 5
    	\end{tabular} 
    	&
    	\begin{tabular}[c]{@{}c@{}} 
    	$\mathbf{x}_1$\\ $\mathbf{x}_2$\\ $\mathbf{x}_3$\\ $\mathbf{x}_4$\\ $\mathbf{x}_5$
    	\end{tabular} 
    	&
    	\begin{tabular}[c]{@{}c@{}} 
    	1\\ 2\\ 3\\ 4\\ 5
    	\end{tabular} 
    	&
    	\begin{tabular}[c]{@{}c@{}} 
    	1\\ 1\\ 1\\ 1\\ 1
    	\end{tabular}
    	&
    	&
    	\begin{tabular}[c]{@{}c@{}} 
    	6\\ 7\\ 8\\ 9\\ 10
    	\end{tabular} 
    	&
    	\begin{tabular}[c]{@{}c@{}} 
    	$\mathbf{x}_6$\\ $\mathbf{x}_7$\\ $\mathbf{x}_8$\\ $\mathbf{x}_9$\\ $\mathbf{x}_{10}$
    	\end{tabular} 
    	&
    	\begin{tabular}[c]{@{}c@{}} 
    	1\\ 1\\ 1\\ 1\\ 2
    	\end{tabular} 
    	&
    	\begin{tabular}[c]{@{}c@{}} 
    	0\\ 0\\ 0\\ 0\\ 0
    	\end{tabular}
    	&
    	&
    	\begin{tabular}[c]{@{}c@{}} 
    	11\\ 12\\ 13\\ 14\\ 15
    	\end{tabular} 
    	&
    	\begin{tabular}[c]{@{}c@{}} 
    	$\mathbf{x}_{11}$\\ $\mathbf{x}_{12}$\\ $\mathbf{x}_{13}$\\ $\mathbf{x}_{14}$\\ $\mathbf{x}_{15}$
    	\end{tabular} 
    	&
    	\begin{tabular}[c]{@{}c@{}} 
    	2\\ 2\\ 3\\ 3\\ 3
    	\end{tabular} 
    	&
    	\begin{tabular}[c]{@{}c@{}} 
    	0\\ 0\\ 0\\ 0\\ 0
    	\end{tabular}
    	 \\
	\cline{1-4} \cline{6-9} \cline{11-14}
	\end{tabular}
	\caption{Synthetic Data}
    \label{tab:rps_issue_data}
\end{table}
Suppose a predictive model yields predicted survival functions in Figure \ref{fig:rps_issue_1}. Please notice that they are step functions taking value of 1 before the event and 0 afterwards. For a censored individual, the survival function takes value of 1 before and after the censoring time. (NB: RPS has assumed, for censored individuals ($\delta_i=0$) whose $t_i \leq t$, that $S_\theta(t|\mathbf{x}_i)=1$.)    
With RPS as in Equation \ref{eq:rps} (repeated in the following), both of the summations \linebreak$\sum_{t=0}^\tau
    \left(S_\theta(t|\mathbf{x}_i)-\mathbb{I}[t<t_i]\right)^2$ and  $\sum_{t=0}^{t_i}\left(S_\theta(t|\mathbf{x}_i)-1 \right)^2$ will be zero.
\begin{equation} 
    RPS_\theta =\sum_{(\mathbf{x}_i,t_i,\delta_i) \in D}
    \left[ \delta_i \sum_{t=0}^\tau
    \left(S_\theta(t|\mathbf{x}_i)-\mathbb{I}[t<t_i]\right)^2
    + \, (1-\delta_i)\sum_{t=0}^{t_i}\left(S_\theta(t|\mathbf{x}_i)-1 \right)^2 \right] \tag{\ref{eq:rps}}
\end{equation}
Again, we can see the contradictory result from the marginalized survival function and the actual survival curve from Kaplan-Meier estimator. The marginalized survival function for population shown as the gold curve in Figure \ref{fig:rps_issue_2} predicts 67\% of population would survive after time 5. On the other hand, the actual survival curve from Kaplan-Meier estimator (blue curve in Figure \ref{fig:rps_issue_2}) suggests 0\% of population survives after time 5. 
Figure \ref{fig:GRADUATE_advantage_rps1} shows all the discrepancies between the marginalization of the predicted individualized survival functions and the survival function from the Kaplan-Meier estimator at every timestep which would be constrained either by $L^2$-Norm or variance-adjusted norm methods in our GRADUATE system. 
Therefore, with $c_i=0$ constraints, the marginalized predicted survival function will be the same as the survival function from Kaplan-Meier estimator as shown in Figure \ref{fig:GRADUATE_advantage_rps2}.      
\section{Practical Implications} \label{sec:practical_implications}

GRADUATE is a novel model for survival analysis which has managerial applications in varied domains such as financial credit \citep{stepanova2002survival}, innovation management \citep{harhoff2009duration}, and narcotics abuse \citep{dekimpe1998long}. In credit risk modeling, survival analysis allows companies to assess the probability of default over time, facilitating strategic customer selection for maximizing profitability. Additionally, this methodology finds application in pricing bonds and other financial investments \citep{stepanova2002survival}. Within the medical realm, survival analysis serves as a valuable tool for aiding healthcare management planning. For instance, \cite{sabouri2017screening} investigate screening strategies for patients on the kidney transplant waiting list using survival analysis. 

Survival analysis systems typically prioritize discriminative performance, focusing on accurately ranking individuals based on their predicted probabilities of event occurrence compared to observed outcomes. However, particularly in clinical contexts, it is equally vital for these systems to be well-calibrated~\citep{pepe2013methods}, ensuring that predicted probabilities closely match ground-truth probabilities across all time points.

\begin{figure}[h]
	\begin{center}
	\includegraphics[scale = 0.6]{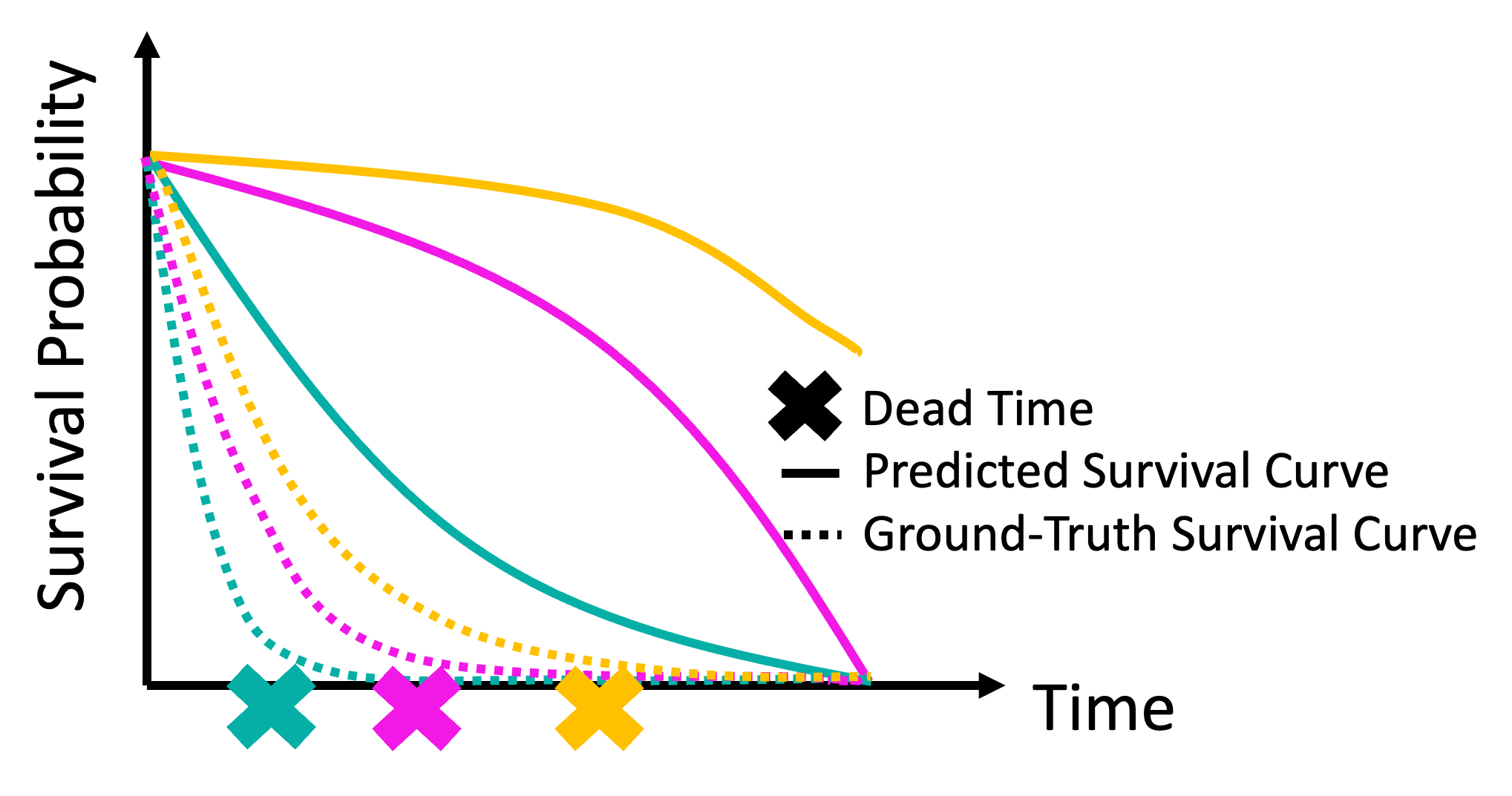}
	\end{center}
	\caption{Illustration of discrimination and calibration. Three crosses indicate the death times of three distinct individuals. The solid lines represent the predicted survival curves of these individuals, while the dotted lines depict the true survival curves.}\label{fig:discri_calib} 
\end{figure}

To exemplify discriminative performance, consider an example depicted in Figure \ref{fig:discri_calib}. The true (dotted) survival curve of the cyan individual lies beneath the other dotted lines, indicating that this individual's survival probabilities are the lowest. Hence, we expect the cyan individual to die before the others. This expectation is consistent with the ground truth in which the cyan individual's time of death, as indicated by the cyan cross in Figure~\ref{fig:discri_calib}, is leftmost on the x-axis. Furthermore, the purple dotted curve lies above the cyan dotted curve but below the gold dotted curve. Therefore, we anticipate that the purple individual will perish after the cyan individual but before the gold individual. This anticipation aligns with the ground truth, where the purple individual's time of death, indicated by the purple cross on the x-axis in Figure~\ref{fig:discri_calib}, occurs after the cyan individual's but before the gold individual's. This arrangement of curves is also observed in the predicted survival curves generated by the model. Therefore, when considering the time of death for these three individuals, the model correctly rank them by relative ordering of each individual's predicted survival curve. This demonstrates the model's satisfactory discriminative performance.

However, there is a significant discrepancy between the predicted survival curves and the ground-truth survival curve. Specifically, the model tends to overestimate survival probabilities, as indicated by the predicted survival curve consistently being notably higher than the ground-truth survival curve for each individual in Figure~\ref{fig:discri_calib}. Consequently, despite achieving excellent discriminative performance, the model exhibits poor calibration, as demonstrated by the disparity between the predicted and ground-truth survival curves.


In the context of medical applications, inaccurately calibrated systems like the one shown in Figure~\ref{fig:discri_calib} can result in potentially harmful decisions. For instance, a physician using the solid survival curve of the orange individual may misinform the patient about his life expectancy, falsely suggesting that he has several years remaining when he actually has limited time remaining (as indicated by the orange dotted curve).   

Cutting-edge survival analysis systems like RPS~\citep{kamran2021estimating} and X-Cal~\citep{goldstein2020x} typically leverage deep learning techniques but exhibit only marginal calibration. This implies that they are calibrated across the entire population represented in their training dataset. They adhere to the conventional supervised learning paradigm, selecting a model that minimizes expected loss on the training data. Consequently, there is a risk of these systems learning models that excel for the majority sub-population but falter on one or more minority sub-populations, whose contributions to the expected loss are comparatively small. Such a characteristic can carry significant legal and ethical ramifications, particularly if the minority sub-populations are delineated by protected characteristics like race and gender. For instance, algorithms may exhibit racial biases in allocating resources like care management \citep{obermeyer2019dissecting}.

\begin{figure}[h]
	\begin{center}
	\includegraphics[scale = 0.5]{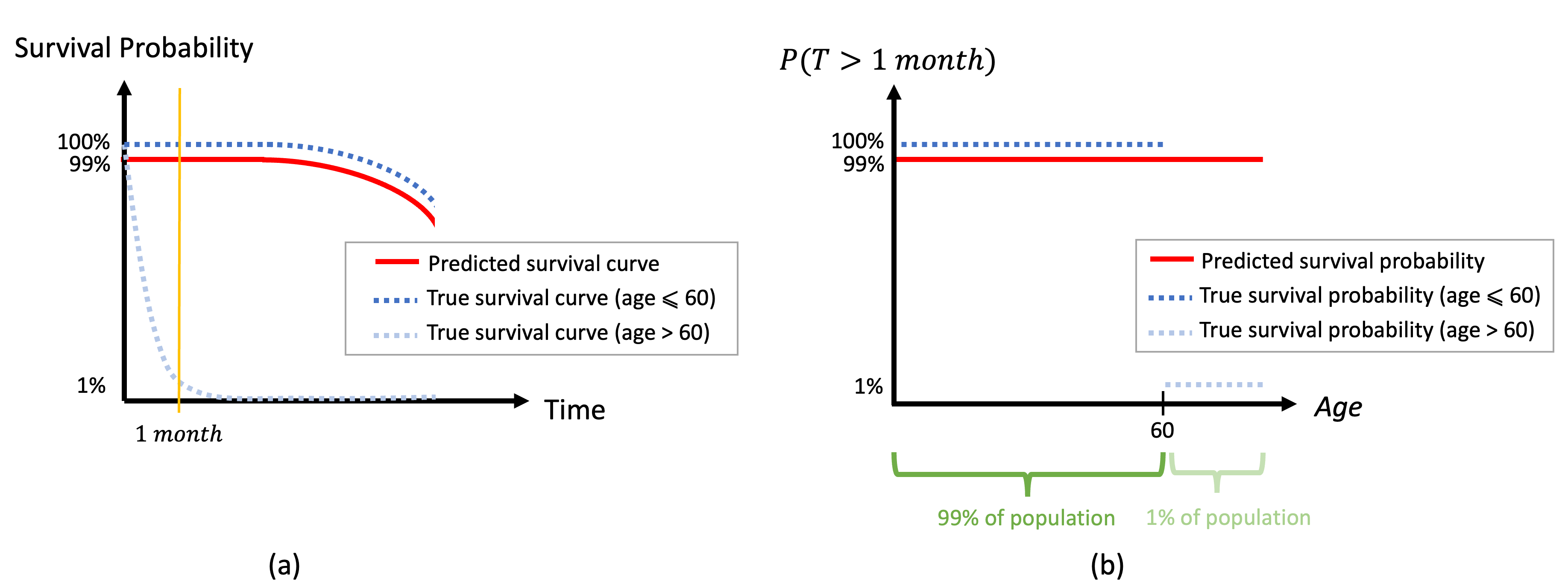}
	\end{center}
	\caption{Illustration of Predicted Survival Functions.}\label{fig:multicalib} 
\end{figure}

In the subsequent discussion, we utilize an illustrative example to elucidate the application of survival analysis models, emphasizing the importance of model calibration in minority subpopulation or multicalibration. We concentrate on a specific time from survival curves shown in Figure~\ref{fig:multicalib} (a), signifying the probabilities of individuals surviving beyond one month (as indicated by the values of the survival curves on the y-axis at the gold vertical line), or equivalently, the probabilities that the time variable $T$ exceed one month. In Figure~\ref{fig:multicalib} (b), these survival probabilities are depicted on the y-axis, while individuals' ages are plotted on the x-axis. Let us consider a hypothetical scenario where the vast majority, roughly 99\%, of individuals in our dataset are under the age of 60, with the remaining 1\% falling into the older age bracket. Consequently, a significant portion of the population comprises younger individuals. Specifically, individuals aged less than 60 demonstrate a 100\% chance of survival, whereas those aged 60 and above exhibit a mere 1\% chance. Notably, the model may incorrectly estimate the survival probability as 99\%, indicating well-calibrated predictions for younger age groups but inadequate calibration for individuals aged over 60.

GRADUATE facilitates the enforcement of calibration on these groups by incorporating constraints within a constrained optimization problem, and it trains its parameters using primal-dual gradient descent. Therefore, it achieves multicalibration, and prevents unfairness within these subpopulations.

Although there exists a system, MCBoostSurv~\citep{becker2021multicalibration}, aimed at achieving multicalibration, it does so through post-processing. However, in employing post-processing, MCBoostSurv sacrifices discriminative performance for multicalibration. To illustrate the necessity of maintaining discriminative performance alongside multicalibration, we refer once more to Figure \ref{fig:multicalib}. The model would exhibit good calibration for by individuals aged over 60 by predicting a 1-month survival probability of 1\% for all individuals within this group. However, it fails to distinguish between individuals within this demographic. For example, if there are three individuals in this demographic, we would prefer a model capable of identifying those at higher risk, enabling more efficient allocation of medical resources. Therefore, solely employing post-processing techniques for multicalibration, such as MCBoostSurv, without addressing discrimination, is still considered suboptimal.

GRADUATE attains multicalibration during its training process for discrimination. Consequently, it strikes a favorable balance between calibration and discrimination.

\section{Conclusion and Future Work} \label{sec:conclusion}

In this essay, we present GRADUATE, a novel model for survival analysis. In contrast to existing systems, GRADUATE achieves both multicalibration and a good balance between calibration and discrimination during its training process. GRADUATE is formulated as a constrained optimization problem, and trains its parameters via primal-dual gradient descent. We mathematically demonstrate that the chosen optimization technique delivers a solution that is approximately optimal and feasible with a high likelihood. In empirical comparisons to four state-of-the-art baselines on three real-world medical datasets, GRADUATE not only has better calibration both marginally and at the subpopulation level, but also has the best trade-off between discrimination and calibration. We elucidate the flaws present in the baselines' loss functions, and explain how GRADUATE obviates their shortcomings. As future work, we want to extend GRADUATE to other domains such as customer lifetime estimation and product upgrade time prediction.   
\bibliographystyle{informs2014} 
\bibliography{Reference}

\end{document}